%% file: main.tex
\documentclass[twoside,11pt]{article}

\usepackage{jmlr2e}
\usepackage{preamble_florian}

\ShortHeadings{Neural Network BaB for Neural Network Verification}{Jaeckle, Lu and Kumar}
\firstpageno{1}

\begin{document}

\title{Neural Network Branch-and-Bound for Neural Network Verification}

\author{
      \name Florian Jaeckle \thanks{equal contribution}
      \email florian@robots.ox.ac.uk \\
      \addr Department of Engineering\\
      University of Oxford\\
      Oxford OX1 3PJ
      \AND
      \name Jingyue Lu \footnotemark[1]
      \email jingyue.lu@spc.ox.ac.uk \\
      \addr Department of Statistics\\
      University of Oxford\\
      Oxford OX1 3LB
      \AND
      \name M. Pawan Kumar \email pawan@robots.ox.ac.uk \\
      \addr Department of Engineering\\
      University of Oxford\\
      Oxford OX1 3PJ}


\editor{}
\maketitle

\import{Sections/}{Abstract.tex}


\newpage
\import{Sections/}{Introduction.tex}
\import{Sections/}{Background.tex}
\import{Sections/}{GNN_Framework.tex}

\import{Sections/}{GNN_Branching.tex}
\import{Sections/}{GNN_Bounding.tex}
\import{Sections/}{Related_Work.tex}
\import{Sections/}{Experiments.tex}
\import{Sections/}{Discussion.tex}



\bibliography{bib_jmlr}

\newpage
\import{Sections/}{Appendix.tex}

\end{document}

%% file: Sections/Abstract.tex
\begin{abstract}
Many available formal verification methods have been shown to be instances of a unified Branch-and-Bound (BaB) formulation. We propose a novel machine learning framework that can be used for designing an effective branching strategy as well as for computing better lower bounds. Specifically, we learn two graph neural networks (GNN) that both directly treat the network we want to verify as a graph input and perform forward-backward passes through the GNN layers. We use one GNN to simulate the strong branching heuristic behaviour and another to compute a feasible dual solution of the convex relaxation, thereby providing a valid lower bound.

We provide a new verification dataset that is more challenging than those used in the literature, thereby providing an effective alternative for testing algorithmic improvements for verification. Whilst using just one of the GNNs leads to a reduction in verification time, we get optimal performance when combining the two GNN approaches. Our combined framework achieves a 50\% reduction in both the number of branches and the time required for verification on various convolutional networks when compared to several state-of-the-art verification methods. In addition, we show that our GNN models generalize well to harder properties on larger unseen networks.

\end{abstract}

%% file: Sections/Introduction.tex
\section{Introduction}

Despite their outstanding performances on various tasks, neural networks are found to be vulnerable to adversarial examples~\citep{goodfellow2014explaining, szegedy2013intriguing} --- examples that are similar to real inputs but ones which the neural network misclassifies with a high probability. They are obtained by applying small but deliberately chosen perturbations that are often imperceptible to the human eye. The brittleness of neural networks can have costly consequences in areas such as autonomous vehicles \citep{bojarski2016end} and personalized medicine \citep{weiss2012machine}.
When one requires robustness to adversarial examples, traditional model evaluation approaches, which test the trained model on a hold-out set, do not suffice. Instead, formal verification of properties such as adversarial robustness becomes necessary. For instance, to ensure self-driving cars make consistent correct decisions even when the input image is slightly perturbed, the required property to verify is that the underlying neural network outputs the same correct prediction for all points within a norm ball whose radius is determined by the maximum perturbation allowed.

Several methods have been proposed for verifying properties on neural networks. \cite{rudy2018} showed that many of the available methods can be viewed as instances of a unified branch-and-bound (BaB) framework. The BaB framework solves a mixed integer programming (MIP) formulation of the verification problem. In other words, the verification problem is formulated as the minimization of a linear objective under linear constraints over variables that can take real values or can be restricted to take only integral values.
A BaB algorithm consists of two key components: branching strategies and bounding methods. Branching strategies decide how the feasible domain of the MIP is recursively split into smaller subdomains. For each subdomain the bounding method then computes an upper and a lower bound of the MIP objective. If the upper bound of a subdomain is less than the lower bound of another, the latter subdomain can be pruned thereby reducing the domain for the optimal solution.

Branching strategies have a significant impact on the overall problem-solving process, as they directly influence the total number of steps, and consequently the total time, required to solve the problem at hand. The quality of a branching strategy is even more important when neural network verification problems are considered, which generally have a very large domain. Each input dimension or each activation unit can be a potential branching option and neural networks of interest often have high dimensional inputs and thousands of hidden activation units. With such a large problem domain, an effective branching strategy could mean a large reduction of the total number of branches required, and consequently of the time required to solve a problem. Developing an effective strategy is thus of significant importance to the success of BaB based neural network verification. 
So far, to the best of our knowledge, branching rules adopted by BaB based verification methods are either random selection~\citep{Katz2017, Ehlers2017} or hand-designed heuristics~\citep{wang2018formal, rudy2018, royo2019fast, journal2019}. Random selection is generally inefficient as the distribution of the best branching decision is rarely uniform. In practice, this strategy often results in an almost exhaustive search to make a verification decision. On the other hand, hand-designed heuristics often involve a trade-off between effectiveness and computational cost. For instance, strong branching is generally one of the best performing heuristics for BaB methods in terms of the number of branches, but it is computationally prohibitive as each branching decision requires an expensive exhaustive search over all possible options. The heuristics that are currently used in practice are either inspired by the corresponding dual problem when verification is formulated as an optimization problem~\citep{rudy2018, royo2019fast} or incorporating the gradient information of the neural network~\citep{wang2018formal}. These heuristics normally have better computational efficiency. However, given the complex nature of the problem domain, it is unlikely that any hand-designed heuristic is able to fully exploit the structure of the problem and the data distribution encountered in practice. As mentioned earlier, for large size neural network verification problems, a slight reduction in the quality of the branching strategy could lead to substantial increase in the total number of branches required to solve the problem. A computationally cheap but high quality branching strategy is thus much needed.

The bounding component of the BaB algorithm consists of two parts: finding upper and lower bounds.
The former is efficient to compute as it involves evaluating the objective for any feasible solution. In contrast, the lower bound computation requires solving a large convex relaxation. Typically, the relaxation is solved using either commercial solvers such as Gurobi~\citep{gurobi} or traditional optimization algorithms such as subgradient descent~\citep{dvijotham2018dual} or proximal minimization~\citep{bunel2020lagrangian}.
However, neither approach scales elegantly with the size of the relaxation, which prevents current formal verification methods from being applied to deep state-of-the-art networks.
In other words, lower bound estimation forms a computational bottleneck for BaB. 
A natural question that arises is why do traditional algorithms fail? We argue that by their very nature, they ignore the rich inherent structure of lower bound estimation for the problem of verification.
Specifically, all lower bounds that one wishes to estimate across multiple subdomains of the same property, across multiple properties of the same network, and across multiple networks share the same form of variables, constraints and objectives. Furthermore, the coefficients of the objective and constraints are determined from network weights, which themselves are not random but are estimated using real data.
Traditional optimization algorithms are agnostic to this complex high-dimensional structure as it is not ``visible'' to human intelligence. 

The aforementioned arguments suggest that, in order to scale-up verification, we require computationally cheap methods that can exploit the inherent structure of the problem and the data.
To this end, we propose a novel machine learning framework that can be used for designing a branching strategy and for the estimation of lower bounds.
Our framework is both computationally efficient and effective, giving branching decisions that are of a similar quality to that of strong branching.
Moreover, we use the same framework to return tighter lower bounds more quickly than other optimization methods. Specifically, we make the following contributions: 


\begin{itemize}
     \item We use a graph neural network (GNN) to exploit the structure of the neural network we want to verify. The embedding vectors of the GNN are updated by a novel schedule, which is both computationally cheap and memory efficient. In detail, we mimic the forward and backward passes of the neural network to update the embedding vectors. In addition, the proposed GNN allows a customised schedule to update embedding vectors via shared parameters. That means, once training is done, the trained GNN model is applicable to various verification properties on different neural network structures.
     
     \item We train two GNNs: one for branching and another for bounding. We provide ways to generate training data cheaply but inclusive enough to represent problems at different stages of a BaB process for various verification properties. With the ability to exploit the neural network structure and a comprehensive training data set, our GNNs are easy to train and converge quickly.
     
     \item Our learnt GNNs also enjoy transferability both horizontally and vertically. Horizontally, although trained with easy properties, the learnt GNNs give similar performance on medium and difficult level properties. More importantly, vertically, given that all other parts of BaB algorithms remain the same, the GNNs trained on small networks perform well on large networks. Since the network size determines the total cost for generating training data and is positively correlated with the difficulty of learning, this vertical transferability allows our framework to be readily applicable to large scale problems. 


     \item Finally, we supply a dataset on convolutional neural network verification problems, covering problems at different difficulty levels over neural networks of different sizes. We hope that by providing a large problem dataset it could allow easy comparisons among existing methods and additionally encourage the development of better methods.
\end{itemize}
Since most available verification methods work on ReLU-based deep neural networks, we focus on neural networks with ReLU activation units in this paper. However, we point out that our framework is applicable to other neural network architectures using different non-linearities such as sigmoid or the hyperbolic tangent.

A preliminary version of this work appeared in the 2020 proceedings of the International Conference on Learning Representations, in which we focused on learning how to branch. We have extended the work to now also include learnt bounding algorithms. Moreover, we have merged the branching and the bounding methods to get a learnt complete verification algorithm that outperforms state-of-the-art solvers.

%% file: Sections/Background.tex
\section{Background}
Formal verification of neural networks refers to the problem of proving or disproving a property over a bounded input domain. Properties are functions of neural network outputs. When a property can be expressed as a Boolean expression over linear forms,  we can modify the neural network in a suitable way so that the property can be simplified to checking the sign of the neural network output~\citep{rudy2018}. Note that all the properties studied in previous works satisfy this form, thereby allowing us to use the aforementioned simplification. Mathematically, given the modified neural network $f$, a bounded convex input domain $\mathcal{C}$, formal verification examines the truthfulness of the following statement:
\begin{equation}\label{eq:verif-prop}
    \forall \x \in \mathcal{C}, \qquad f(\x)\geq 0.
\end{equation}
If the above statement is true, the property holds. Otherwise, the property does not hold. 

\subsection{Branch-and-Bound}
Verification tasks are often treated as a global optimization problem. We want to find the minimum of $f(\x)$ over $\mathcal{C}$ in order to compare it with the threshold $0$. Specifically, we consider an $L$ layer feed-forward neural network, $f: \mathbb{R}^{d} \rightarrow \mathbb{R}$, with non-linear activations $\sigma$ such that for any $\x_0\in \mathcal{C} \subseteq \mathbb{R}^{d}$, $f(\x_0)= \hat{\x}_{L}\in\mathbb{R}$ where
\begin{subequations}\label{eq:nn-formula}
\begin{align}
        \hat{\x}_{i+1} &= W^{i+1}\x_{i} + \bb^{i+1}, \qquad &&\text{for }i=0,\dots, L-1, \label{eq:linear} \\
        \x_{i} &= \sigma(\hat{\x}_i), \qquad &&\text{for }i=1,\dots, L-1. \label{eq:relu}
\end{align}

\end{subequations}
\noindent The terms $W^{i}$ and $\bb^{i}$ refer to the weights and biases of the $i$-th layer of the neural network f. 
Domain $\mathcal{C}$ can be an $\ell_p$ norm ball with radius $\epsilon$.
In our case, we use the ReLU activation defined as $\sigma(x) = \max(x, 0)$ as it is widely used in machine learning in general \citep{krizhevsky2012imagenet, maas2013rectifier} and in neural network verification in particular \citep{bunel2018unified, dvijotham2018training, ehlers2017formal}. However, we note that our approach works for other non-linearities such as the sigmoid activation or the hyperbolic tangent \citep{de2021improved}. See Appendix \ref{sec:app:non-linearities} for a more detailed description of using other non-linearities. 
 Finding the minimum of $f$ is a challenging task, as the optimization problem is generally NP hard~\citep{Katz2017}. To deal with the inherent difficulty of the optimization problem itself, BaB~\citep{rudy2018} is generally adopted. In detail, BaB based methods divide the feasible domain defined by equation (\ref{eq:linear}) and (\ref{eq:relu}) for all ${\bf x} \in {\cal C}$ into sub-domains, each of which defines a new sub-problem (branching). They then compute a relaxed lower bound of the minimum on each sub-problem (bounding). The minimum of the lower bounds of all the generated sub-domains constitutes a valid global lower bound of the global minimum over $\mathcal{C}$. As a recursive process, BaB keeps partitioning the sub-domains to tighten the global lower bound. The process terminates when the computed global lower bound is above zero (property is true) or when an input with a negative output is found (property is false). A detailed description of the BaB algorithm is provided in the appendices. In what follows, we provide a brief description of the two components, bounding methods and branching strategies, that is necessary for the understanding of our novel learning framework. 
  
\subsection{Bounding}


\begin{figure}[!tb]
\begin{subfigure}{0.32\textwidth}
 \centering
  \input{Figures/relu/convexN.tex}
  \caption{Naive relaxation}
\end{subfigure}
~
\begin{subfigure}{0.32\textwidth}
  \centering
  \input{Figures/relu/convexF.tex}
  \caption{Linear bounds relaxation}
\end{subfigure}
~
\begin{subfigure}{0.32\textwidth}%
  \centering
  \input{Figures/relu/convexkw.tex}
  \caption{Planet relaxation~\citep{Ehlers2017}}
\end{subfigure}
\caption{\footnotesize Different convex relaxations introduced. For each plot, the black line shows the output of a ReLU activation unit for any input value between $l_{i[j]}$ and $u_{i[j]}$ and the green shaded area shows the convex relaxation introduced. Naive relaxation (a) is the loosest relaxation. Linear bounds relaxation (b) is tighter and is introduced in~\citet{Weng2018}. Finally, Planet relaxation (c) is the tightest linear relaxation among the three considered~\citep{Ehlers2017}. Among them, (a) and (b) have closed form solutions which allow fast computations while (c) requires an iterative procedure to obtain an optimal solution.}
\label{fig:convex-relax}
\end{figure}
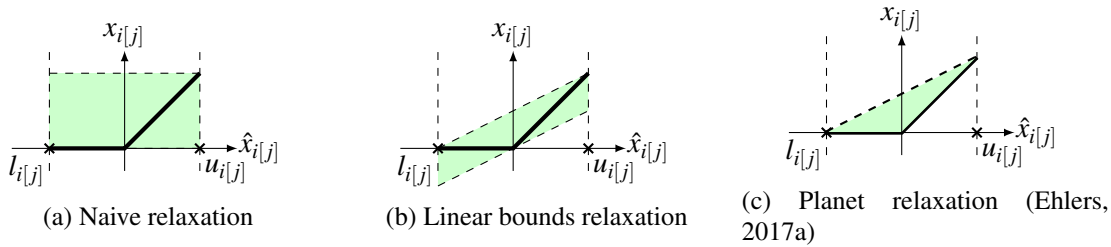

The bottleneck of most BaB algorithms is the estimation of a lower bound on the output of the neural network we are trying to verify on a given subdomain.
As the neural network is highly non-convex and thus hard to optimize over we use convex relaxations. Different linear-sized relaxations have been proposed in the literature such as naive relaxation, linear bounds relaxation, and Planet relaxation. In our work we use the Planet relaxation as it is the tightest relaxation as can be seen in Figure~\ref{fig:convex-relax}. Specifically, we focus on the decomposition based approach of \citet{bunel2020lagrangian} which is described in more detail below.

\paragraph{Planet Relaxation.} 
We denote the  output of the $i$-th layer before the application of the ReLU as $\zhatbold_i$ and the output of applying the ReLU to $\zhatbold_i$ as $\z_i$. Given the lower bounds $\lbk$ and upper bounds $\ubk$ of the values of $\zhatbold_i$, we relax the ReLU activations $\z_i = \sigma(\zhatbold_i)$ to its convex hull $\cvxhull$, defined as follows:
\begin{equation}
    \label{eq:cvx_hull}
    \cvxhull \equiv
    \begin{cases}
    x_{i[j]} \geq 0 \;\;\; x_{i[j]} \geq \zhat_{i[j]}\\
    x_{i[j]} \leq \frac{u_{i[j]}(\zhat_{i[j]} - l_{i[j]})}{u_{i[j]} - l_{i[j]}}  \quad &\iif l_{i[j]} < 0 \aand u_{i[j]} > 0\\
    x_{i[j]} = 0													   &\iif u_{i[j]} \leq 0 \\
    x_{i[j]} = \zhat_{i[j]}										&\iif l_{i[j]} \geq 0. \\
    \end{cases}
    \; ,\forall j
\end{equation}
Here, $x_{i[j]}$ denotes the $j$-th element of ${\bf x}_i$.
Note that the computation of the convex hull requires the knowledge of the lower and upper bounds (i.e. $\lb_i$ and $\ub_i$) for each intermediate node. The bounds do not have to be optimal. However, the tighter the bounds are, the tighter the relaxation will be as well. 
There are different ways of computing said bounds that have been proposed in the literature \citep{gowal2018effectiveness, raghunathan2018certified, Wong2018}. In our experiments we use the method proposed by \citet{Wong2018}.
For the sake of clarity, we introduce the following notations for the constraints corresponding to the input and the $i$-th layer respectively:

\noindent\begin{minipage}{.5\linewidth}
\begin{equation*}
    \mathcal{P}_0(\z_0, \zhatbold_{1}) \equiv 
    \begin{cases}
    \z_0 \in C \\
    \zhatbold_1 = W_1 \z_0 + \bb_1 \\
    \end{cases}
\end{equation*}
\end{minipage}%
\begin{minipage}{.5\linewidth}
\begin{equation}
    \label{eq:P_constrains}
    \mathcal{P}_i(\zhatbold_{i}, \zhatbold_{i+1}) \equiv 
    \begin{cases}
    \exists \z_i \text{ s.t. }\\
    \lbk \leq \zhatbold_i \leq \ubk \\
    \cvxhull \\
    \zhatbold_{i+1} = W_{i+1} \z_i + \bb_{i+1}.
    \end{cases}
\end{equation}
\end{minipage}
Using the above notation, the Planet relaxation for computing the lower bound can be written as:
\begin{equation}\label{eq:primal}
    \min_{\z, \zhatbold} \zhat_L \text{ s.t. } \mathcal{P}_0(\z_0, \zhatbold_{1}); \mathcal{P}_i(\zhatbold_{i}, \zhatbold_{i+1}) \forkone.
\end{equation}

\paragraph{Lagrangian Decomposition.} We often merely need approximations of the bounds rather than the precise values of them. We can therefore make use of the primal-dual formulation of the problem as every feasible solution to the dual problem provides a valid lower bound for the primal problem.
Following the work of \citet{bunel2020lagrangian} we will use the Lagrangian decomposition \citep{guignard1987lagrangean}.
To this end, we first create two copies $\zhatbold_{A, i}, \zhatbold_{B, i}$ of each variable $\zhatbold_{i}$:
\begin{equation}\label{primal}
    \begin{aligned}
    \min_{\z, \zhatbold} \zhatbold_{A, L} \text{ s.t. } &\PO; \Pk &&\forkone\\
    &\zhatbold_{A, i} = \zhatbold_{B, i} &&\forkone.
    \end{aligned}
\end{equation}
Next we obtain the dual by introducing Lagrange multipliers $\rrho$ corresponding to the equality constraints of the two copies of each variable:
\begin{equation}\label{eq:dual}
    \begin{aligned}
    q(\rrho) = &\min_{\z, \zhatbold} &&\zhatbold_{A, n} + \sum_{i = 1, \dots, n-1} \rrho_i^\top (\zhatbold_{B, i} - \zhatbold_{A, i})\\
    &\text{ s.t. } &&\PO; \;\Pk \,\forkone.
    \end{aligned}
\end{equation}
Problem (\ref{eq:dual}) is unconstrained with respect to $\rrho$. In other words, any possible $\rrho$ is a feasible solution and thus by duality provides a lower bound for the primal problem ($\ref{primal}$).
We therefore aim to maximize $q(\rrho)$ to get the tightest possible lower bound.
Given an assignment to the dual variables, \citet{bunel2020lagrangian} showed that the minimization over $\z^*_0, \, \zhatbold^*_{A}, \, \zhatbold^*_{B}$ can be done efficiently.
The supergradients can then be easily computed as $\nabla_{\rrho}(q) = \zhatbold^*_B - \zhatbold^*_A$.
This is used to come up with lower bounds via supergradient ascent (see appendix \ref{sec:app:supergradient_ascent} for a more detailed explanation). Unfortunately, supergradient ascent is known to be quite slow. We therefore take a different approach that learns to estimate a better ascent direction, thereby providing larger lower bounds more efficiently.

\subsection{Branching}
Branching is of equal importance as bounding in the BaB framework. Especially for large scale networks $f$, each branching step has a large number of putative choices. In these cases, the effectiveness of a branching strategy directly determines the possibility of verifying properties over these networks within a given time limit. On neural networks, two types of branching decisions are used: input domain split and hidden activation unit split. 

Assume we want to split a parent domain $\mathcal{D}$. Input domain split selects an input dimension and then makes a cut on the selected dimension while the rest of the dimensions remain the same. The common choice is to cut the selected dimension in half and the dimension to cut is decided by the branching strategy used. Available input domain split strategies are~\citet{rudy2018} and~\citet{royo2019fast}. The strategy of \citet{royo2019fast} is based on a sensitivity test of the LP on $\mathcal{D}$ while~\citet{rudy2018} use the formula provided in~\citet{Wong2018} to estimate final output bounds for sub-domains after splitting on each input dimension and selects the dimension that results in the highest output lower bound estimates.

In our setting, we refer to a ReLU activation unit $x_{i[j]} = \max(\hat{x}_{i[j]}, 0)$ as ambiguous over $\mathcal{D}$ if the upper bound $u_{i[j]}$ and the lower bound $l_{i[j]}$ for $\hat{x}_{i[j]}$ have different signs. Activation unit split chooses among ambiguous activation units and then divides the original problem into cases of different activation phases of the chosen activation unit. If a branching decision is made on $x_{i[j]}$, we divide the ambiguous case into two determinable cases: \mbox{$\{x_{i[j]}=0, l_{i[j]}\leq \hat{x}_{i[j]}\leq 0\}$} and \mbox{$\{x_{i[j]}=\hat{x}_{i[j]}, 0\leq \hat{x}_{i[j]} \leq u_{i[j]}\}$}. After the split, the originally introduced convex relaxation is removed, since the above sets are themselves convex. We expect large improvements on the output lower bounds of the newly generated sub-problems if a good branching decision is made. Apart from random selection, employed in~\citet{Ehlers2017} and~\citet{Katz2017}, available ReLU split heuristics are~\citet{Wang2018} and~\citet{journal2019}. \citet{Wang2018} compute scores based on gradient information to prioritise ambiguous ReLU nodes. \citet{journal2019} use scores to rank ReLU nodes that are computed with a formula developed on the estimation equations in~\citet{Wong2018}. We note that for both branching strategies, after the split, intermediate bounds are updated accordingly on each new sub-problem. For neural network verification problems, either domain split or ReLU split can be used at each branching step. When compared with each other, ReLU split is a more effective choice for large scale networks, as shown in~\citet{journal2019}. 


All the aforementioned existing branching strategies use hand-designed heuristics. In contrast, we propose a new framework for branching strategies by learning to imitate strong branching heuristics. This allows us to harness the effectiveness of strong branching strategies while retaining the computational efficiency.

%% file: Figures/relu/convexN.tex
\begin{tikzpicture}
  \tikzset{dummy/.style= {inner sep=0, outer sep=0}}
  \tikzset{cross/.style={cross out, draw,
      minimum size=3*(#1-\pgflinewidth),
      inner sep=0pt, outer sep=0pt,
      thick}}

 \fill [white!80!green] (0, 0) to (0,1) to (2, 1) to (2,0);
  \draw[-, ultra thick](0, 0) to (1, 0) to (2, 1);
  \draw[dashed](0, 1) to (2, 1);    
  \draw[dashed](0, 0) to (2, 0);

  \draw[dashed](0, -0.3) to (0, 1.3);
  \draw[dashed](2, -0.3) to (2, 1.3);

  
  
  \node[cross=2pt] at (0, 0) {};
  \node[dummy](lb-lab) at (-0.3, -0.3) {$l_{i[j]}$};
  \node[cross=2pt] at (2, 0) {};
  \node[dummy](ub-lab) at (2.35, -0.3) {$u_{i[j]}$};

  \draw[-latex](-0.5,0) to (2.5, 0);
  \node[dummy](x-label) at (2.8, 0) {$\hat{x}_{i[j]}$};
  \draw[-latex](1,-0.3) to (1, 1.3);
  \node[dummy](x-label) at (1, 1.5) {$x_{i[j]}$};
      
\end{tikzpicture}

%% file: Figures/relu/convexF.tex
\begin{tikzpicture}
  \tikzset{dummy/.style= {inner sep=0, outer sep=0}}
  \tikzset{cross/.style={cross out, draw,
      minimum size=3*(#1-\pgflinewidth),
      inner sep=0pt, outer sep=0pt,
      thick}}

 \fill [white!80!green] (0, 0) to (2,1) to (2, 0.5) to (0, -0.5);
  \draw[-, ultra thick](0, 0) to (1, 0) to (2, 1);
  \draw[dashed](0, 0) to (2, 1);    
  \draw[dashed](0, -0.5) to (2, 0.5);

  \draw[dashed](0, -0.3) to (0, 1.3);
  \draw[dashed](2, -0.3) to (2, 1.3);

  
  
  \node[cross=2pt] at (0, 0) {};
  \node[dummy](lb-lab) at (-0.3, -0.3) {$l_{i[j]}$};
  \node[cross=2pt] at (2, 0) {};
  \node[dummy](ub-lab) at (2.35, -0.3) {$u_{i[j]}$};

  \draw[-latex](-0.5,0) to (2.5, 0);
  \node[dummy](x-label) at (2.8, 0) {$\hat{x}_{i[j]}$};
  \draw[-latex](1,-0.3) to (1, 1.3);
  \node[dummy](x-label) at (1, 1.5) {$x_{i[j]}$};
      
\end{tikzpicture}

%% file: Figures/relu/convexkw.tex
\begin{tikzpicture}
  \tikzset{dummy/.style= {inner sep=0, outer sep=0}}
  \tikzset{cross/.style={cross out, draw,
      minimum size=3*(#1-\pgflinewidth),
      inner sep=0pt, outer sep=0pt,
      thick}}

  \draw[-, ultra thick](0, 0) to (1, 0) to (2, 1);
  \draw[dashed, ultra thick](0, 0) to (2, 1);    

  \draw[dashed](0, -0.3) to (0, 1.3);
  \draw[dashed](2, -0.3) to (2, 1.3);

  \draw[fill=white!80!green](0, 0) -- (1,0) -- (2, 1);

  \node[cross=2pt] at (0, 0) {};
  \node[dummy](lb-lab) at (-0.3, -0.3) {$l_{i[j]}$};
  \node[cross=2pt] at (2, 0) {};
  \node[dummy](ub-lab) at (2.35, -0.3) {$u_{i[j]}$};

  \draw[-latex](-0.5,0) to (2.5, 0);
  \node[dummy](x-label) at (2.8, 0) {$\hat{x}_{i[j]}$};
  \draw[-latex](1,-0.3) to (1, 1.3);
  \node[dummy](x-label) at (1, 1.5) {$x_{i[j]}$};
      
\end{tikzpicture}

%% file: Sections/GNN_Framework.tex
\section{GNN Framework}

The bounding part of the BaB algorithm aims to estimate the lower bound for the final layer of the neural network. The strong branching strategy, that our branching method is based on, estimates the final lower bound for the two subdomains that splitting on any given node would create. It then chooses the node which leads to the subdomains with the smallest lower bounds. 
Both the branching and the bounding methods therefore rely on an accurate estimation of the final lower bound.
Previously known lower bound computation techniques such as supergradient ascent and proximal maximization~\citep{bunel2020lagrangian} can be thought of as performing forward-backward style passes through the network to update the dual variables. However, the exact form of the passes is restricted to those suggested by standard optimization algorithms, which are agnostic to the special structure of neural lower bound computation.
This observation suggests using a GNN framework to parameterize the forward and backward passes, and estimate the parameters using a training data set so as to exploit the problem and data structure more successfully.

Having motivated the use of a GNN we begin with a brief overview of our overall framework that our branching and bounding approaches are based on. Moreover, we provide a detailed description of its specialized implementation, first for the branching and then for the bounding problem.

\subsection{GNNs for Verification}

\begin{figure}[t]

\begin{center}
\includegraphics[width=\textwidth]{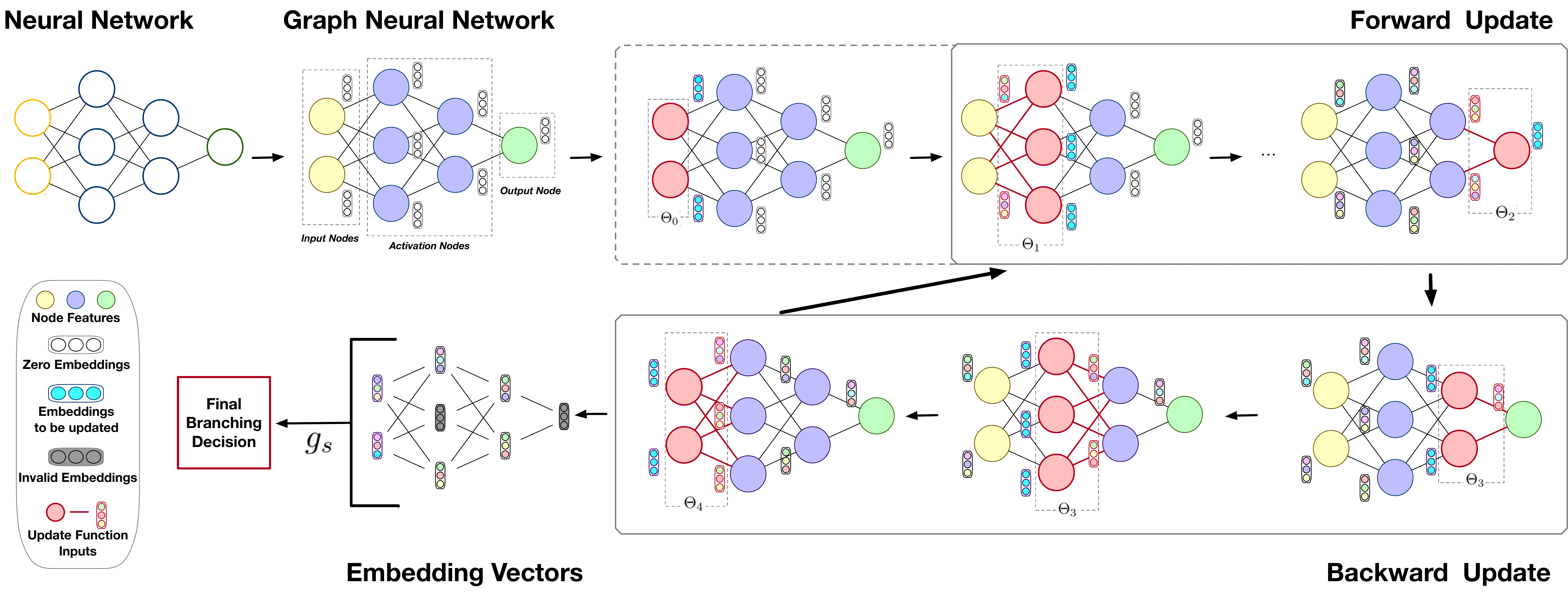}
\end{center}

\caption{\footnotesize Illustration of our proposed GNN framework. An all zeros embedding network mimicking the neural network is initialised. Embedding vectors are updated via several rounds of forward backward passes using updating Eqs.~(\ref{eq:f-inp})-(\ref{eq:b-inp}). We obtain the final branching decision by calling a score function $g_s$ over all embedding vectors of the potential branching decision nodes.}
\label{fig:mesage-passing}

\end{figure}

\paragraph{Nodes and Edges.}
 A graph neural network $\GGNN$ is represented by two components: a set of nodes $\VGNN$ and a set of edges $\EGNN$, such that $\GGNN=(\VGNN, \EGNN)$.
 The precise structure of $\GGNN$ depends on its input. We treat the neural network $f$ that we're trying to verify as a graph $G_f=(V_f, E_f)$ and provide it as input to the GNN. 
 The set of nodes in the GNN corresponds to the nodes in the original neural network. 
 Specifically, they are either based on the primal or on the dual variables.
 The set $\EGNN$ consists of all edges connecting nodes in $\VGNN$, which are exactly the connecting edges in $f$. Edges are characterized by the weight matrices that define the parameters of the network $f$ such that for an edge $e^{i}_{jk}$ connecting $v_{i[j]}$ and $v_{i+1[k]}$, we assign $e^{i}_{jk} = W^i_{jk}$.

\paragraph{Node Features.}
 For every node $v \in V_{GNN}$ we first compute a feature vector $\feat$, which contains local information about the node and will depend on the task that we want to solve. There is an inherent trade-off between using simple and easily computable features and using more informative but complex ones. Our methods use simple node features and do not rely on extensive feature engineering. Instead we rely on the powerful GNN framework - in particular the forward-backward passes described below - to generate accurate solutions. 
 
 \paragraph{Embeddings.}
 We associate a $p$-dimensional embedding vector $\mmu_{v}$ with each node $v \in \VGNN$. All embedding vectors are initialised either as zero vectors or using a learnt function that takes the corresponding feature vectors as input.
 
 \paragraph{Forward and Backward Passes.}
 In general, a graph neural network learns signals from a graph by acting as a function of two inputs: a feature matrix consisting of the embedding vectors, and an adjacency matrix representing the graph structure. Under this formulation, all node embedding vectors are updated at the same time and there is no particular order between nodes. In this work, instead, we propose an update scheme where only the nodes corresponding to the same layer of the network $f$ are updated at the same time
 . As described above, our approach is modelled on existing lower bound computation techniques:
 in order to pass information between the embedding vectors, we update them using a forward- and backward update schedule simulating runs through the original network. The update steps normally take as input the node feature vectors, the embedding vectors of all neighbouring nodes, as well as the edge values. At the end of the forward-backward updates, embedding vectors capture important information about the corresponding node, the structure of the neural network, and the state of the optimization algorithm.
 We point out that our forward-backward update scheme does not depend on the underlying neural network structure and thus should be applicable to network architectures that differ from the one we use for training. Furthermore, our forward-backward update is memory efficient, as we are dealing with one layer at a time and only the updated embedding vectors of the layer are used to update the embedding vectors in the next (forward-pass) and the previous (backward-pass) layer. This makes it readily applicable to large networks.
Specifically, during the forward update, for $i = 1,\dots, L-1$, we have, for all possible $j$,
\begin{align}
    \mmu_{0[j]} &\longleftarrow F_{inp}(\feat_{0[j]} ; \ttheta_0), \quad \text{if }\mmu_{0[j]}=\boldsymbol{0}, \label{eq:f-inp}
    \end{align}
    \vspace{-10pt}
  \noindent\begin{minipage}{0.5\textwidth}
  \vspace{-15pt}
    \begin{align}  
    \mmu_{i[j]} &\longleftarrow F_{act}(\fki, \mmu_{i-1}, \mmu_{i[j]}, e^{i}; \ttheta_1), \label{eq:f-act} 
    \end{align}
  \end{minipage}%
  \begin{minipage}{.5\textwidth}
  \vspace{-15pt}
    \begin{align}
    \mmu_{L} &\longleftarrow F_{out}(\feat_{L}, \mmu_{L-1}, e^{L} ; \ttheta_2). \label{eq:f-out}
    \end{align}
    \end{minipage}
    \vspace{5pt}\\
During the backward update, for $i=L-1, \dots, 1$, we have\\
\vspace{5pt}
\begin{minipage}{0.5\textwidth}
     \vspace{-2pt}
    \begin{align}
   \mmu_{i[j]} &\longleftarrow B_{act}(\fki, \mmu_{i+1}, \mmu_{i[j]}, e^{i+1} ; \ttheta_3), \label{eq:b-act}
    \end{align}
  \end{minipage}%
  \begin{minipage}{.5\textwidth}
       \vspace{-2pt}
       \begin{align}
       \mmu_{0[j]} &\longleftarrow B_{inp}(\feat_{0[j]}, \mmu_{1}, \mmu_{0[j]}, e^{1}; \ttheta_4). \label{eq:b-inp}
    \end{align}
  \end{minipage}
%
In other words, both the forward and the backward passes are performed by functions $F$ and $B$ with parameters $\Theta_0, \cdots, \Theta_4$, which can be estimated using a training data set. The structure of the functions $F$ and $B$ depends on the specific problem we are trying to solve.

\paragraph{Output Function.}
 Once the embedding vectors have been updated using the forward-backward passes they are gathered and treated as inputs to a learnt output function ${g}_{s}(\cdot; {\ttheta}_5): \mathbb{R}^{p} \rightarrow \mathbb{R}$. It transforms them into a solution to the problem we are aiming to solve: branching scores for every node, or a dual ascent direction. Our framework and the update steps are visualised in Figure \ref{fig:mesage-passing}.

 We will train two different GNNs, one for branching and one for bounding. They are both based on the framework described above, but the exact design of the feature vectors, the update steps, the output functions, and the training procedures differ. We will first describe the components of the branching GNN before outlining the similarities and differences with the GNN used for bounding.

\subsection{GNN for Branching}
In the previous section we described a general GNN framework. We now explain how to specialize it for the branching problem. The aim of the branching GNN is to output the next ReLU node to branch on.


\paragraph{Nodes.} The set of nodes $\VGNNhat$ is based on the primal variables including all input nodes $v_{0[j]}$, all hidden activation nodes $v_{i[j]}$ and the output node $v_{L}$. In our framework, we combine every pre-activation variable and its associated post-activation variable and treat them as a single node. Pre- and post-activation nodes together contain the information about the amount of convex relaxation introduced at this particular activation unit, so dealing with the combined node simplifies the learning process. In terms of Equation~(\ref{eq:nn-formula}), let $x'_{i[j]}$ denote the combined node of $\hat{x}_{i[j]}$ and $x_{i[j]}$. The nodes $v_{0[j]}$, $v_{i[j]}$ and $v_{L}$ are thus in one-to-one correspondence with $x_{0[j]}$, $x'_{i[j]}$ and $x_{L}$. We note that $\VGNNhat$ is larger than the set of all potential branching decisions as it includes unambiguous activation nodes and output nodes. 

\paragraph{Node Features.} 
Different types of nodes have different sets of features. In particular, input node features contain the corresponding domain lower and upper bounds and the primal solution. For activation nodes, the node features consist of associated intermediate lower and upper bounds, the layer bias, primal and dual solutions and new terms computed using previous features. Finally, the output node has features including the associated output lower and upper bounds, the layer bias and the primal solution. We denote input node features as $\widehat\feat_{0[j]}$, activation node features as $\widehat\feat_{i[j]}$ and output node features as $\widehat\feat_{L}$. 
All embedding vectors are initialised as zero vectors.

\paragraph{Forward and Backward Embedding Updates.}

The forward and backward update steps are as described in equations (\ref{eq:f-inp} - \ref{eq:b-inp}).
Update functions $\widehat{F}$ and $\widehat{B}$ take the form of multi-layered fully-connected networks with ReLU activation functions or composites of these simple update networks. The terms $\widehat{\ttheta}_i$ denote the parameters of the networks. A detailed description of update functions is provided in appendix \ref{app:branching_implementation}.

\paragraph{Scores.}
At the end of the forward-backward updates, embedding vectors for potential branching decision nodes (all input nodes and ambiguous activation nodes) are gathered and treated as inputs of a score function $\widehat{g}_{s}(\cdot; \widehat{\ttheta}_5): \mathbb{R}^{p} \rightarrow \mathbb{R}$, which takes the form of a fully-connected network with parameters $\widehat{\ttheta}_5$. It assigns a scalar score for each input embedding vector. The final branching decision is determined by picking the node with the largest score.

\subsection{GNN for Bounding}\label{sec:GNN_framework}

 The main structure of the GNN is similar to the branching GNN. The main difference being that once we have gotten a learnt representation of each node we will convert the embedding vectors into a dual ascent direction, which will be used to update the dual variables.

 \paragraph{Nodes.}
 We create a node $v_{i[j]}$ in our bounding GNN for every dual variable $\rho_{i[j]}$. Every dual variable corresponds to the output of the non-linear activation and the input to the next linear layer. 

 \paragraph{Node Features.}
 For each node $v_{i[j]}$ we define a corresponding $d$-dimensional feature vector $\featbarij \in \RR^d$ describing the current state of that node as follows: 
 \begin{equation}
    \featbarij := \left( \rho_{i[j]}, \zhat_{A, i[j]}, \zhat_{B, i[j]}, {\zhat_{B, i[j]} - \zhat_{A, i[j]}} \right)^{\top}.
 \end{equation}
 Here, $\rho_{i[j]}$ is the current assignment to the corresponding dual variable and $\zhat_{A, i[j]}$ and $\zhat_{B, i[j]}$ are the closed-form solutions to the inner minimization problem of the dual problem as explained above. The term ${\zhat_{B, i[j]} - \zhat_{A, i[j]}}$ corresponds to the supergradient of $q$.
 While more complex features could be included, we deliberately chose the simple features described above and rely on the power of GNNs to efficiently compute an accurate ascent direction.

\paragraph{Embeddings.}
For every node $v_i[j]$ we compute a corresponding $p$-dimensional embedding vector $\widebar \mmu_{i[j]} \in \RR^p$ using a learnt function $\widebar{g}_{init}: \mathbb{R}^{d} \rightarrow \mathbb{R}^p$ that takes the feature vector as input:
\begin{equation}
    \widebar \mmu_{i[j]} := \widebar{g}_{init} (\featbarij; \widebar{\ttheta}_0).
\end{equation}
In our case $\widebar{g}_{init}$ is a multilayer perceptron (MLP).

\paragraph{Forward and Backward Passes.}
The bounding GNN does not have any nodes corresponding to the input nodes of the original network. We thus don't have the update function $\widebar{F}_{inp}$. We also treat all forward passes the same, i.e. $\widebar{F} := \widebar{F}_{act} = \widebar{F}_{out}$. Similarly, we only have a single update function for all backward passes: $\widebar{B}$.
During the forward update, for $i=2, \dots, L$, we have
\begin{equation}
    \widebar \mmu_{i[j]} \longleftarrow \widebar{F}(\widebar \mmu_{i-1}, \widebar \mmu_{i[j]}, e^{i}; \widebar{\ttheta}_1), \label{eq:fbar-act} 
\end{equation}
and during the backward update, for $i=L-1, \dots, 1$, we have
\begin{equation}
    \widebar \mmu_{i[j]} \longleftarrow \widebar{B} (\widebar \mmu_{i+1}, \widebar \mmu_{i[j]}, e^{i+1}; \widebar{\ttheta}_2 ). \label{eq:fvar-act} 
\end{equation}

\paragraph{Update Step and Output Function.}
Finally, we need to reduce each $p$-dimensional embedding vector to a single value to get an ascent direction $\rrhobar_i^{t+1}$.
We use an output function $\widebar{g}_{out}(\cdot; \widebar{\ttheta}_3): \mathbb{R}^{p} \rightarrow \mathbb{R}$ with learnable parameter $\widebar{\Theta}$ to get a set of dual variables: $\rrhobar_i^{t+1} = \widebar{g}_{out}(\widebar \mmu_i;  \widebar{\Theta}_3).$
 Ideally the GNN would output a new ascent direction that will lead us directly to the global optimum of equation (\ref{eq:dual}). However, as the dual problem is complex this may not be feasible in practice without making the GNN very large, thereby resulting in computationally prohibitive inference. Instead, we propose to run the GNN a small number of times to return ascent directions that gradually move towards the optimum.
 Given a step size $\eta^{t+1}$, previous dual variables $\rrho^{t}$, and the new ascent direction $\rrhobar^{t+1}$ we update the dual variables as follows: 
 \begin{equation}
     \rrho^{t+1} = \rrho^{t} + \eta^{t+1} \rrhobar^{t+1}.
 \end{equation}
 Similar to many iterative optimization methods we decay our stepsize as we want to take smaller steps the closer we get to the optimal solution. Given an initial step size $\eta_0$, we define the step size at time $t$ as follows: $\eta^t = \eta_0 * \sqrt{t}$.
 %

%% file: Sections/GNN_Branching.tex
\section{Parameter Estimation for Branching}
Having described the structure of the GNNs we will now explain how to train them. Whereas the structure of the different GNNs was similar, the training procedure differs more strongly due to the difference in nature between the tasks they aim to solve. We will first outline how to estimate the learnable parameters for the branching GNN before doing the same for the bounding GNN.

\paragraph{Training.} We train a GNN via supervised learning. To estimate ${\widehat{\Theta}}\coloneqq (\widehat{\Theta}_0, \widehat{\Theta}_1, \widehat{\Theta}_2, \widehat{\Theta}_3, \widehat{\Theta}_4, \widehat{\Theta}_5)$, we propose a new hinge rank loss function that is specifically designed for our framework. Before we give details of the loss, we introduce a relative improvement measure $m$ first. Given a domain $\mathcal{D}$, for each branching decision node $v$, the two generated sub-problems have output lower bounds $l^{L}_{s_1}$ and $l^{L}_{s_2}$. We measure the relative improvement of splitting at the node $v$ over the output lower bound $l^{L}_{\mathcal{D}}$ as follows 
\begin{equation}\label{eq:improvement}
m_v \coloneqq (\min(l^{L}_{s_1},0) + \min(l^{L}_{s_2},0) -2\cdot l^{L}_{\mathcal{D}})/(-2\cdot l^{L}_{\mathcal{D}}).
\end{equation}
Intuitively, $m$ ($0\leq m \leq 1$) measures the average relative sub-problem lower bound improvement to the maximum improvement possible, that is $-l^{L}_{\mathcal{D}}$. Any potential branching decision node $v$ can be compared and ranked via its relative improvement value $m_v$. Since we are only interested in branching nodes with large improvement measures, ranking loss is a natural choice. A direct pairwise rank loss might be difficult to learn for NN verification problems, given the large number of branching decision nodes on each domain $\mathcal{D}$. In addition, many branching decisions may give similar performance, so it is redundant and potentially harmful to the learning process if we learn a ranking among these similar nodes. To deal with these issues, we develop our loss by first dividing all potential branching nodes into $M$ classes ($M$ is much smaller than the total number of branching decision nodes) through the improvement value $m_v$ of a node. We denote the class label as $Y_v$ for a node $v$. Labels are assigned in an ascending order such that $Y_v >= Y_{v'}$ if $m_v > m_{v'}$. We then compute the pairwise hinge-rank loss on these newly assigned labels as
\begin{equation}\label{eq:hinge-rank-loss}
\vspace{-5pt}
    loss_{\mathcal{D}} (\boldsymbol{\widehat{\Theta}}) = \frac{1}{K}\sum_{i=1}^{N}\big(\sum^{N}_{j=1}\phi(g_{s}(\boldsymbol{\mu}_j;
    \boldsymbol{\widehat{\Theta}})-g_{s}(\boldsymbol{\mu}_i;
    \boldsymbol{\widehat{\Theta}}))\cdot  \mathbf{1}_{\mathrm{Y_j>Y_i}}\big),
\vspace{-4pt}
\end{equation}
where $\phi(z) = (1-z)_{+}$ is the hinge function, $N$ is the total number of branching decision nodes and $K$ is the total number of pairs where $Y_j>Y_i$ for any branching decision nodes $v_i, v_j$. The loss measures the average hinge loss on score difference ($g_{s}(\boldsymbol{\mu}_j; \boldsymbol{\widehat{\Theta}})-g_{s}(\boldsymbol{\mu}_i; \boldsymbol{\widehat{\Theta}})$) for all pairs of branching decision nodes $v_i, v_j$ such that $Y_j >Y_i$. Finally, we evaluate $\boldsymbol{\widehat{\Theta}}$ by solving the following optimization problem:
\begin{equation}\label{eq:min-theta}
\vspace{-3pt}
    \boldsymbol{\widehat{\Theta}} = \arg \min_{\boldsymbol{\widehat{\Theta}}} \frac{\lambda}{2}\Vert\boldsymbol{\widehat{\Theta}} \Vert^2 + \frac{1}{n}\sum_{i}^n loss_{\mathcal{D}_i}(\boldsymbol{\widehat{\Theta}}),
\vspace{-3pt}
\end{equation}
where the $loss_{\mathcal{D}_i}$ is the one introduced in Equation~(\ref{eq:hinge-rank-loss}) and $n$ is the number of training samples. 


\paragraph{Fail-safe Strategy.}
We introduce a fail-safe strategy employed by our framework to ensure that consistent high-quality branching decisions are made throughout a BaB process. The proposed framework uses a GNN to imitate the behavior of the strong branching heuristic. Although computationally cheap, in some cases, the output decision by the learnt graph neural network might be suboptimal. When this happens, it could lead to considerably deteriorated performance for two reasons. Firstly, we observed that for certain problems, which requires multiple splits to reach a conclusion on this problem, if a few low-quality branching decisions are made at the beginning or the middle stage of the branching process, the total number of splits required might increase substantially. The total BaB path is thus, to some extent, sensitive to the quality of each branching decision apart from those made near the end of the BaB process. Secondly, once a low-quality decision is made on a given problem, a decision of similar quality is likely to be made on the two newly generated sub-problems, leading to exponential decrease in performance. Features for newly generated sub-problems are normally similar to those of the parent problem, especially in the cases where the branching decision of the parent problem is made on the later layers and loose intermediate bounds are used. Thus, it is reasonable to expect the GNN fails again on the resulting sub-problems.

To deal with this issue, we keep track of the output lower bound improvement for each branching decision, as introduced in Equation~(\ref{eq:improvement}). We then set a pre-determined threshold parameter. If the improvement is below the threshold, a computationally cheap heuristic is called to make a branching decision. Generally, the back-up heuristic is able to give an above-threshold improvement and generate sub-problems sufficiently different from the parent problem to allow the learnt GNN to recover from the next step onwards.

%% file: Sections/GNN_Bounding.tex
\section{Parameter Estimation for Bounding}
\paragraph{Training.}
 
We now describe how to train the learnable parameters for the bounding GNN.
We aim to maximize the dual values returned by the bounding GNN method.
The dual value directly corresponds to the final layer lower bound for a given subdomain. In particular, if the dual value is strictly positive, then the corresponding lower bound is greater than zero as well and we can prune away that subdomain thereby reducing the size of the BaB tree.

Recall that we do not use the GNN to directly compute the optimum dual solution. Instead, we run it iteratively, where each iteration computes an update direction for the dual variables.
In order for the training procedure to more closely resemble its behaviour at inference time, it is crucial to train the GNN using a loss function that takes into account the dual values across a large number of iterations $K$.
In order to ensure that a single training sample does not dominate the loss by reaching a large positive value, we truncate the loss values for each sample.
The natural point to clamp the individual losses at is the value returned by supergradient ascent ($q^i_{SupG}$) plus a small positive threshold $\kappa$. Inference time of our GNN is shorter than supergradient ascent as we run it for significantly fewer iterations (100 and 500 respectively); so as long as the duals returned by the GNN are as good as those returned by supergradient ascent, the GNN will outperform the baseline in the BaB setting.
%
  Given the $i$-th training sample $d_i$, the corresponding dual objective $q^i$, and the dual variables returned by the bounding GNN $\rrho_{GNN}^{i,t}$, we define its loss to be:
 \begin{equation}\label{loss_function}
     \loss_i = -\sum_{t=1}^K q^i(\rrho_{GNN}^{i,t})* \gamma^t
     *\mathbf{1}_{q^i(\rrho^{i,K}_{GNN}) < q^i_{SupG} + \kappa}.
 \end{equation} 
 Instead of maximizing over the dual value, we minimize over the negative dual instead.
 If the decay factor $\gamma \in (0,1)$ is low then we encourage the model to make as much progress in the first few steps as possible, whereas if $\gamma$ is closer to $1$, then more emphasis is placed on the final output of the GNN, sacrificing progress in the early stages.
 Readers familiar with reinforcement learning may be reminded by the discount rates used in algorithms such as Q-learning and policy-gradient methods.
 We sum over the individual loss values corresponding to each data point to get the final training objective $\loss$:
$
    \loss = 
    \sum_{i=1}^{\mid D \mid} \loss_i.
$

\paragraph{Fail-safe Strategy.}
 As our approach is based on learning, we do not have any convergence guarantees. For a few subdomains our GNN might diverge rather than improve on the value returned for its parent. We therefore introduce a fail-safe strategy that ensures our algorithm performs well even when our GNN fails.
 We compare whether the final bound of a given subdomain outputted by the GNN beats the bound returned for its parent domain by a given absolute threshold. If it fails to do so, then we add the subdomain into a second set of current subdomains. We use supergradient ascent to solve these subdomains on which our GNN performed poorly. This way we reduce the risk of our branch-and-bound algorithm timing out on certain properties.

\paragraph{Running Standard Algorithms using the GNN.}
As mentioned earlier, the motivation behind our GNN framework is to offer a parameterized generalization of previous methods for lower bound computation. We now formalize the generalization using the following proposition.
\begin{proposition}
    \label{prop:simulate_methods}
    Our GNN architecture can simulate supergradient ascent \citep{bunel2020lagrangian}
    (proof in appendix \ref{sec:app:proof_prop}).
\end{proposition}

%% file: Sections/Related_Work.tex
\section{Related Works}


Learning has already been used in solving combinatorial optimization problems~\citep{bello2016neural, Dai2017} and mixed integer linear programs (MILP)~\citep{ khalil2016learning, alvarez2017machine, hansknecht2018cuts, gasse2019exact}. In these areas, instances of the same underlying structure are solved multiple times with different data values, which opens the door for learning. Among them, \citet{khalil2016learning}, \citet{alvarez2017machine}, \citet{hansknecht2018cuts}, and~\citet{gasse2019exact} proposed learnt branching strategies for solving MILP with BaB algorithms. These methods imitate the strong branching strategy. Specifically, \citet{khalil2016learning} and~\citet{hansknecht2018cuts} learn a ranking function to rank potential branching decisions while~\citet{alvarez2017machine} uses regression to assign a branching score to each potential branching choice. Apart from imitation, \citet{Anderson2019} proposed utilizing Bayesian optimization to learn verification policies. There are two main issues with these methods. Firstly, they rely heavily on hand-designed features or priors and secondly, they use a generic learning structure which is unable to exploit the neural network architecture. 
The approach most relevant to ours is the concurrent work by~\citet{gasse2019exact}. They managed to reduce feature reliance by exploiting the bipartite structure of an MILP through a GNN. The bipartite graph is capable of capturing the network architecture, but cannot exploit it effectively. Specifically, it treats all the constraints the same and updates them simultaneously using the same set of parameters. This limited expressiveness can result in a difficulty in learning and hence in a high generalization error for NN verification problems. Our proposed framework is specifically designed for NN verification problems. By exploiting the neural network structure, and designing a customized schedule for embedding updates, our framework is able to scale elegantly both in terms of computation and memory. 

Most bounding methods create relaxations of the original problem. There are a variety of relaxations that have easily computable closed-form solutions such as Interval Bound Propagation \citep{gowal2018effectiveness} or WK, the method introduced by \citet{Wong2018}. However, these relaxations tend to be quite loose and therefore lead to bad estimates of the final layer output of a neural network. Hence different linear programming (LP) relaxations were proposed that provide tighter bounds: Planet introduced by \citet{ehlers2017formal}, Reluplex by \citet{katz2017reluplex}, or the Anderson relaxation \citep{anderson2019strong}. However, these often require an iterative solver to optimize them, which tends to not scale well.
 We will therefore use machine learning approaches to come up with better bounds than the best current iterative methods. 
 However, only limited work has been done on learning the bound computation in the BaB algorithm:
 \citet{dvijotham2018training, dvijotham2018dual} propose several different learnt methods: one that treats every layer separately, and another that uses a simple forward-backward architecture.
 \citet{gowal2019dual} propose a method called predictor-verifier  (PVT) that learns a robust training procedure and a verifier simultaneously.
 However, these methods end up beating Interval Bound propagation, a comparatively weak baseline, by a small margin only.
 There is thus a large scope for improvement which we explore in our work.

%% file: Sections/Experiments.tex
\section{Experiments}

We now validate the effectiveness of our proposed framework through comparative experiments against other available neural network verification methods. A comprehensive study of neural network verification methods has been done in~\citet{journal2019}. We thus design our experiments based on the results presented in~\citet{journal2019}. 

We start by describing our new verification dataset, which consists of properties of varying levels of difficulty defined on three different network architectures. We show that our branching GNN approach beats the best available hand-designed branching strategy. We then run experiments highlighting that the bounding GNN method outperforms conventional lower bounding techniques.
Finally, we combine our two GNNs to create a learnt branch-and-bound framework that beats the individual GNN methods as well as other state-of-the-art verification methods.
 
 \subsection{OVAL Verification Dataset}

 
Many existing datasets consist of a neural network along with a set of inputs and a constant epsilon value \citep{Balunovic2020Adversarial, katz2017reluplex, katz2019marabou}.
However, this results in a lot of properties being either SAT, meaning that a counter-example exists, or they are very easily verifiable. 
In both cases the property does not enable us to compare the lower bounding part of different verification methods in a meaningful way. 
In the former case the run-time and outcome are determined by the sub-routine that generates adversarial examples rather than the one that generates lower bounds, and in the latter case all verification methods verify the property very quickly.
To alleviate this inefficiency, we provide a dataset where the allowed input perturbation is uniquely determined for every image in the dataset. This ensures that all properties can be verified but doing so is challenging enough that it highlights the difference in performance between different methods --- we call it the OVAL verification dataset.

The OVAL benchmark consists of sets of adversarial robustness properties specifically designed for three adversarially trained CIFAR-10 convolutional neural networks with ReLU activations.
Two networks are composed of 2 convolutional layers followed by 2 fully connected layers: a ``\textit{Base}" model, and a wider ``\textit{Wide}" model. A  ``\textit{Deep}" model has 2 additional convolutional layers, with a width analogous to the ``\textit{Base}" model. All three models are trained robustly using the method introduced by \citet{Wong2018} to achieve robustness against $l_\infty$ perturbations of size up to $\epsilon$ = 8/255 (the amount typically considered in empirical works). We run our experiments on adversarially trained models, as standard trained networks are not robust for most images, and hence not suited as well for comparing verification methods.



Finally, we consider the following verification properties. Given an image $\mathbf{x}$ for which the model correctly predicted the label $y_c$, we randomly choose a label $y_{c'}$ such that for a given $\epsilon$, we want to prove
$(\ve^{(c)}-\ve^{(c')})^Tf'(\mathbf{x}') > 0, \, \forall \mathbf{x}' \text{ s.t } \Vert \mathbf{x}-\mathbf{x}' \Vert_{\infty} \leq \epsilon$.
Here, $f'$ is the original neural network, $\ve^{(c)}$ and $\ve^{(c')}$ are one-hot encoding vectors for labels $y_c$ and $y_{c'}$.
We want to verify that for a given $\epsilon$, the trained network will not make a mistake by labelling the image as $y_{c'}$.
Since BaBSR (Branch-and-Bound with Smart branching on ReLUs, \citep{journal2019}) is claimed to be the best performing method on convolutional networks, we use it to determine the $\epsilon$ values, which govern the difficulty level of verification properties.
Small $\epsilon$ values mean that most ReLU activation units are fixed so their associated verification properties are easy to prove while large $\epsilon$ values could lead to easy detection of counter-examples. The most challenging $\epsilon$ values are those at which a large number of activation units are ambiguous. We use binary search with BaBSR method to find the largest $\epsilon$ values possible that result in true properties. We further include a few timed out properties, that with a high probability are true properties to make the dataset even more challenging which might be beneficial when comparing stronger verification methods in the future. The binary search process is simplified by our choice of robustly trained models. Since these models are trained to be robust over a $\delta$-ball, the predetermined value $\delta$ can be used as a starting value for binary search.

Properties are generated for the ``\textit{Base}" model using binary search with BaBSR and a 3600s timeout. We categorise verification properties solved by BaBSR within 800s as easy, between 800s and 2400s as medium and more than 2400s as hard. In total, we generated 467 easy properties, 773 medium properties and 426 hard properties.
We use a larger timeout of 7200s to generate 300 properties for the ``\textit{Wide}" model and 250 properties for the ``\textit{Deep}" model.\\


\subsection{Training Dataset}
We now describe the training procedures for the branching and the bounding GNNs. Only the ``\textit{Base}" model is used for training the GNNs.
\paraorsub{Training Dataset for Branching.}
In order to generate training data, we firstly pick 565 random images and for each image, we randomly select an incorrect class. For each property, the $\epsilon$ value is determined by running binary search with BaBSR and 800 seconds timeout, so the final set of properties consists of mainly easily solvable properties and a limited number of timed out properties. This is consistent with the easy properties from the OVAL verification dataset.

We collect training data along a BaB process for solving a verification property. We would like to mimic strong branching as it is the most accurate branching strategy, however, as it very slow in practice, we have come up with the following regime to create a training dataset:
at each given domain, given the large number of potential branching decisions, we perform the strong branching heuristic on a selected subset of all potential branching decisions. The subset consists of branching decisions that are estimated to be of high quality by the BaBSR heuristic and randomly selected ones, which ensure a minimum $5\%$ coverage on each layer.

To construct a training dataset that is representative enough of the whole problem space, we need to cover a large number of properties. In addition, within a BaB framework, it is important to include training data at different stages of a BaB process. However, running a complete BaB process with the strong branching heuristic for hundreds of properties is computationally expensive and considerably time consuming. We thus propose the following procedure for generating a training dataset to guarantee a wide coverage both in terms of the verification properties and BaB stages. For generated verification properties, we randomly select $25\%$ of non-timeout property to conduct a complete BaB process with the strong branching heuristic. For the rest of the properties, we try to generate at least $\mathcal{B}=20$ training data for each verification property. Given the maximum number of branches $q=10$ and an effective and computationally cheap heuristic, we first generate a random integer $k$ from $[0,q]$. Then, we run a BaB process with the selected cheap heuristic for $k$ steps. Finally, we call the strong branching heuristic to generate a training sample. We repeat the process until $B$ training samples are generated or the BaB process terminated. A detailed algorithm is provided in appendix \ref{sec:app:training_branching}. 
Generating the training dataset takes less than 24 hours if we run it on 10 CPUs in parallel, and training the GNN takes about 6 hours. 

\paraorsub{Training Dataset for Bounding.}
We would like to train the bounding GNN on the same samples that we will encounter during inference time. However, that is impossible as the structure and the elements of the BaB tree computed at test time depend on the lower bound computation and thus on the GNN. 
To resolve that problem we dynamically create a training dataset as follows.
We first pick a fixed number of images from the training dataset used to train the branching GNN together with the corresponding properties that we are verifying our network against and epsilon values defining the input domain. We then create the first part of the training dataset by running a complete BaB algorithm on these properties using the supergradient method. We record the intermediate bounds and parent dual variables for each subdomain visited to create a dataset to train a first GNN on.
Once we have finished training the first version of the GNN we extend the dataset by running another complete BaB algorithm on the same properties; this time using the first version of the GNN instead of supergradient ascent to compute the lower bounds. We subsequently resume training the first GNN on the extended dataset for a fixed number of epochs to get a second GNN. We then repeat this process of extending the dataset and further training the GNN for a fixed number of iterations.
For most properties in the training dataset we acquire a large number of samples over the different iterations. To speed up training, we reduce the proportion of the training dataset on which we train our GNNs by only picking a small subset of the samples for each property. We make sure to pick subdomains from different stages of the BaB algorithm in order to get a more diverse training dataset.
We train the GNN using the loss function described in the previous section. A more detailed description can be found in Appendix \ref{sec:app:training_bounding}. The total training time for the bounding GNN, including the generation of the training dataset is 72 hours. Similar to the branching GNN there is an additional cost associated with our approach, that is training a GNN on a similar structure that we wish to verify. However, we argue that the significantly improved performance compared to state-of-the-art attack methods shown in various experiments below far outweighs this. In most use cases we want to repeatedly verify a network, or run lower bounding methods as a subroutine to calculate robustness. The more often we run our method, the more the improved performance of the GNN that is not just quicker but also more powerful than other verification methods makes up for the one-time cost.



\subsection{GNN Branching Only}
\import{Tables/}{branching_all.tex}
We now demonstrate the practical performance of the branching GNN method. 
We compute intermediate bounds using linear bounds relaxations (Figure~\ref{fig:convex-relax}(b)). For the output lower bound, we use Planet relaxation (Figure~\ref{fig:convex-relax}(c)) and solve the corresponding LP with Gurobi. For the output upper bound, we compute it by directly evaluating the network value at the input provided by the LP solution. 
In terms of our branching strategy we focus on ReLU split only in our experiments. As shown in~\citet{journal2019}, domain split only outperforms ReLU split on low input dimensional and small scale networks. Also, since one of the Baseline method BaBSR employs a ReLU-split heuristic, we consider ReLU split only for a fair comparison. However, we emphasize that our framework is readily applicable to work with a combined domain and ReLU split strategy.
 
Our method uses the branching GNN to return branching decisions.
We compare against the following two baselines: (i) MIPplanet, a mixed integer solver backed by the commercial solver Gurobi \citep{bunel2018unified}; (ii) BaBSR, a BaB based method utilising a ReLU-split heuristic and using the same Gurobi based LP solver as the GNN Branching method \citep{journal2019}. 
The branching GNN method and BaBSR only differ in the branching strategy used making it a fair comparison.
The two baselines using Gurobi are run on one CPU each.
The implementation of our method is based on Pytorch \citep{paszke2017automatic}.

Results for the branching GNN are given in Table~\ref{table:branching_all}. Methods are compared in three perspectives: the average time over all properties, average number of branches required over the properties that are solved by all methods (we exclude timed out properties) and also the percentage of timed out properties. BaBSR and the branching GNN only differ in the branching strategy used. 

\begin{figure}[h]

\centering
\begin{subfigure}{0.31\textwidth}
  \includegraphics[width=0.9\linewidth]{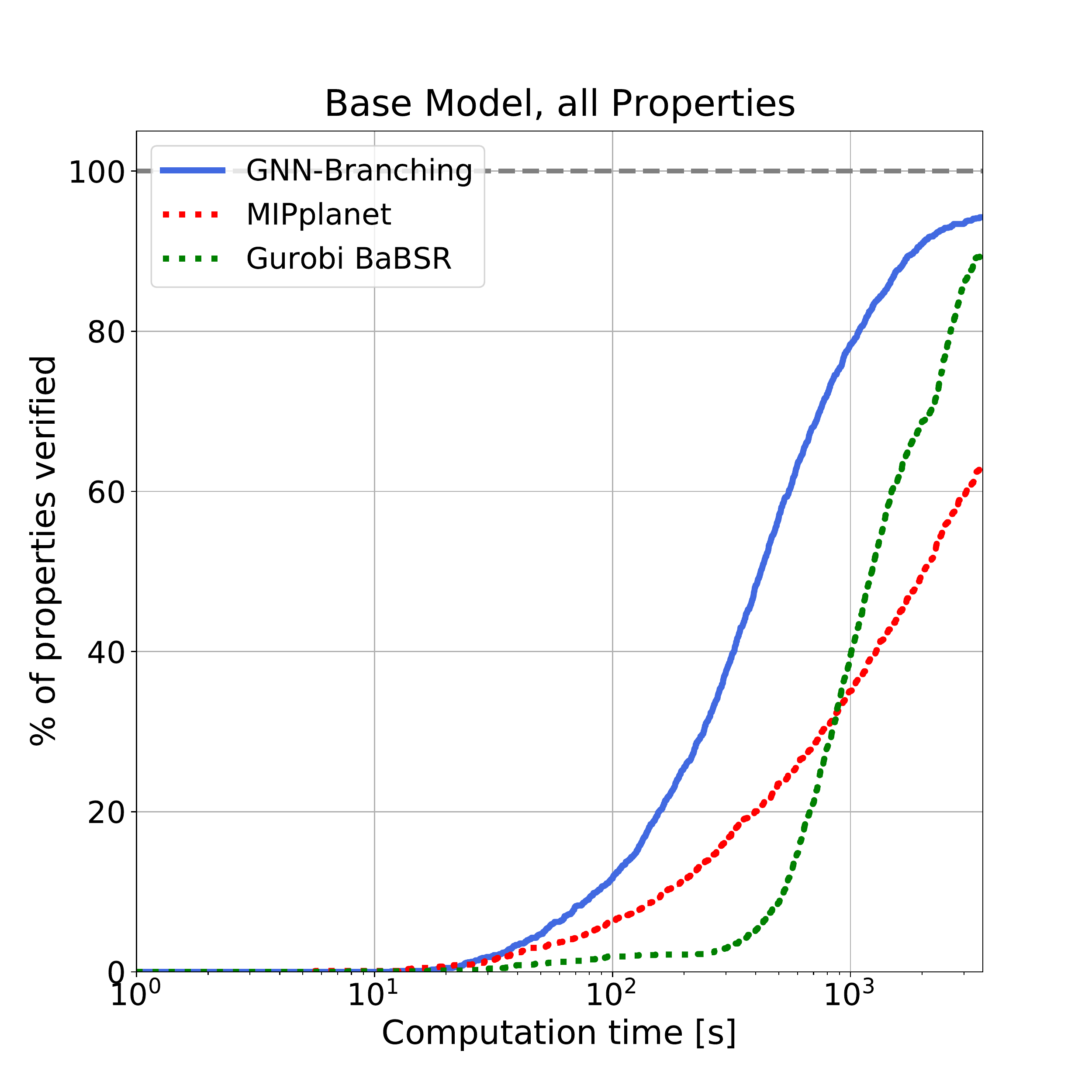}
  \vspace{-5pt}
  \caption{Base model}\label{fig:m2_total}
  \vspace{-10pt}
\end{subfigure}
\begin{subfigure}{0.31\textwidth}
  \includegraphics[width=0.9\linewidth]{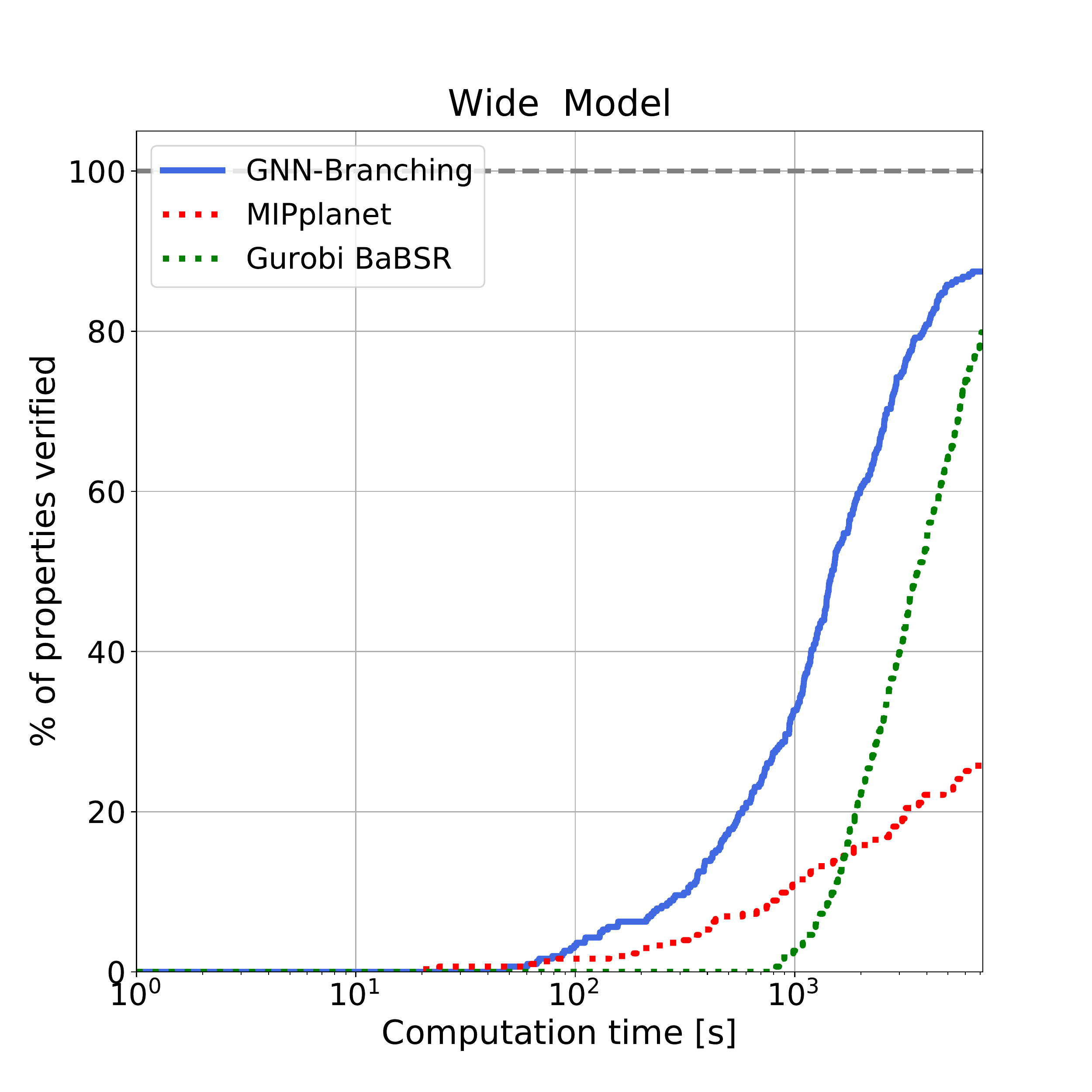}
  \vspace{-5pt}
  \caption{Wide large model}\label{fig:wide}
  \vspace{-10pt}
\end{subfigure}
\begin{subfigure}{0.31\textwidth}%
  \includegraphics[width=0.9\linewidth]{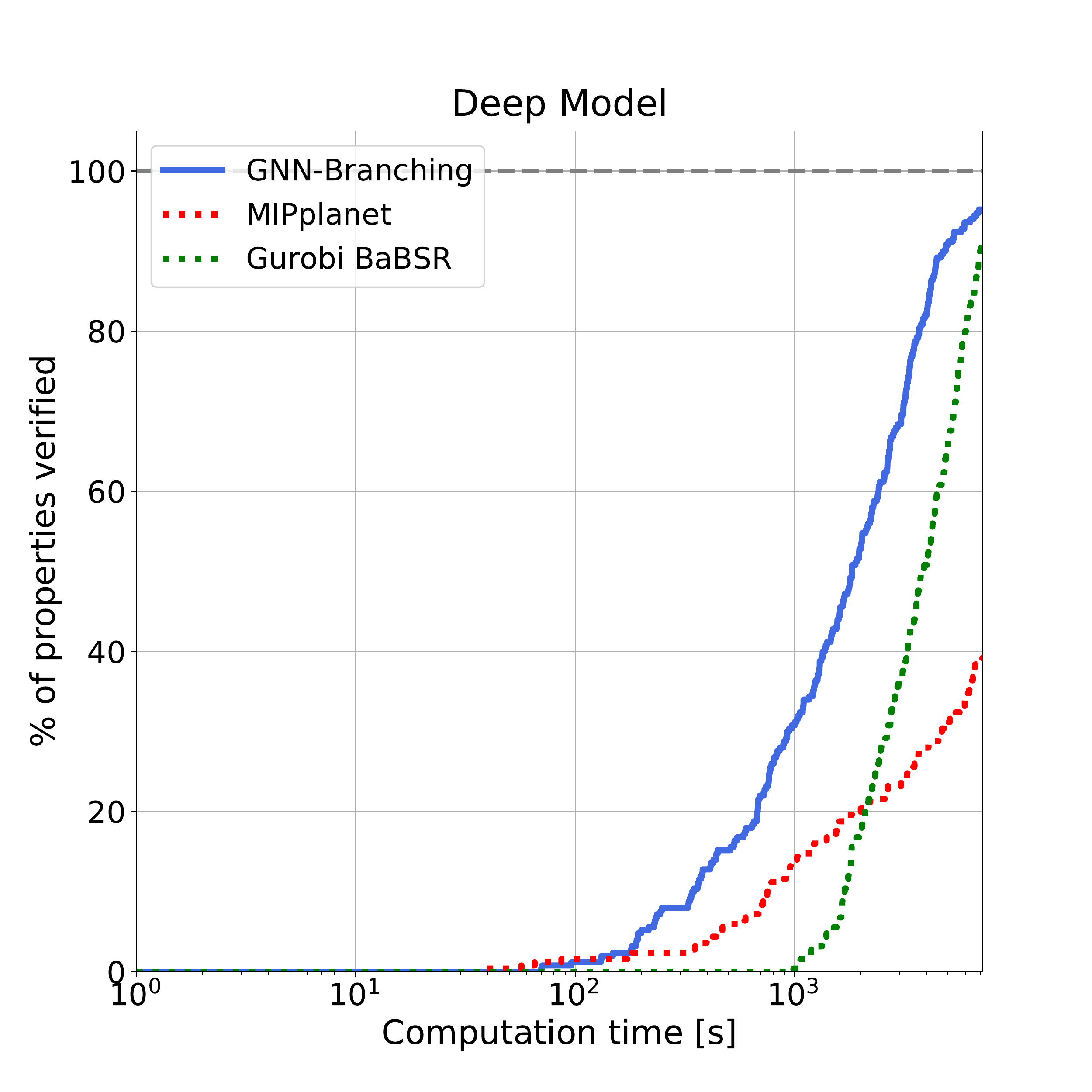}
  \vspace{-5pt}
  \caption{Deep large model}\label{fig:deep}
   \vspace{-10pt}
\end{subfigure}
\caption{Cactus plots for the ``\textit{Base}" model (left), ``\textit{Wide}" large model (middle) and ``\textit{Deep}" large model (right). For each model, we plot the percentage of properties solved in terms of time for each method. Consistent performances are observed on all three models. BaBSR beats MIPplanet on the majority of properties. GNN consistently outperforms BaBSR and MIPplanet. Baselines are represented by dotted lines.}
\vspace{-10pt}
\end{figure}

On all three sets of properties, we see that our learned branching strategy has led to a more than $50\%$ reduction in the total average number of branches required for a property. As a direct result, the average time required achieves at least a $50\%$ reduction as well.
%
We also provide a time cactus plot (Figure \ref{fig:m2_total}) for all properties on the ``\textit{Base}" model. Time cactus plots for each category of properties can be found in the appendices. All these time cactus plots look similar. 
The branching GNN method is capable of giving consistent high quality performance over all properties tested. 


 For the two larger models, each LP called for solving a sub-problem output lower bound is much more time consuming, especially for the ``\textit{Deep}" model. This is reason that the average number of branches considered is much fewer that those of the ``\textit{Base}" model within the given time limit.
The model learned on the ``\textit{Base}" network is tested on verification properties of large networks. Experimental results are given in the Table~\ref{table:branching_all} and time cactus plots (Figures \ref{fig:wide}, \ref{fig:deep}) are also provided. All results are similar to what we observed on the ``\textit{Base}" model, which show that our framework enjoys vertical transferability.


\subsection{GNN Bounding Only}

We will now show the practical effectiveness of our bounding GNN model by comparing its performance with several state-of-the-art bounding methods.
The bounding GNN method uses the hand-designed branching heuristic of BaBSR and the bounding GNN to return final lower bounds.
 Similar to the branching GNN experiments we compare to the Gurobi based baselines MIPplanet and BaBSR. We also compare against supergradient ascent together with Adam as proposed by \citet{bunel2020lagrangian} (for a more detailed description see appendix \ref{sec:app:supergradient_ascent}). 
 We run supergradient ascent for 500 steps with a learning rate of 1e-4; both of these hyper-parameters have been optimized over the validation dataset.
Note that whereas BaBSR aims to solve the subdomains to optimality, the supergradient and bounding GNN methods aim to trade accuracy for speed. As they return estimated bounds faster they lead to a quicker BaB algorithm despite generating more subdomains. 
 
 To speed up the BaB algorithm we parallelize over the lower bound computation for the different subdomains. A detailed description of the parallelized BaB algorithm can be found in Appendix \ref{sec:app:bab}.
 We run the bounding GNN with a batch-size of 200 and the supergradient ascent method with the highest batch-size possible for all experiments: 1600 on the ``\textit{Base}" model, 900 on the ``\textit{Deep}", and 350 on the ``\textit{Wide}" model. For the baseline, the maximum possible batch-size for a given memory constraint depends on the size of the model we are trying to verify, and the computation of the intermediate bounds, as well as the gradient. The actual update step is more memory efficient and thus does not influence the batch-size. In case of the bounding GNN, if the embedding size $p$ is small, then the maximum embedding size is the same as for supergradient ascent. For larger embedding sizes, the maximum embedding size is significantly lower.
 
 We run our GNN for 100 steps with an absolute fail-safe threshold of 0.05 for the ``\textit{Base}" experiment and a threshold of 0.1 for the larger models.
 We run both the supergradient baseline and the bounding GNN method on a single GPU and 4 CPUs each. 
One advantage of our combined method compared to off-the-shelf solvers is precisely that we can run it on GPUs and can therefore use more efficient parallelized implementations of mathematical operations.

 Our method leads to an over 70\% reduction in average time taken compared to BaBSR and MIPplanet (Table \ref{table:base_wide_deep}) and it times out on significantly fewer properties.
 Our bounding GNN also outperforms supergradient ascent.
 It leads to an over 20\% reduction in both time taken and number of subdomains visited.
 Even though the GNN is trained on easy properties only it generalizes well to harder ones (see appendix \ref{sec:app:easy_med_hard_exp}).

 We show further the generalization performance of our GNN without the need to perform fine-tuning or online learning by testing it on the two larger neural networks. 
 As shown in Figure \ref{fig:verification-main} and 
 Table \ref{table:base_wide_deep},
 the GNN still outperforms all baselines on the unseen networks both in terms of average time taken and the percentage of properties that time out.

 \import{Figures/}{cactus_bounding.tex}
\import{Tables/}{table_base_wide_deep.tex}

\subsection{Combined GNN Approach}
Having shown that the branching GNN outperforms the branching heuristic and the bounding GNN beats supergradient ascent for the estimation of the lower bounds we now combine the two GNNs. We use the branching GNN as the branching strategy and the bounding GNN for the computation of the final lower bounds.
We train a new branching GNN using the bounding GNN during the training procedure instead of the Gurobi based LP solver.
We run the combined GNN method on the same number of GPUs and CPUs as the bounding GNN and the supergradient ascent baseline.

As shown in Table \ref{table:combined_base_exp_rebuttal} the combined GNN method outperforms both the individual GNN methods as well as all other baselines on the ``\textit{Base}" model.
Moreover, it achieves good horizontal transferability as it was trained on easy properties only. In fact it performs particularly well on the medium and hard properties. It reduces the number of properties timing out by over 65\% compared to the best performing baseline.

Furthermore, the combined GNN also performs well on the unseen ``\textit{Wide}" and ``\textit{Deep}" models as shown in Table \ref{table:combined_base_wide_deep_rebuttal} and Figure \ref{fig:cactus:combined}. It times out on significantly fewer properties and reduces the average solving time by over 50\%.
Good generalisation performance from easy properties to difficult ones and from small networks to larger ones is beneficial as the complexity of training the GNNs depends on both the difficulty of the training properties and the size of the model. Moreover, it allows us to train a single pair of GNNs and use them for different verification tasks on various different networks.

More experimental results can be found in the appendices, including a more detailed analysis of our two GNNs (Appendices \ref{sec:app:further_experiments_branching} and \ref{sec:app:further_experiments_bounding}), a comparison of our models with \textbf{ERAN} \citep{singh2020eran}, another state-of-the-art complete verification method (Appendix \ref{sec:app:further_experiments_combined}), and experiments on another CIFAR10 dataset that uses fixed constant epsilon values (Appendix \ref{sec:app:costant_epsilons}).

\import{Figures/}{cactus_combined.tex}

\import{Tables/}{table_all_base_rebuttal.tex}
\import{Tables/}{table_all_rebuttal.tex}

%% file: Tables/branching_all.tex
\sisetup{detect-weight=true,detect-inline-weight=math,detect-mode=true}
\begin{table}[t]
\centering
\caption{\footnotesize Methods' Performance on all three models. We compare methods' average solving time, average number of branches required and the percentage of timed out properties. We use a timeout of 3600s for the ``\textit{Base}" model and a timeout of 7200s for the ``\textit{Wide}" and the ``\textit{Deep}" models. GNN-branching outperforms both baselines in all aspects.}
\label{table:branching_all}
\vspace{-10pt}
\scriptsize
\setlength{\tabcolsep}{4pt}
\aboverulesep = 0.1mm  
\belowrulesep = 0.2mm  
	\begin{adjustbox}{center}
		\begin{tabular}{
				l
				S[table-format=4.3]
				S[table-format=4.3]
				S[table-format=4.3]
				S[table-format=4.3]
				S[table-format=4.3]
				S[table-format=4.3]
				S[table-format=4.3]
				S[table-format=4.3]
				S[table-format=4.3]
			}
			& \multicolumn{3}{ c }{Base} & \multicolumn{3}{ c }{Wide} & \multicolumn{3}{ c }{Deep} \\
			\toprule
			
			\multicolumn{1}{ c }{Method} &
			\multicolumn{1}{ c }{time(s)} &
			\multicolumn{1}{ c }{subdomains} &
			\multicolumn{1}{ c }{$\%$Timeout} &
			\multicolumn{1}{ c }{time(s)} &
			\multicolumn{1}{ c }{subdomains} &
			\multicolumn{1}{ c }{$\%$Timeout} &
			\multicolumn{1}{ c }{time(s)} &
			\multicolumn{1}{ c }{subdomains} &
			\multicolumn{1}{ c }{$\%$Timeout} \\
			
			\cmidrule(lr){1-1} \cmidrule(lr){2-4} \cmidrule(lr){5-7} \cmidrule(lr){8-10}

    \multicolumn{1}{ c }{\textsc{BaBSR}}
		&1588.02
		&1340.08
		&10.57

        &4137.467
        &843.476
        &20.13

        &4016.336
        &416.824
        &9.20  \\

    \multicolumn{1}{ c }{\textsc{MIPplanet}}
		&2036.65
		& 
		&36.40
			
        &5855.059
        & 
        &74.30

        &5426.160
        & 
        &60.80 \\

    \multicolumn{1}{ c }{\textsc{GNN-branching}}
		&\B 752.94
		&\B 604.16
		&\B 5.77
			
        &\B 2367.693
        &\B 387.403
        &\B 12.70

        &\B 2308.612
        &\B 208.760
        &\B 4.80 \\

    \bottomrule

		\end{tabular}
	\end{adjustbox}

\end{table}

%% file: Figures/cactus_bounding.tex
 \begin{figure*}[h]
	\centering
	\begin{subfigure}{.32\textwidth}
		\centering
		\includegraphics[width=\textwidth]{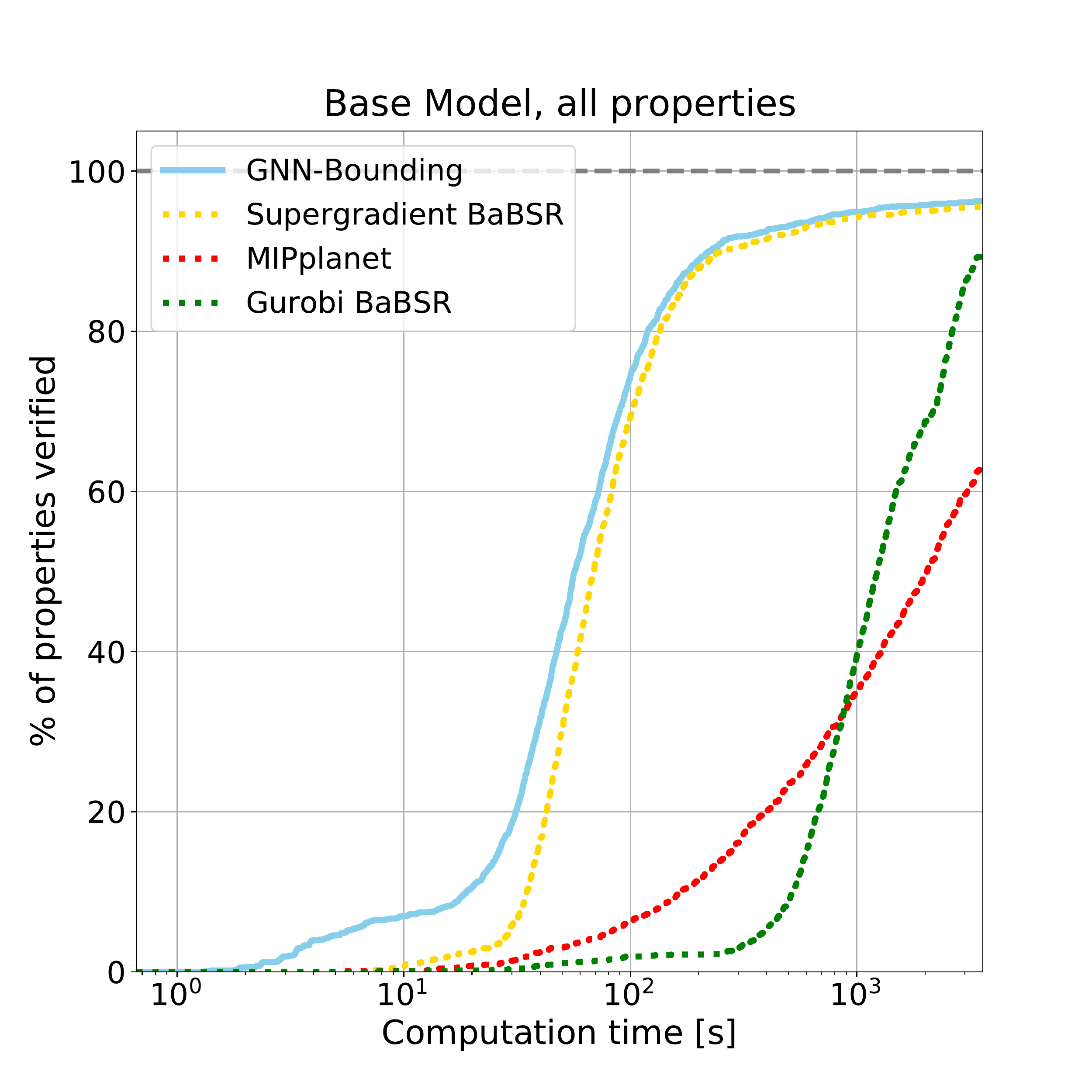}
	\end{subfigure}
	\begin{subfigure}{.32\textwidth}
		\centering
		\includegraphics[width=\textwidth]{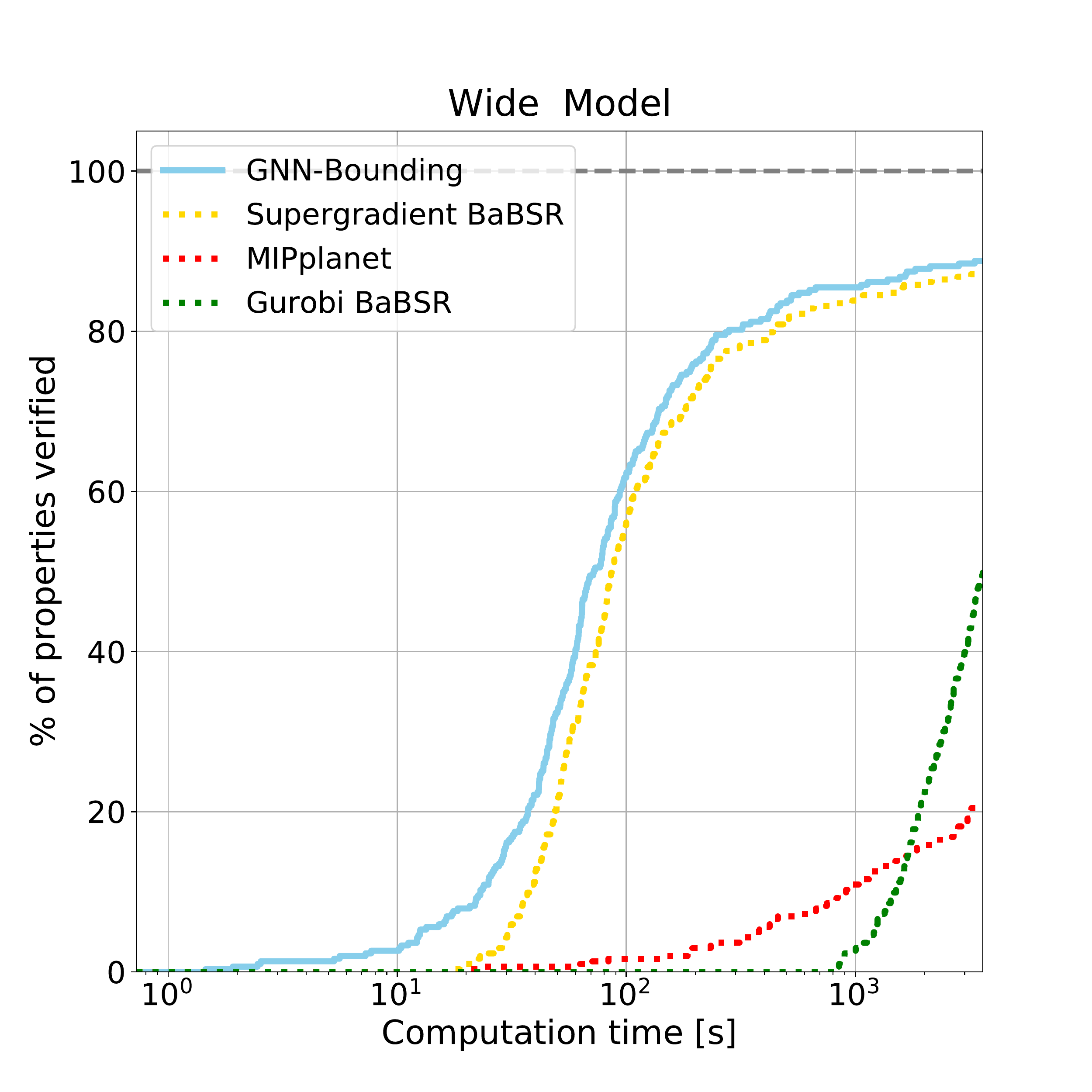}
	\end{subfigure}
	\begin{subfigure}{.32\textwidth}
		\centering
		\includegraphics[width=\textwidth]{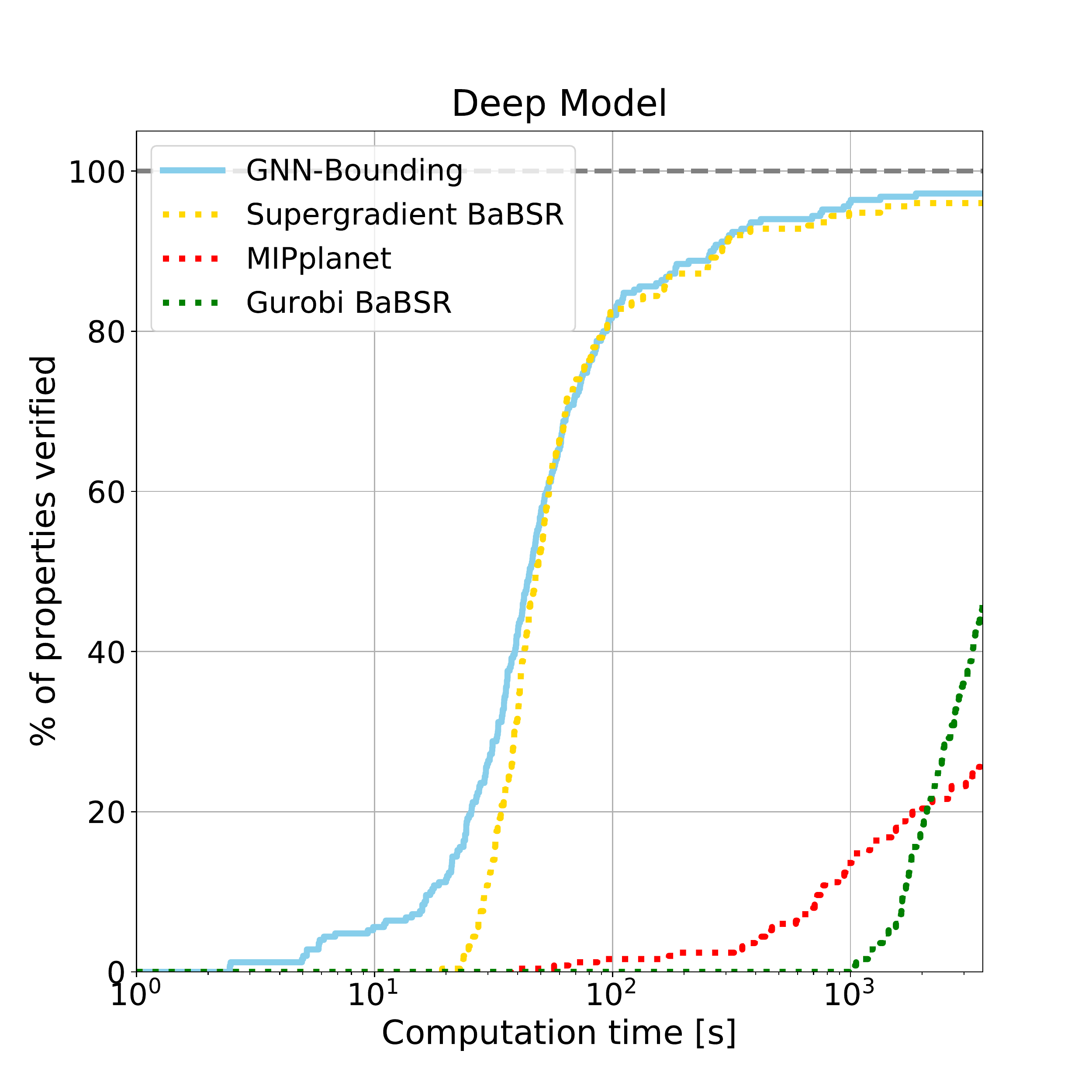}
	\end{subfigure}
	\caption{Cactus plots for the base, wide and deep models. For each, we compare the bounding methods and complete verification algorithms by plotting the percentage of solved properties as a function of runtime. Baselines are represented by dotted lines.}
	\label{fig:verification-main}
\end{figure*}

%% file: Tables/table_base_wide_deep.tex
 \sisetup{detect-weight=true,detect-inline-weight=math,detect-mode=true}
\begin{table*}[t]
	\centering
	\scriptsize
	\setlength{\tabcolsep}{4pt}
	\aboverulesep = 0.1mm  
	\belowrulesep = 0.2mm  
	
    
	\begin{adjustbox}{center}
		\begin{tabular}{
				l
				S[table-format=4.3]
				S[table-format=4.3]
				S[table-format=4.3]
				S[table-format=4.3]
				S[table-format=4.3]
				S[table-format=4.3]
				S[table-format=4.3]
				S[table-format=4.3]
				S[table-format=4.3]
			}
			& \multicolumn{3}{ c }{Base} & \multicolumn{3}{ c }{Wide} & \multicolumn{3}{ c }{Deep} \\
			\toprule
			
			\multicolumn{1}{ c }{Method} &
			\multicolumn{1}{ c }{time(s)} &
			\multicolumn{1}{ c }{subdomains} &
			\multicolumn{1}{ c }{$\%$Timeout} &
			\multicolumn{1}{ c }{time(s)} &
			\multicolumn{1}{ c }{subdomains} &
			\multicolumn{1}{ c }{$\%$Timeout} &
			\multicolumn{1}{ c }{time(s)} &
			\multicolumn{1}{ c }{subdomains} &
			\multicolumn{1}{ c }{$\%$Timeout} \\
			
			\cmidrule(lr){1-1} \cmidrule(lr){2-4} \cmidrule(lr){5-7} \cmidrule(lr){8-10}
			
			
			
			
			
			
			
			
			
			
			
			
		
			\multicolumn{1}{ c }{\textsc{Gurobi BaBSR}}
			&1588.02
			&1340.08
			&10.57
			
			&2917.95
			&855.37
			&51.11
			
			&3007.24
			&428.52
			&54.00 \\
			
			\multicolumn{1}{ c }{\textsc{MIPplanet}}
			& 2036.65
			& 
			&36.40
			
			& 3108.50
			& 
			&79.37
			
			& 2997.12
			& 
			&73.60 \\
			
			\multicolumn{1}{ c }{\textsc{Supergradient}}
			
			
			
            &277.22
            &8654.03
            &4.50

            &624.36
            &8885.64
            &12.87

            &241.80
            &3313.73
            &4.00\\

			\multicolumn{1}{ c }{\textsc{GNN-bounding}}
			&\B 219.62
			& 6427.10
			&\B 2.94
			
			&\B 513.63
			& 5274.19
			&\B 10.23
			
			& \B 214.55
			& 2784.49
			& \B 2.80\\

			\bottomrule
			
		\end{tabular}
	\end{adjustbox}
	\caption{\small We compare average (mean) solving time, average number of subdomains solved, and the percentage of properties that the methods time out on when using a cut-off time of 3600s. The best performing method for each subcategory is highlighted in bold.}
	\label{table:base_wide_deep}
\end{table*}

%% file: Figures/cactus_combined.tex
 \begin{figure*}[h]
	\centering
	\begin{subfigure}{.32\textwidth}
		\centering
		\includegraphics[width=\textwidth]{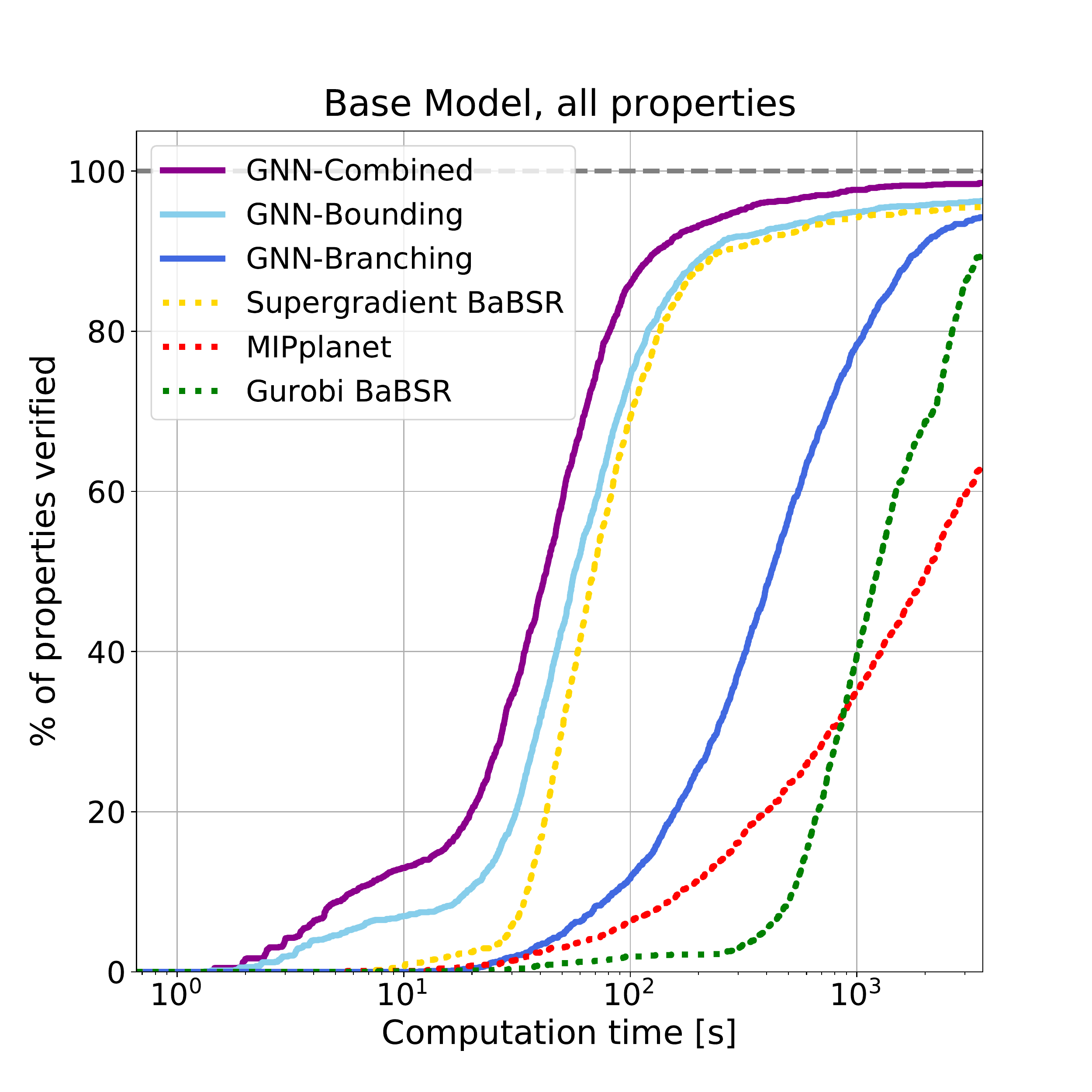}
	\end{subfigure}
	\begin{subfigure}{.32\textwidth}
		\centering
		\includegraphics[width=\textwidth]{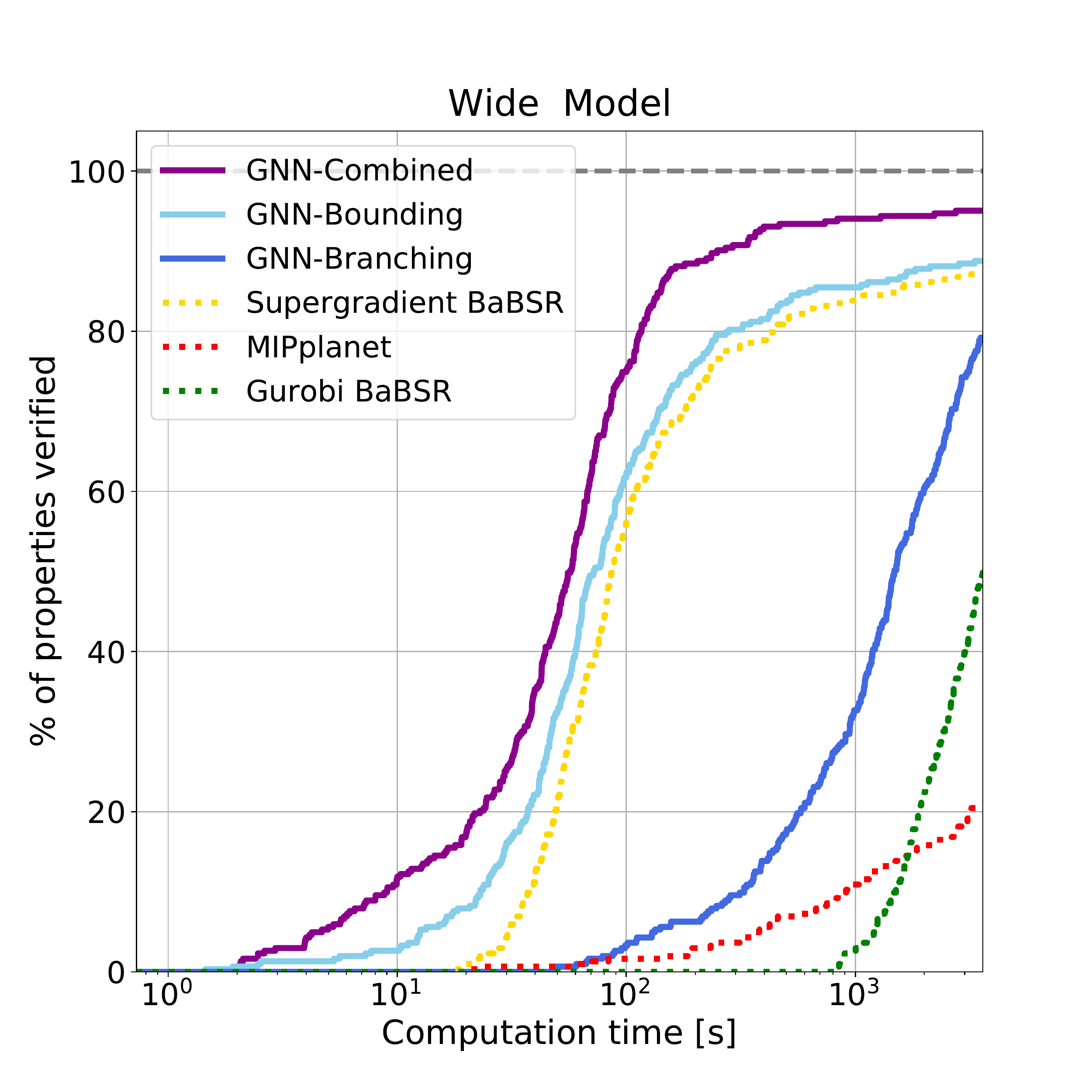}
	\end{subfigure}
	\begin{subfigure}{.32\textwidth}
		\centering
		\includegraphics[width=\textwidth]{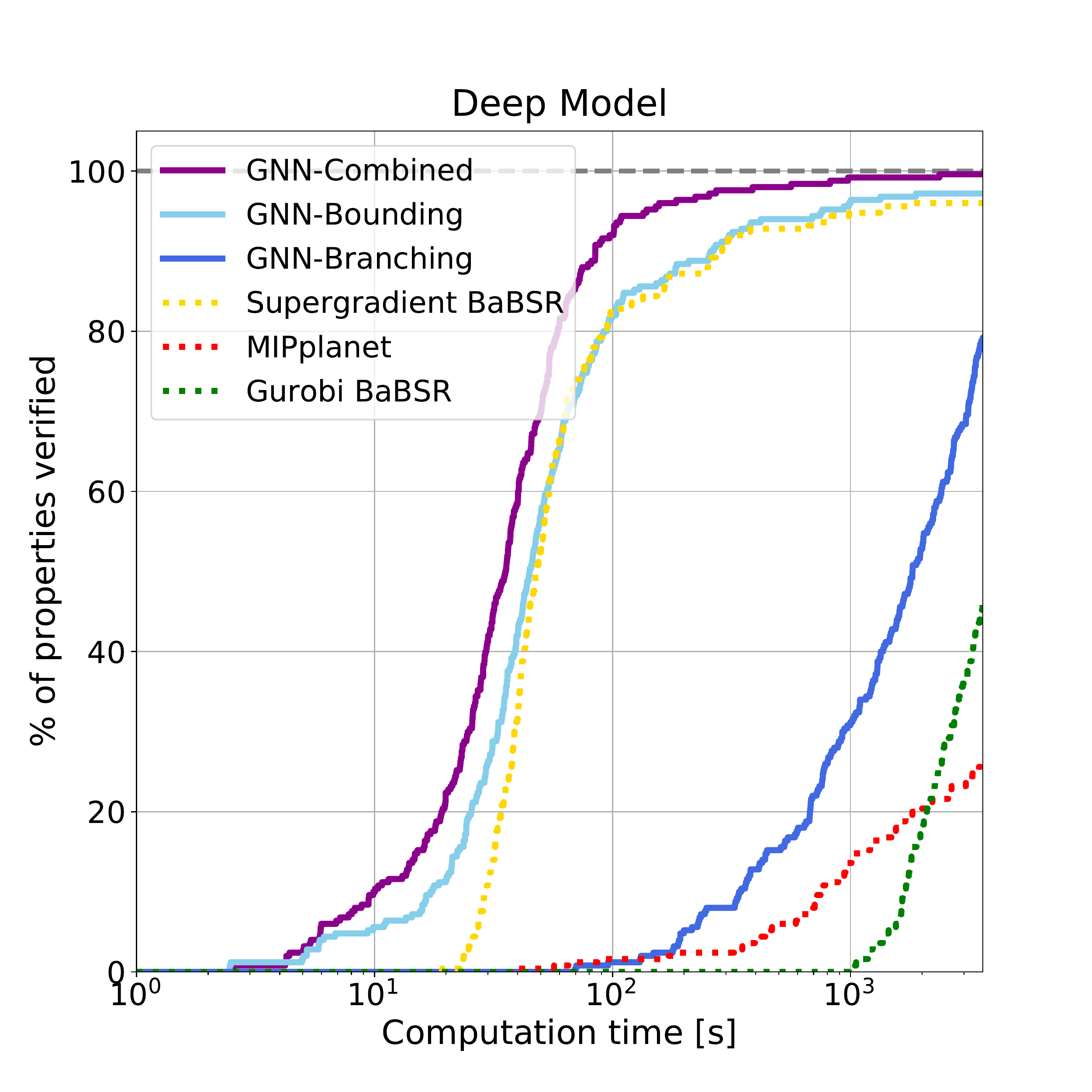}
	\end{subfigure}
	\caption{Cactus plots for the base, wide and deep models. For each, we compare the bounding methods and complete verification algorithms by plotting the percentage of solved properties as a function of runtime. Baselines are represented by dotted lines.}
	\label{fig:cactus:combined}
\end{figure*}

%% file: Tables/table_all_base_rebuttal.tex
 \sisetup{detect-weight=true,detect-inline-weight=math,detect-mode=true}
\begin{table*}[t]
	\centering
	\scriptsize
	\setlength{\tabcolsep}{4pt}
	\aboverulesep = 0.1mm  
	\belowrulesep = 0.2mm  
	

	\begin{adjustbox}{center}
		\begin{tabular}{
				l
				S[table-format=4.3]
				S[table-format=4.3]
				S[table-format=4.3]
				S[table-format=4.3]
				S[table-format=4.3]
				S[table-format=4.3]
				S[table-format=4.3]
				S[table-format=4.3]
				S[table-format=4.3]
			}
			& \multicolumn{3}{ c }{Easy} & \multicolumn{3}{ c }{Med} & \multicolumn{3}{ c }{Hard} \\
			\toprule
			
			\multicolumn{1}{ c }{Method} &
			\multicolumn{1}{ c }{time(s)} &
			\multicolumn{1}{ c }{subdomains} &
			\multicolumn{1}{ c }{$\%$Timeout} &
			\multicolumn{1}{ c }{time(s)} &
			\multicolumn{1}{ c }{subdomains} &
			\multicolumn{1}{ c }{$\%$Timeout} &
			\multicolumn{1}{ c }{time(s)} &
			\multicolumn{1}{ c }{subdomains} &
			\multicolumn{1}{ c }{$\%$Timeout} \\
			
			\cmidrule(lr){1-1} \cmidrule(lr){2-4} \cmidrule(lr){5-7} \cmidrule(lr){8-10}
			
			\multicolumn{1}{ c }{\textsc{Gurobi BaBSR}}
			&550.48
			&580.43
			&0.00
			
			&1374.32
			&1408.75
			&0.00
			
			&3129.08
			&2551.63
			&42.41 \\
			
			\multicolumn{1}{ c }{\textsc{MIPplanet}}
			&1499.35
			& 
			&16.49
			
			&2240.92
			& 
			&42.95

			&2255.53
			& 
			&46.35 \\
			
			\multicolumn{1}{ c }{\textsc{Supergradient}}
            &57.49
            &2529.30
            &0.21

            &82.66
            &5292.58
            &\B0.00

            &872.54
            &24188.27
            &17.41\\
			
			\multicolumn{1}{ c }{\textsc{GNN-Branching}}
			&272.69
			&285.68
			&0.21
			
			&592.12
			&583.21
			& 0.39
			
			&1573.16
			&1098.28
			&21.65 \\

			\multicolumn{1}{ c }{\textsc{GNN-Bounding}}
			&46.70
			&2201.70
			&\B0.00
			
			&78.85
			&4108.27
			&\B 0.00
			
			&665.64
			&16442.30
			&11.53 \\

			\multicolumn{1}{ c }{\textsc{GNN-Combined}}
			
            &\B37.09
            &930.33
            &0.21

            &\B57.97
            &2481.79
            &\B0.00

            &\B367.45
            &9397.48
            &\B5.65\\

			\bottomrule
			
		\end{tabular}
	\end{adjustbox}
	\caption{\small We compare average (mean) solving time, average number of subdomains solved, and the percentage of properties solved for easy, medium, and hard properties on the base model. The best performing method for each subcategory is highlighted in bold (note that by definition Gurobi BaBSR doesn't time out on easy and med experiments).}.
	\label{table:combined_base_exp_rebuttal}
\end{table*}

%% file: Tables/table_all_rebuttal.tex
 \sisetup{detect-weight=true,detect-inline-weight=math,detect-mode=true}
\begin{table*}[t]
	\centering
	\scriptsize
	\setlength{\tabcolsep}{4pt}
	\aboverulesep = 0.1mm  
	\belowrulesep = 0.2mm  
	
    
	\begin{adjustbox}{center}
		\begin{tabular}{
				l
				S[table-format=4.3]
				S[table-format=4.3]
				S[table-format=4.3]
				S[table-format=4.3]
				S[table-format=4.3]
				S[table-format=4.3]
				S[table-format=4.3]
				S[table-format=4.3]
				S[table-format=4.3]
			}
			& \multicolumn{3}{ c }{Base} & \multicolumn{3}{ c }{Wide} & \multicolumn{3}{ c }{Deep} \\
			\toprule
			
			\multicolumn{1}{ c }{Method} &
			\multicolumn{1}{ c }{time(s)} &
			\multicolumn{1}{ c }{subdomains} &
			\multicolumn{1}{ c }{$\%$Timeout} &
			\multicolumn{1}{ c }{time(s)} &
			\multicolumn{1}{ c }{subdomains} &
			\multicolumn{1}{ c }{$\%$Timeout} &
			\multicolumn{1}{ c }{time(s)} &
			\multicolumn{1}{ c }{subdomains} &
			\multicolumn{1}{ c }{$\%$Timeout} \\
			
			\cmidrule(lr){1-1} \cmidrule(lr){2-4} \cmidrule(lr){5-7} \cmidrule(lr){8-10}
			
			\multicolumn{1}{ c }{\textsc{Gurobi BaBSR}}
			&1588.02
			&1340.08
			&10.57
			
			&2917.95
			&855.37
			&51.11
			
			&3007.24
			&428.52
			&54.00 \\
			
			\multicolumn{1}{ c }{\textsc{MIPplanet}}
			& 2036.65
			& 
			&36.40
			
			& 3108.50
			& 
			&79.37
			
			& 2997.12
			& 
			&73.60 \\
			
			\multicolumn{1}{ c }{\textsc{Supergradient}}
            &277.22
            &8654.03
            &4.50

            &624.36
            &8885.64
            &12.87

            &241.80
            &3313.73
            &4.00\\
			
			\multicolumn{1}{ c }{\textsc{GNN-branching}}
			& 752.94
			& 604.16
			& 5.77

			& 1817.97
			& 418.68
			& 21.27

			& 1966.90
			& 202.73
			& 20.80\\

			\multicolumn{1}{ c }{\textsc{GNN-bounding}}
			& 219.62
			& 6427.10
			& 2.94
			
			& 513.63
			& 5274.19
			& 10.23
			
			& 214.55
			& 2784.49
			& 2.80\\

			\multicolumn{1}{ c }{\textsc{GNN-combined}}

            &\B 131.11
            &3731.92
            &\B 1.50

            &\B 269.11
            &2985.55
            &\B 4.95

            &\B76.10
            &1355.20
            &\B0.40\\

			\bottomrule
			
		\end{tabular}
	\end{adjustbox}
	\caption{\small We compare average (mean) solving time, average number of subdomains solved, and the percentage of properties that the methods time out on when using a cut-off time of 3600s. The best performing method for each subcategory is highlighted in bold.}
	\label{table:combined_base_wide_deep_rebuttal}
\end{table*}

%% file: Sections/Discussion.tex
\newpage
\section{Discussion}
We have shown how to improve the complete verification procedure using GNNs that learn how to use the underlying structure of the problem to return better bounds more quickly and to improve branching strategies. We show that our method consistently beats the existing state-of-the-art algorithms. Our GNNs trained on easy properties on a small network show good generalization performance on harder properties and on larger unseen networks. We've taken an important step towards creating verification methods for larger state-of-the-art networks. Further work might include extending our approach to work on different relaxations, such as the one proposed by \cite{anderson2019strong}, which is tighter than Planet but has significantly more constraints. Alternatively, one could learn a lazy verifier that only solves subdomains for which there is a high chance of pruning and further divides them into more subdomains otherwise.

%% file: Sections/Appendix.tex
\appendix

\section{Branch-and-Bound Algorithms}
\label{sec:app:bab}
The following generic Branch-and-Bound Algorithm is provided in \citet{journal2019}. Given a neural network \textit{net} and a verification property \textit{problem} we wish to verify, the BaB procedure examines the truthfulness of the property through an iterative procedure. During each step of BaB, we first use the \textit{pick$\_$out} function (line $6$) to choose a problem \textit{prob} to branch on. The split function (line $7$) determines the branching strategy and splits the chosen problem \textit{prob} into sub-problems. We compute output upper and lower bounds on each sub-problem with functions \textit{compute$\_$UB} and \textit{compute$\_$LB} respectively. Newly computed output upper bounds are used to tighten the global upper bound, which allows more sub-problems to be pruned. We prune a sub-problem if its output lower bound is greater than or equal to the global upper bound, so the smaller the global upper bound the better it is. Newly calculated output lower bounds are used to tighten the global lower bound, which is defined as the minimum of the output lower bounds of all remained sub-problems after pruning. We consider the BaB procedure converges when the difference between the global upper bound and the global lower bound is smaller than $\epsilon$.

In our case, our interested verification problem Equation~(\ref{eq:verif-prop}) is a satisfiability problem. We thus can simplify the BaB procedure by initialising the global upper bound \textit{global$\_$ub} as $0$. As a result, we prune all sub-problems whose output lower bounds are above $0$. In addition, the BaB procedure is terminated early when a below $0$ output upper bound of a sub-problem is obtained, which means a counterexample exits. 


\import{Algorithms/}{BAB.tex}

\section{Implementation of forward and backward passes for the branching GNN}\label{app:branching_implementation}
We give implementation details of forward and backward updates for embedding vectors for the model used in the experiments section. Choices of forward and backward update functions are based on the bounding methods used. In our experiments, we used linear bound relaxations for computing intermediate bounds and Planet relaxation for computing the final output lower bound. We start with a graph neural network mimicking the structure of the network we want to verify. We denote domain lower and upper bounds as $\boldsymbol{l}_{0}$ and $\boldsymbol{u}_{0}$ respectively. Similarly, we denote the intermediate bounds (pre-activation) for layers $i=1, \dots, L-1$ as $\boldsymbol{l}_i$ and $\boldsymbol{u}_i$. Since an LP solver is called for the final output lower bound, we have primal values for all nodes of $V$ and dual values for all ambiguous nodes of $V$. Finally, let $W^{1},\dots, W^{L}$ be the layer weights and $\boldsymbol{b}^1, \dots, \boldsymbol{b}^{L}$ be the layer biases of the network $f$, which we wish to verify.

\subsection{Motivation for Forward and Backward Passes}
We argue that the forward-backward updating scheme is a natural fit for our problem. In more detail, for a given problem $\mathcal{D}$, each branching decision (an input node or an ambiguous activation node) will generate two sub-problems $s_1$ and $s_2$, with each sub-domain having an output lower bound $l^{L}_{s_1}$ and $l^{L}_{s_2}$ respectively, equal to or higher than $l^{L}_{\cal{D}}$ the lower bound that of $\cal{D}$. Strong branching heuristic uses a predetermined function to measure the combined improvement of $l^{L}_{s_1}$ and $l^{L}_{s_2}$ over $l^{L}_{\mathcal{D}}$ and makes the final branching decision by selecting the node that gives the largest improvement. Thus, to maximise the performance of a graph neural network, we want a node embedding vector to maximally capture all information related to the computation of $l^{L}_{s_1}$ and $l^{L}_{s_2}$. For estimating $l^{L}_{s_1}$, $l^{L}_{s_2}$ of splitting on a potential branching decision node $v$, we note that these values are closely related to two factors. The first factor is the amount of convex relaxations introduced at a branching decision node $v$, when $v$ corresponds to an ambiguous activation node. The second factor considers that the impact that splitting node $v$ will have on the convex relaxations introduced to nodes on layers after that of $v$. Recall that, if there are no ambiguous activation nodes, the neural network $f$ is simply a linear operator, whose minimum value can be easily obtained. When ambiguous activation nodes are present, the total amount of relaxation introduced determines the tightness of the lower bound to $f$. We thus treat embedding vectors as a measure of local convex relaxation and its contribution to other nodes' convex relaxation.  

As shown in Figure~\ref{fig:convex-relax} in the main paper, at each ambiguous activation node $x'_{i[j]}$, the area of convex relaxation introduced is determined by the lower and upper bounds of the pre-activate node $\hat{x}_{i[j]}$. We observe that intermediate lower and upper bounds of a node $\hat{x}_{i[j]}$ are significantly affected by the layers prior to it and have to be computed in a layer-by-layer fashion. Based on the observation, we utilise a forward layer-by-layer update on node embedding vectors. This should allow these embedding vectors to capture the local relaxation information. In terms of the impact of local relaxation change to that of other nodes, we note that by splitting an ambiguous node into two fixed cases, all intermediate bounds of nodes on later layers will be affected, leading to relaxation changes at those nodes. We thus employ a backward layer-by-layer update to account for the impact the local change has over other nodes. Theoretically, by fixing an ambiguous ReLU node, intermediate bounds of nodes at previous layers and on the same layer might change as well. For a naturally trained neural network, the changes for these nodes should be relatively small compared to nodes on the later layers. To account for these changes, we rely on multiple rounds of forward-and-backward updates.

\subsection{Implementation of the Forward Pass}
Unless otherwise stated, all functions $F_{\ast}$ are 2-layer fully connected network with ReLU activation units.

\subsubsection{Input nodes}
We update the embedding vectors of input nodes only during the first round of forward pass. That is we update $\boldsymbol{\mu}_{0[j]}$ when it is zero for all $j$. After that, input nodes embedding vectors are updated only in backward pass. For each input node, we form the feature vector $\boldsymbol{z}_{0[j]}$ as a vector of $l_{0[j]}$, $u_{0[j]}$ and its associated primal solution. The input node embedding vectors are computed as
\begin{align}
    \boldsymbol{\mu}_{0[j]} = F_{inp}(\boldsymbol{z}_{0[j]} ; \boldsymbol{\theta}_0). 
\end{align}

\subsubsection{Activation nodes} 
The update function $F_{act}$ can be broken down into three parts: 1) compute information from local features 2) compute information from neighbourhood embedding vectors and 3) combine information from 1) and 2) to update current layer's embedding vectors.

\paragraph{Information from local features}
Since we compute the final lower bound with the Planet relaxation (Figure~\ref{fig:convex-relax}(c)), we introduce a new feature related to the relaxation: the intercept of the relaxation triangle, shown in Figure~\ref{fig:interceptKW}. We denote an intercept as $\beta$ and compute it as
\begin{align}\label{eq:intercept}
    \beta_{i[j]} = \frac{-l_{i[j]}\cdot u_{i[j]}}{u_{i[j]}-l_{i[j]}}.
\end{align}
The intercept of a relaxation triangle can be used as a measure of the amount of relaxation introduced at the current ambiguous node.

Therefore, the local feature vector $\boldsymbol{z}_{i[j]}$ of an ambiguous node $x'_{i[j]}$ consists of $l_{i[j]}$, $u_{i[j]}$, $\beta_{i[j]}$, its associated layer bias value, primal values (one for pre-activation variable and one for post-activation variable) and dual values. We obtain information from local features via
\begin{align}\label{eq:local-relax}
    R_{i[j]} = \begin{cases} F_{act-lf}(\boldsymbol{z}_{i[j]}; \boldsymbol{\theta}^0_1) & \text{if }x'_{i[j]}\text{ is ambiguous,} \\
    \boldsymbol{0} &\text{otherwise.}
    \end{cases}
\end{align}
where $R_{i[j]}\in \mathbb{R}^{p}$.

\begin{figure}[h]
  \centering
  \input{Figures/relu/interceptKW.tex}
\caption{\footnotesize Red line represents the intercept of the convex relaxation. It is treated as a measure of the shaded green area.}
\label{fig:interceptKW}
\end{figure}
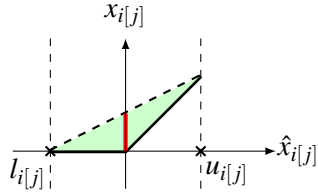

\begin{figure}[!h]
  \centering
  \input{Figures/relu/relusplitN.tex}
\caption{\footnotesize Depending on the value of $l_{i[j]}$ and $u_{i[j]}$, relaxed activation function can take three forms. The left figure shows the case where $l_{i[j]}$ and $u_{i[j]}$ are of different signs. In this case, for any input value between $l_{i[j]}$ and $u_{i[j]}$, the maximum output achievable is indicated by the red line. The middle figure shows the case where both $l_{i[j]}$ and $u_{i[j]}$ are no greater than zero. In this case, the activation function completely blocks all input information by outputting zero for any input value. The right figure shows the case where $l_{i[j]}$ and $u_{i[j]}$ are greater or equal to zero. In this case, the activation function allows complete information passing by outputting a value equal to the input value.}
\label{fig:intergate}
\end{figure}
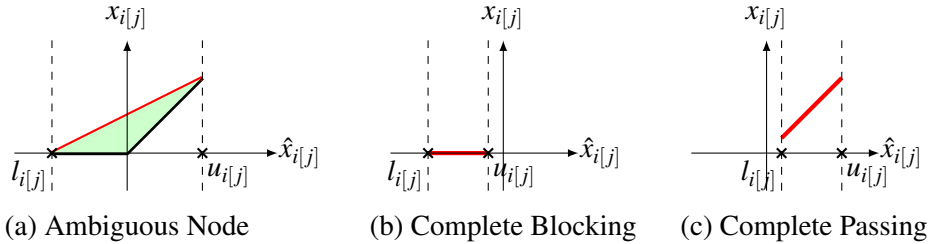
\paragraph{Information from neighbourhood embedding vectors}
During the forward pass, we focus on embedding vectors of the previous layer only. To update an embedding vector on layer $i$, we first combine embedding vectors of the previous layer with edge weights via
\begin{align}\label{eq:embedding}
    E_{i[j]} = \sum_{k} W^i_{kj}\cdot \boldsymbol{\mu}_{i-1[k]}.
\end{align}
\noindent To compute the information from neighbourhood embedding vectors to an arbitrary activation node $x'_{i[j]}$, we consider each activation unit as a \textit{gate}. We observe that the amount of the information from neighbourhood embedding vectors that remains after passing through a \textit{gate} is dependent on the its lower bound $l_{i[j]}$ and upper bound $u_{i[j]}$. When $l_{i[j]}$ and $u_{i[j]}$ are of different signs, $x'_{i[j]}$ is an ambiguous node. With relaxation, for any input value between $l_{i[j]}$ and $u_{i[j]}$, the maximum output achievable after passing an activation unit is shown by the red slope in Figure~\ref{fig:intergate}(a). The red slope $s_{i[j]}$ is computed as 
\begin{align}\label{eq:slope}
    s_{i[j]}(\hat{x}_{i[j]}) = \frac{u_{i[j]}}{u_{i[j]}-l_{i[j]}} \cdot \hat{x}_{i[j]} + \beta_{i[j]}.
\end{align}
Thus, the amount of information from neighbourhood embedding vectors that remains after passing through an ambiguous \textit{gate} is related to the ratio $\alpha \coloneqq \frac{u_{i[j]}}{u_{i[j]}-l_{i[j]}}$. When $u_{i[j]}$ is no greater than zero, the activation node $x'_{i[j]}$ completely blocks all information. For any input value, the output value is zero after passing the activation unit, as shown by the red line in Figure~\ref{fig:intergate}(b). We have $\alpha=0$ in this case. Finally, when $l_{i[j]}$ is no less than $0$, the activation node $x'_{i[j]}$ allows a complete passing of information and $\alpha=1$. It is shown by the red line in Figure~\ref{fig:intergate}(c). We incorporate these observations into our evaluations and compute the information from neighbourhood embedding vectors as 
\begin{align}
    N_{i[j]} = f_{act-nb}([\alpha\cdot E_{i[j]}, \alpha'\cdot E_{i[j]}];\boldsymbol{\theta}^1_1),
\end{align}
where $\alpha' = 1-\alpha$ when $0< \alpha <1$ and $\alpha' = \alpha$ otherwise. Here, we use $[\boldsymbol{a},\boldsymbol{b}]$ to denote the concatenation of two vectors $\boldsymbol{a}, \boldsymbol{b} \in \mathbb{R}^{p}$ into a vector of $\mathbb{R}^{2p}$. We introduce $\alpha'$ to be more informative. We do not consider the information that relate to the intercept $\beta_{i[j]}$ in the ambiguous case for the sake of simplicity. Improved performance could be expected if the $\beta_{i[j]}$ related information is incorporated as well. 

\paragraph{Combine previous information}
Finally, we combine the information from local features and the information from neighbourhood embedding vectors to update the embedding vectors of activation nodes. Specifically, 
\begin{align}
    \boldsymbol{\mu}_{i[j]} = F_{act-com}([R_{i[j]}, N_{i[j]}];\boldsymbol{\theta}^2_1).
\end{align}

\subsubsection{Output node} 
Embedding vectors of output nodes are updated in a similar fashion to that of activation nodes. We first compute information from local features. 
\begin{align}
    R_{L{j}} = F_{out-lf}(\boldsymbol{z}_{L{j}};\boldsymbol{\theta}^0_2)
\end{align}
For output nodes, the vector of local features $\boldsymbol{z}_{L}$ consists of output lower bound, output upper bound, primal solution and layer bias. $F_{out-lf}$ is a one-layer fully-connected network with ReLU activation units. We then compute information from neighbourhood embedding vectors. Since the output node does not have an activation unit associated with it, we directly compute the information of neighbourhood embedding vectors as
\begin{align}
    E_{L[j]} = \sum_{k} W^{L}_{kj}\cdot \boldsymbol{\mu}_{L-1[k]}.
\end{align}
Finally, we update the embedding vector of the output node as 
\begin{align}
    \boldsymbol{\mu}_{L{j}} = F_{out-com}([R_{L[j]}, E_{L[j]}]; \boldsymbol{\theta}^1_2).
\end{align}

\subsection{Implementation of the Backward Pass}
During backward message passing, for $i=L-1, \dots, 1$, we update embedding vectors for activation nodes and input node. Again, all functions $B_{\ast}$ are 2-layer fully-connected networks unless specified otherwise.

\subsubsection*{B.2.1 Activation nodes}
Similar to updates of embedding vectors carried out for activation nodes in a forward pass, we update embedding vectors of activation nodes using the same three steps in the backward pass, but with minor modifications.

\paragraph{Information from local features}
We use the same feature $\boldsymbol{z}_{i[j]}$ as the one used in the forward pass and compute the information from local features as 
\begin{align}
     R^b_{i[j]} = \begin{cases} B_{act-lf_1}(\boldsymbol{z}_{i[j]};\boldsymbol{\theta}_3^0) & \text{if }x'_{i[j]}\text{ is ambiguous,} \\
    \boldsymbol{0} &\text{otherwise.}
    \end{cases}
\end{align}
We recall that a dual value indicates how the final objective function is affected if its associated constraint is relaxed by a unit. To better measure the importance of each relaxation to the final objective function, we further update the information from local features by 
\begin{align}
    R^{b'}_{i[j]} = \begin{cases} B_{act-lf_2}([\boldsymbol{d}_{i[j]} \odot R^b_{i[j]}, R^b_{i[j]}]; \boldsymbol{\theta}_3^1) & \text{if }R^b_{i[j]}\neq \boldsymbol{0} \\
    \boldsymbol{0} &\text{otherwise.}
    \end{cases}
\end{align}
Here, $\boldsymbol{d}_{i[j]}$ is the vector of dual values corresponding to the activation node $x'_{i[j]}$. We use $\odot$ to mean that we multiply $R^b_{i[j]}$ by each element value of $\boldsymbol{d}_{i[j]}$ and concatenate them as a singe vector.

\paragraph{Information from neighbourhood embedding vectors}
During the backward pass, we focus on embedding vectors of the next layer only. In order to update an embedding vector on layer $i$, we compute the neighbourhood embedding vectors as  
\begin{align}\label{eq:embedding-b}
    E^b_{i[j]} = \sum_{k} W^{i+1}_{jk}\cdot \boldsymbol{\mu}_{i+1[k]}.
\end{align}
We point out that there might be an issue with computing $E_{i[j]}$ if the layer $i+1$ is a convolutional layer in the backward pass. For a convolutional layer, depending on the padding number, stride number and dilation number, each node $x'_{i[j]}$ may connect to a different number of nodes on the layer $i+1$. Thus, to obtain a consistent measure of $E_{i[j]}$, we divide $E_{i[j]}$ by the number of connecting node on the layer $i+1$, denoted as $E^{b'}_{i[j]}$ and use the averaged $E^{b'}_{i[j]}$ instead. Let 
\begin{align}\label{eq:nb-embedding-2}
    E^{b\ast}_{i[j]} = \begin{cases} E^{b'}_{i[j]} & \text{if layer i+1 convolutional,} \\
    E^b_{i[j]} &\text{otherwise.}
    \end{cases}
\end{align}
The following steps are the same as the forward pass. We first evaluate 
\begin{align}
    N^{b}_{i[j]} = B_{act-nb}([\alpha\cdot E^{b\ast}_{i[j]}, \alpha'\cdot E^{b\ast}_{i[j]}]; \boldsymbol{\theta}_3^2),
\end{align}
and the update embedding vectors as
\begin{align}
    \boldsymbol{\mu}_{i[j]} = B_{act-com}([R^{b'}_{i[j]}, N^{b}_{i[j]}]; \boldsymbol{\theta}_3^3).
\end{align}

\subsubsection*{B.2.2 Input nodes}
Finally, we update the input nodes. We use the feature vector $\boldsymbol{z}^b_0$, which consists of domain upper bound and domain lower bound. Information from local features is evaluated as 
\begin{align}
    R_{0{j}} = B_{inp-lf}(\boldsymbol{z}^b_{0[j]}; \boldsymbol{\theta}_4^0).
\end{align}
We compute the information from neighbourhood embedding vectors in the same manner as we do for activation nodes in the backward pass, shown in Eq~(\ref{eq:nb-embedding-2}). Denote the computed information as $E^{b\ast}_{0[j]}$. The embedding vectors of input nodes are updated by
\begin{align}
    \boldsymbol{\mu}_{0[j]} = B_{inp-com}([R^{b'}_{0[j]}, E^{b\ast}_{0[j]}]; \boldsymbol{\theta}_4^1).
\end{align}

\section{Implementation of forward and backward passes for the bounding GNN}
In our case $g$ is a multilayer perceptron (MLP), which is made up of a series of linear layers $\Theta_i$ and non-linear activations $\sigma$.
We have the following set of trainable parameters:
\begin{equation}
    \Theta_0 \in \RR^{d \times p}, \quad
    \Theta_1, \dots, \Theta_{T_1} \in \RR^{p \times p}, \quad
    \bb_0, \dots, \bb_{T_1} \in \RR^{p}.
\end{equation}
Given a feature vector $\feat$ we compute the following set of vectors:
\begin{equation}
    \mmu^0 = \Theta_0 \cdot \feat + b_0, \quad
    \mmu^{l+1} = \Theta_{l+1} \cdot \relu(\mmu^l) + \bb_{l+1}.\\
\end{equation}
We initialize the embedding vector to be $\mmu = \mmu^{T_1}$, where $T_1+1$ is the depth of the MLP.

 The hyper-parameters for the GNN computation of the duals are the depth of the MLP ($T_1$), how many forward and backward passes we run ($T_2$), and the embedding size ($p$).

 The forward pass consists of a weighted sum of three parts: the first term is the current embedding vector, the second is the embedding vector of the previous layer passed through the corresponding linear or convolutional filters, and the third is the average of all neighbouring embedding vectors:
\begin{equation}
    \mmu'_{i[j]} = \relu \left( \thetafor_1 \mmu_{i[j]} + \thetafor_2 \left(W_i \mmu_{i-1} + \bb_{i-1}\right){[j]} + \thetafor_3 \left(\sum_{k \in N(j)} \mmu_{{i-1}[k]} / Q_{{i}[j]}\right){[j]} \right).
\end{equation}
Both the second and the third term can be implemented using existing deep learning functions.
Similarly, we perform a backward pass as follows:
\begin{equation}
    \mmu_{i[j]} = \relu \left( \thetaback_1 \mmu'_{i[j]} + \thetaback_2 (W_{i+1}^{T} \left(\mmu'_{i+1} - \bb_{i+1}\right))[j] + \thetaback_3 \left(\sum_{k \in N'(j)} \mmu'_{{i+1}[k]} / Q'_{{i}[j]}\right)[j] \right).
\end{equation}
Here $\thetafor_1, \thetafor_2, \thetafor_3, \thetaback_1, \thetaback_2, \thetaback_3 \in \RR^{p \times p}$ are all learnable parameters.
To ensure better generalization performance to unseen neural networks with a different network architecture we include normalization parameters $Q$ and $Q'$. These are matrices whose elements are the number of neighbouring nodes in the previous and following layer respectively for each node.
We repeat this process of running a forward and backward pass $T_2$ times. The high-dimensional embedding vectors are now capable of expressing the state of the corresponding node taking the entire problem structure into consideration as they are directly influenced by every single other node, even if we set $T_2 = 1$.

\section{Algorithm for Generating the Training Dataset}
\subsection{Branching}\label{sec:app:training_branching}
Algorithm~\ref{alg:train-gen} outlines the procedure for generating the training dataset. The algorithm ensures the generated training date have a wide coverage both in terms of the verification properties and BaB stages while at the same time is computationally efficient. Specifically, we randomly pick $25\%$ of all properties that do not time out and run a complete BaB procedure on each of them with the strong branching heuristic to generate training samples (line $3$-$5$). For the remaining properties, we attempt to generate $B$ training samples for each of them. To cover different stages of a BaB process of a property, we use a computationally cheap heuristic together with the strong branching heuristic. Given a property, we first use the cheap heuristic for $k$ steps (line $10$-$15$) to reach a new stage of the BaB procedure and then call the strong branching heuristic to generate a training sample (line $16$). We repeat the process until $B$ training samples are generated or the BaB processs terminates.

\begin{algorithm}[H]
      \caption{Generating Training Dataset}\label{alg:train-gen}
      \begin{algorithmic}[1]
      \State{Provided: total $P$ properties; minimum $B$ training data for each property; a maximum $q$ branches between strong branching decisions}
        \For{$p = 1, \dots, P$}:
         \State{$\alpha \longleftarrow \text{random number from }[0, 1]$}
         \If{$p$ is not a timed out property and $\alpha \leq 0.25$ }
         \State{Running a complete BaB process with the Strong Branching Heuristic}
         \Else
            \State{$b=0$}
            \While{$b\leq B$}    
                 \State{$k \longleftarrow \text{random integer from} [0, q]$}
                 \While{$k >0$}
                 \State{Call a computationally cheap heuristic}
                 \If{BaB process terminates}
                     \Return
                 \EndIf
                 \State{$k = k -1 $}
            \EndWhile
            \State{Call the strong branching heuristic and generate a training sample}
            \If{BaB process terminates}
                 \Return
            \EndIf
            \State{$b=b+1$}
        \EndWhile
        \EndIf
        \EndFor
      \end{algorithmic}
    \end{algorithm}

\subsection{Bounding}
\label{sec:app:training_bounding}
We train our bounding GNN on 100 properties, taken from the training dataset used for the branching GNN. All of the training properties used are easy; that is BaBSR takes less than 800 seconds to solve them.
We train the bounding GNN for three iterations, as explained above.
For each iteration we train the GNN on 10,000 subdomains for a total of 50 epochs.
At the start of each of the three iterations, we randomly select the subdomains to train on, choosing the same number of subdomains for each property.
We aim to minimize the loss function (\ref{loss_function}) using a horizon of 100 and a decay factor $\gamma = 0.99$.
We train the GNN using the Adam optimizer \cite{kingma2014adam} with a learning rate of $1e^{-2}$ and no weight decay; we manually decay the learning rate by a factor of 10 if the loss function doesn't improve for two consecutive epochs.
For the update step we use an initial step size of $\mu = 1e^{-3}$, and decay it as explained above.
We set the embedding size to be 32 for all GNNs. Moreover, we set $T_1 = 1$ and $T_2 = 1$. That is, we use a 2-layer MLP to initialize the embedding vectors and perform just one set of forward and backward passes.
At the beginning of the first iteration we create the dataset by running the BaB algorithm using supergradient ascent and Adam to compute the lower bounds; we set the learning rate to be $1e^{-4}$.
For the second and third iterations we further extend the dataset, this time using the current version of the GNN to compute the final lower bounds.


\section{Experiment Details for the Branching GNN}
All the hyper-parameters used in the experiments are determined by testing a small set of numbers over the validation set. Due to the limited number of tests, we believe better sets of hyper-parameters could be found.

\subsection{Training Details}
\paragraph{Training dataset} To generate a training dataset, 565 random images are selected. Binary serach with BaBSR and 800 seconds timeout are used to determine $\epsilon$ on the Base model. Among 565 verification properties determined, we use 430 properties to generate 17958 training samples and the rest of properties to generate 5923 validation samples. Training samples and validation samples are generated using Algorithm~\ref{alg:train-gen} with $B=20$ and $q=10$.

For a typical epsilon value, each sub-domain generally contains 1300 ambiguous ReLU nodes. Among them, approximately 140 ReLU nodes are chosen for strong branching heuristics, which leads to roughly 200 seconds for generating a training sample. We point out that the total amount of time required for generating a training sample equals the 2*(per LP solve time)*(number of ambiguous ReLU nodes chosen). Although both the second and the third terms increase with the size of the model used for generating training dataset, the vertical transferability of our GNN enables us to efficiently generate training dataset by working with a small substitute of the model we are interested in. In our case, we trained on the Base model and generalised to Wide and Deep model.

\paragraph{Training} We initialise a GNN by assigning each node a $64$-dimensional zero embedding vector. GNN updates embedding vectors through two rounds of forward and backward updates. To train the GNN, we use hinge rank loss (Equation~(9)) with $M=10$. Parameters $\boldsymbol{\Theta}$ are computed and updated through Adam optimizer with weight decay rate $\lambda=1e^{-4}$ and learning rate $1e^{-4}$. If the validation loss does not decrease for $10$ consecutive epochs, we decrease the learning rate by a factor of $5$. If the validation loss does not decrease for $20$ consecutive epochs, we terminate the learning procedure. The batch size is set to $2$.  In our experiments, each training epoch took less than 400 seconds and the GNN converges within 60 epochs.

In terms of the training accuracy, we first evaluate each branching decision using the metric defined by Equation~(\ref{eq:improvement}) \footnote{we have tried various other metrics, including picking the minimum of the two subdomain lower bounds and the maximum of the two lower bounds. Among these metrics, metric defined by Equation~(\ref{eq:improvement}) performs the best.}. Since there are several branching choices that give similar performance at each subdomain, we considered all branching choices that have $m_v$ above 0.9 as correct decisions. Under this assumption, our trained GNN achieves $85.8\%$ accuracy on the training dataset and $83.1\%$ accuracy on the validation dataset.

\subsection{Verification Experiment Details}
We ran all verification experiments in parallel on $16$ CPU cores, with one property being verified on one CPU core. We observed that although we specifically set the thread number to be one for MIPplanet (backed by the commercial solver Gurobi), the time required for solving a property depends on the total number of CPUs used. For a machine with 20 cpu cores, MIPplanet requires much less time on average for proving the same set of properties on fewer (say 4) CPU cores in parallel than on many (say 16) CPU cores in parallel (the rest of CPU cores remain idle). Since BaBSR, and the branching GNN both use Gurobi for the bounding problems, similar time variations, depending on the number of CPU cores used, are observed. We ran each method in the same setting and on 16 CPUs in parallel, so our reported results and time are comparable. However, we remind readers to take the time variation into consideration when replicating our experiments or using our results for comparison. 

\paragraph{Fail-safe strategy}
Since, to the best of our knowledge, the branching heurisitc of BaBSR is the best performing one on convolutional neural networks so far, we choose it for our fail-safe strategy. The threshold is set to be $0.2$. Every time when the relative improvement $m_{gnn}$ of a GNN branching decision $v_{gnn}$ is less than $0.2$, we call the heuristic to make a new branching decision $v_{h}$. We solve the corresponding LPs for the new branching decision and compute its relative improvement $m_h$. The node with higher relative improvement is chosen to be the final branching decision.


\subsection{Baselines}
We decided our baselines based on the experiment results of \citet{journal2019}. In \citet{journal2019}, methods including MIPplanet, BaBSR, planet~\citep{Ehlers2017}, reluBaB and reluplex~\citep{Katz2017} are compared on a small convolutional MNIST network. Among them, BaBSR and MIPplanet significantly outperform other methods. We thus evaluate our methods against these two methods only in the experiments section. In order to strengthen our baseline, we compare against two additional methods here.

\paragraph{Neurify~\citep{Wang2018}} Similar to BaBSR, Neurify splits on ReLU activation nodes. It makes a branching decision by computing gradient scores to prioritise ReLU nodes. Since the updated version of Neurify's released code supports verification, we conducted a comparison experiment between between Neurify and BaBSR for inclusiveness. 

Neurify does not support CIFAR dataset. To evaluate the performance of Neurify, we obtained the trained ROBUST MNIST model and corresponding verification properties from \citet{journal2019}. We ranked all verification properties in terms of the BaBSR solving time and selected the first 200 properties, which are solved by BaBSR within one minute, as our test properties. For a fair comparison, we have restricted Neurify to use one CPU core only and set the timeout limit to be two minutes. Among all test properties, Neurify timed out on 183 out of 200 properties. BaBSR thus outperforms Neurify significantly. Combining with the results of \citet{journal2019}, BaBSR is indeed a fairly strong baseline to be compared against. 

\paragraph{MIP based algorithm~\citep{newMIP}} We also compared our MIPplanet baseline against a new MIP based algorithm \citep{newMIP}, published in ICLR 2019. To test these two methods, we randomly selected 100 verification properties from the CIFAR Base experiment with timeout 3600s. In terms of solving time, MIPplanet requires 1732.18 seconds on average while the new MIP algorithm requires 2736.60 seconds. Specifically, MIPplanet outperforms the new MIP algorithm on 78 out of 100 properties. MIPplanet is therefore a strong baseline for comparison.   

As a caveat, we mention that the main difference between MIPplanet and the algorithm of \citep{newMIP} is the intermediate bound computation, which is complementary to our focus. If better intermediate bounds are shown to help verification, we can still use our approach to get better branching decisions corresponding to those bounds.

\subsection{Model Architecture}
\label{sec:app:network_arch_verification}
We provide the architecture detail of the neural networks verified in the experiments in the following table. 
\begin{table}[h!]
\centering
\small
  \begin{tabular}{|c|c|c|}
    \hline
    \textbf{Network Name}\TBstrut & \textbf{No. of Properties} \TBstrut & \textbf{Network Architecture} \TBstrut\\
    \hline
    \begin{tabular}{l}
      BASE \\ 
      Model
   \end{tabular} & \begin{tabular}{@{}c@{}}
        Easy: 467  \\
        Medium: 773 \\
        Hard: 426
   \end{tabular} & \begin{tabular}{@{}c@{}} 
                    \footnotesize
                      Conv2d(3,8,4, stride=2, padding=1) \Tstrut \\
                      Conv2d(8,16,4, stride=2, padding=1)\\
                      linear layer of 100 hidden units \\
                      linear layer of 10 hidden units\\
                      (Total ReLU activation units: 3172) \Bstrut
                      \end{tabular}\\
    \hline
    WIDE & 300  & \begin{tabular}{@{}c@{}} 
                        \footnotesize
                      Conv2d(3,16,4, stride=2, padding=1) \Tstrut\\
                      Conv2d(16,32,4, stride=2, padding=1)\\
                      linear layer of 100 hidden units \\
                      linear layer of 10 hidden units\\
                      (Total ReLU activation units: 6244) \Bstrut
                      \end{tabular}\\
    \hline
    DEEP & 250 & \begin{tabular}{@{}c@{}} 
                        \footnotesize
                      Conv2d(3,8,4, stride=2, padding=1) \Tstrut \\
                      Conv2d(8,8,3, stride=1, padding=1) \\
                      Conv2d(8,8,3, stride=1, padding=1) \\
                      Conv2d(8,8,4, stride=2, padding=1)\\
                      linear layer of 100 hidden units \\
                      linear layer of 10 hidden units\\
                      (Total ReLU activation units: 6756) \Bstrut
                      \end{tabular}\\
    \hline
  \end{tabular}
  \caption{\label{tab:problem_size} For each CIFAR experiment, the network architecture used and the number of verification properties tested. }
\end{table}
\pagebreak
\section{Additional Experiment Results --- Branching GNN}
\label{sec:app:further_experiments_branching}
\subsection{Fail-safe heuristic dependence}
In all our experiments, we have compared against BaBSR, which employs only the fail-safe heuristic for branching. In other words, removing the GNN and using only the fail-safe heuristic is equivalent to BaBSR. The fact that GNN significantly outperforms BaBSR demonstrates that GNN is doing most of the job. To better evaluate the GNN's reliance on a fail-safe heuristic, we study the ratio of times that a GNN branching decision is used for each verification property of a given model. Results are listed in Table~\ref{table:fail-safe}. On all three models, GNN accounts for more than $90\%$ of branching decisions employed on average, ensuring the effectiveness of our GNN framework. 

\sisetup{detect-weight=true,detect-inline-weight=math,detect-mode=true}
\begin{table}[H]
\centering
\caption{\footnotesize Evaluating GNN's dependence on the fail-safe strategy. Given a CIFAR model, we collected the percentage of times GNN branching decision is used and the percentage of times the fail-safe heuristic (BaBSR in our case) is employed for each verification property. We report the average ratio of all verification properties of the same model. To account for extreme cases, we also list the minimum and maximum usage ratios of the fail-safe heuristic for each model. }
\label{table:fail-safe}
\scriptsize
\setlength{\tabcolsep}{4pt}
\aboverulesep = 0.1mm  
\belowrulesep = 0.2mm  
\begin{tabular}{
    l
    S[table-format=4.3]
    S[table-format=4.3]
    S[table-format=4.3]
    S[table-format=4.3]
    S[table-format=4.3]
    S[table-format=4.3]
    }
   
    \toprule

    \multicolumn{1}{ c }{Model} &
    \multicolumn{1}{ c }{GNN(avg)} &
    \multicolumn{1}{ c }{BaBSR(avg)} &
    \multicolumn{1}{ c }{BaBSR(min)} &
    \multicolumn{1}{ c }{BaBSR(max)} \\

    \cmidrule(lr){1-1} \cmidrule(lr){2-5} 

    \multicolumn{1}{ c }{\textsc{Base}}
        &0.934
        &0.066
        &0.0
        &0.653 \\

    \multicolumn{1}{ c }{\textsc{Wide}}
        &0.950
        &0.050
        &0.0
        &0.274\\

    \multicolumn{1}{ c }{\textsc{Deep}}
        &0.964
        &0.036
        &0.0
        &0.290 \\

    \bottomrule

\end{tabular}
\end{table}

\subsection{GNN feature analysis}
We evaluate the importance of different features used in GNN. We note that two types of features are used in GNN. The first type (including intermediates bounds, network weights and biases) can be collected at negligible costs. The other type is LP features (primal and dual values) that are acquired by solving a strong LP relaxation, which are expensive to compute but potentially highly informative. To evaluate their effect, we trained a new GNN with LP features removed and tested the new GNN on 260 randomly selected verification properties on the Base model.
Among the selected properties, 140 are categorised as easy, 70 as medium and 50 as hard. We denote the model trained on all features as GNN and the newly trained model as GNN-R (we use R to indicate reduced features).
\sisetup{detect-weight=true,detect-inline-weight=math,detect-mode=true}
\begin{table}[H]
\centering
\caption{\footnotesize Measuring the importance of features used by GNN. For easy, medium and difficult level verification properties, we compare methods' average solving time, average number of branches required and the percentage of timed out properties.}
\label{table:nodual}
\vspace{-5pt}
\scriptsize
\setlength{\tabcolsep}{4pt}
\aboverulesep = 0.1mm  
\belowrulesep = 0.2mm  
\begin{adjustbox}{center}
\begin{tabular}{
    l
    S[table-format=4.3]
    S[table-format=4.3]
    S[table-format=4.3]
    S[table-format=4.3]
    S[table-format=4.3]
    S[table-format=4.3]
    S[table-format=4.3]
    S[table-format=4.3]
    S[table-format=4.3]
    }
    & \multicolumn{3}{ c }{Easy} & \multicolumn{3}{ c }{Medium} & \multicolumn{3}{ c }{Hard} \\
    \toprule

    \multicolumn{1}{ c }{Method} &
    \multicolumn{1}{ c }{time(s)} &
    \multicolumn{1}{ c }{branches} &
    \multicolumn{1}{ c }{$\%$Timeout} &
    \multicolumn{1}{ c }{time(s)} &
    \multicolumn{1}{ c }{branches} &
    \multicolumn{1}{ c }{$\%$Timeout} &
    \multicolumn{1}{ c }{time(s)} &
    \multicolumn{1}{ c }{branches} &
    \multicolumn{1}{ c }{$\%$Timeout} \\

    \cmidrule(lr){1-1} \cmidrule(lr){2-4} \cmidrule(lr){5-7} \cmidrule(lr){8-10}

    \multicolumn{1}{ c }{\textsc{BaBSR}}
        & 429.589
        & 641.300
        & 0.0

        & 1622.669
        & 1504.366
        & 0.0

        & 2466.712
        & 1931.098
        &0.0 \\

    \multicolumn{1}{ c }{\textsc{GNN}}
        &268.592
        &319.386
        &0.0

        &724.883
        &529.070
        &0.0

        &1025.826
        &772.667
        &0.0 \\

    \multicolumn{1}{ c }{\textsc{GNN-R}}
         &348.482 
         &441.043
         &0.0

         &898.011
        &720.958
        &0.0

         &1340.559
         &967.804
         &0.0 \\

    \bottomrule

\end{tabular}
\end{adjustbox}
\end{table}
From Table~\ref{table:nodual}, we observe that removing primal and dual information deteriorates the GNN performance, but GNN-R still outperforms the baseline heuristic BaBSR. We believe cheap features are the most important.  Depending on the cost of LP, potential users can either remove expensive LP features or train a GNN with a smaller architecture. 



\sisetup{detect-weight=true,detect-inline-weight=math,detect-mode=true}
\begin{table}[H]
\centering
\caption{\footnotesize Methods' performance on randomly selected properties. We show methods' average solving time, average number of branches required and the percentage of timed out properties. We emphasize that MIPplanet branch number is not comparable with those of other methods. }
\label{table:grb-branches}
\scriptsize
\setlength{\tabcolsep}{4pt}
\aboverulesep = 0.1mm  
\belowrulesep = 0.2mm  
\begin{adjustbox}{center}
\begin{tabular}{
    l
    S[table-format=4.3]
    S[table-format=4.3]
    S[table-format=4.3]
    S[table-format=4.3]
    S[table-format=4.3]
    S[table-format=4.3]
    S[table-format=4.3]
    S[table-format=4.3]
    S[table-format=4.3]
    }
    & \multicolumn{3}{ c }{Base} & \multicolumn{3}{ c }{Wide} & \multicolumn{3}{ c }{Deep} \\
    \toprule

    \multicolumn{1}{ c }{Method} &
    \multicolumn{1}{ c }{time(s)} &
    \multicolumn{1}{ c }{branches} &
    \multicolumn{1}{ c }{$\%$Timeout} &
    \multicolumn{1}{ c }{time(s)} &
    \multicolumn{1}{ c }{branches} &
    \multicolumn{1}{ c }{$\%$Timeout} &
    \multicolumn{1}{ c }{time(s)} &
    \multicolumn{1}{ c }{branches} &
    \multicolumn{1}{ c }{$\%$Timeout} \\

    \cmidrule(lr){1-1} \cmidrule(lr){2-4} \cmidrule(lr){5-7} \cmidrule(lr){8-10}

    \multicolumn{1}{ c }{\textsc{BaBSR}}
        & 1472.508
        & 1420.839
        & 0.067

        & 2985.199
        & 918.167
        & 0.111

        & 3811.712
        & 482.167
        &0.111 \\

    \multicolumn{1}{ c }{\textsc{MIPplanet}}
        &  1783.800
        &  3780.408$^{\ast}$
        &  0.258
        
        & 5254.134
        & 2949.625$^{\ast}$
        & 0.556

        & 4566.080
        & 4332.375$^{\ast}$
        & 0.407 \\

    \multicolumn{1}{ c }{\textsc{GNN}}
        & 714.224
        & 817.017
        & 0.017
        
        & 996.811
        & 268.333
        & 0.074

        &1893.081
        &201.500
        &0.0 \\

    


    \bottomrule

\end{tabular}
\end{adjustbox}
\end{table}

\subsection{LP solving time and GNN computing time}
 We mention that LP solving time is the main bottleneck for branch-and-bound based verification methods. Although both GNN evaluation time and LP solving time increase with the size of network, LP solving time grows at a significantly faster speed. For instance, in CIFAR experiments, GNN requires on average 0.02, 0.03, 0.08 seconds to make a branching decision on Base, Wide and Deep model respectively but the corresponding one LP solving time on average are roughly 1.1, 4.9, 9.6 seconds. GNN evaluation is almost negligible for large neural networks when compared to LP solving time. 

\newpage

\section{Supergradient Method}\label{sec:app:supergradient_ascent}
We will now outline the supergradient ascent method used in \citet{bunel2020lagrangian}.

\import{Algorithms/}{Supergradient.tex}\\

\section{Regaining Supergradient Ascent}
\label{sec:app:proof_prop}

We show that our method is strictly more expressive than supergradient ascent by showing that it can simulate it exactly.

 The supergradient ascent step is equivalent to update step
 $\rrho^{t+1} = \rrho^{t} + \eta^{t+1} \rrhohat^{t+1}$, where 
 \begin{equation}
     \rrhohat_k^{t+1} = \zhat_{B, k} - \zhat_{A, k}.
 \end{equation}

 Let $\Theta_0$ be the zero-matrix with non-zero elements  $\Theta_0[{1,4}]=1$, $\Theta_0[{2,4}]=-1$.
 Moreover, setting $T_1=1$, $\Theta_1 = \eye$ and $\bb_0 = \bb_1=\zero$, we get
    \begin{align}
     \mmu^0_k &= \left({\zhat_{B, k} - \zhat_{A, k}}, -{\zhat_{B, k} + \zhat_{A, k}}, \zero, \dots, \zero \right)^{\top},\\
     \mmu &= \left( \left(\zhat_{B, k} - \zhat_{A, k}\right)_+, -\left(\zhat_{B, k} - \zhat_{A, k}\right)_-, \zero, \dots, \zero \right)^{\top}.
    \end{align}

 If we set $\thetafor_2 = \thetafor_3 = \thetaback_2 = \thetaback_3 = \zero$ and $\thetafor_1 = \thetaback_1 = \eye$, then the forward and backward passes don't change the embedding vector.
 We now just need to set $\thetascore = (1, -1, 0, \dots, 0)^{\top}$ to get the final ascent direction:
 \begin{equation}
     \rrhohat_k^{t+1}
     = \left(\zhat_{B, k} - \zhat_{A, k}\right)_+ + \left(\zhat_{B, k} - \zhat_{A, k}\right)_-
     = \zhat_{B, k} - \zhat_{A, k}.
 \end{equation}
 We have shown that we can simulate supergradient ascent using our GNN architecture.

\newpage
\section{Experiment Setup --- Bounding GNN}
We will now explain in greater detail how the experiments described in this paper were run including all hyperparameters used in the verification experiments.

 \subsection{Verification Experiments}
 For all verification experiments mentioned in this work we run the BaB algorithm outlined in appendix \ref{sec:app:bab}.
We run 100 iterations of the GNN compared to 500 when using supergradient ascent. This together with the significant reduction in subdomains visited in the BaB algorithm when using the GNN more than compensates for the fact that one iteration of the GNN takes longer than one iteration of supergradient ascent.
 If the GNN performs poorly on a subdomain and the fail-safe method is used, it is likely to also not do well on the child subdomains. We therefore use supergradient ascent to solve all subdomains that result from further subdividing the current one.
For all experiment we use a batch-size of 300 for both the GNN and supergradient descent. For the base experiments we store all current subdomains in memory because it is quicker; for the deep and wide models we store them as files because the experiments are more memory expensive.




\section{Further Experimental Results --- Combined GNN}
\label{sec:app:further_experiments_bounding}
We will now compare our method with the different baselines in greater depth. We first use different statistics to more accurately explain the performances of the different methods. 

\subsection{Median}
\label{sec:app:median}
Unlike the arithmetic mean, the median is not skewed by outliers.
The randomly picked threshold at which we stop experiments (3600s) has a larger impact on the mean than the median so as long as methods time-out on less than half of all images.

\import{Tables/}{table_all_median.tex}

\newpage
\subsection{Geometric Mean}
\label{sec:app:geometric_mean}
The geometric mean is less skewed than the arithmetic mean but still encapsulates a lot more information than the median, and is arguably the best suited measure to compare the different methods. The geometric mean of a set of numbers $x_1, \dots, x_n$ is defined to be:
$
    \left( \prod^n_{i=1} x_i \right)^{\frac{1}{n}}.
$
As shown in Table \ref{table:geometric_mean} the bounding GNN is about 20\% faster than supergradient ascent and more than 6 times faster than GUROBI and MIPplanet when using the geometric mean. The combined GNN is almost twice as fast as all other baselines.
\import{Tables/}{table_geometric_mean.tex}

\newpage
\subsection{Base Model Experiments} 
\label{sec:app:easy_med_hard_exp}
\begin{figure*}[h!]
	\centering
	\begin{subfigure}{.33\textwidth}
		\centering
		\includegraphics[width=\textwidth]{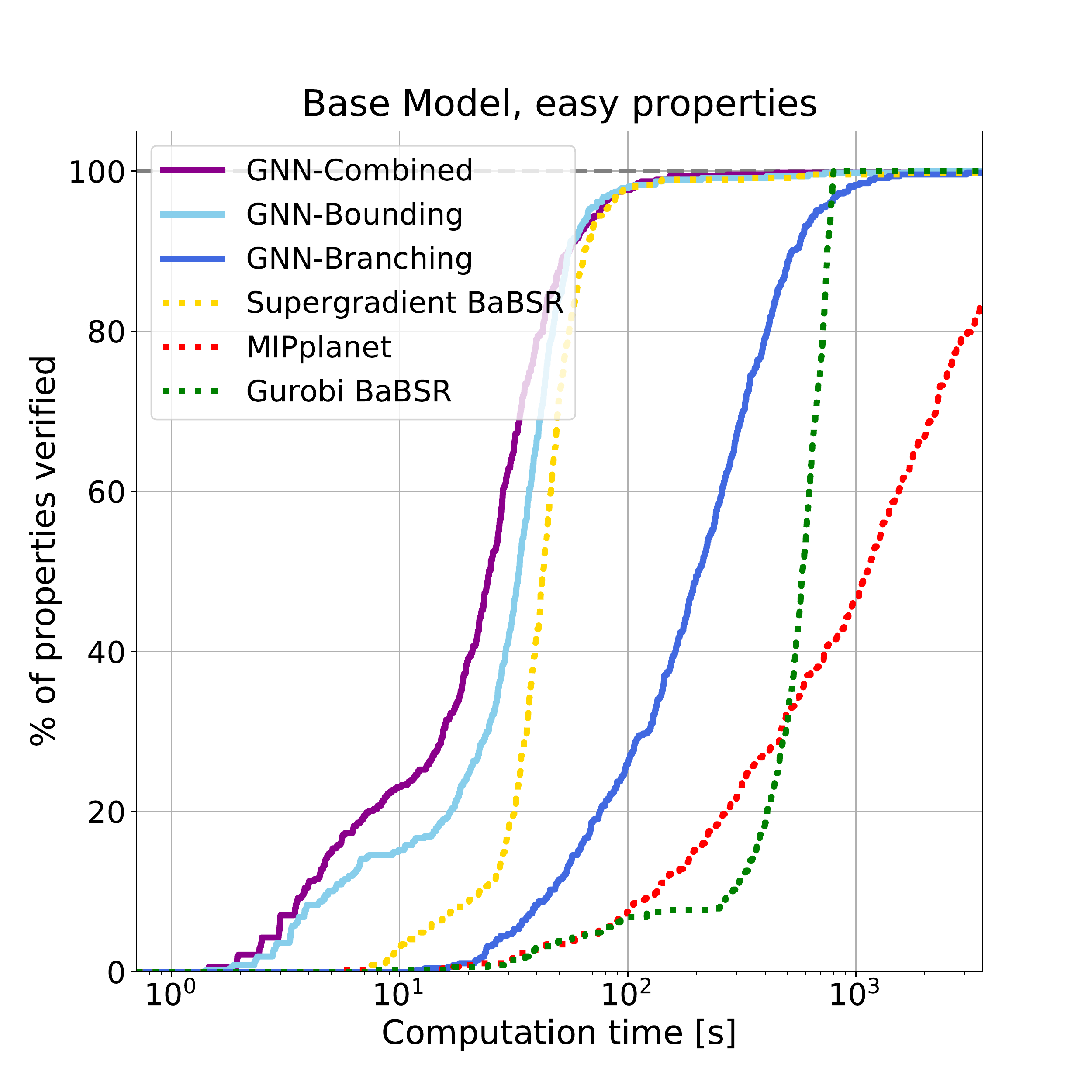}
	\end{subfigure}
	\begin{subfigure}{.33\textwidth}
		\centering
		\includegraphics[width=\textwidth]{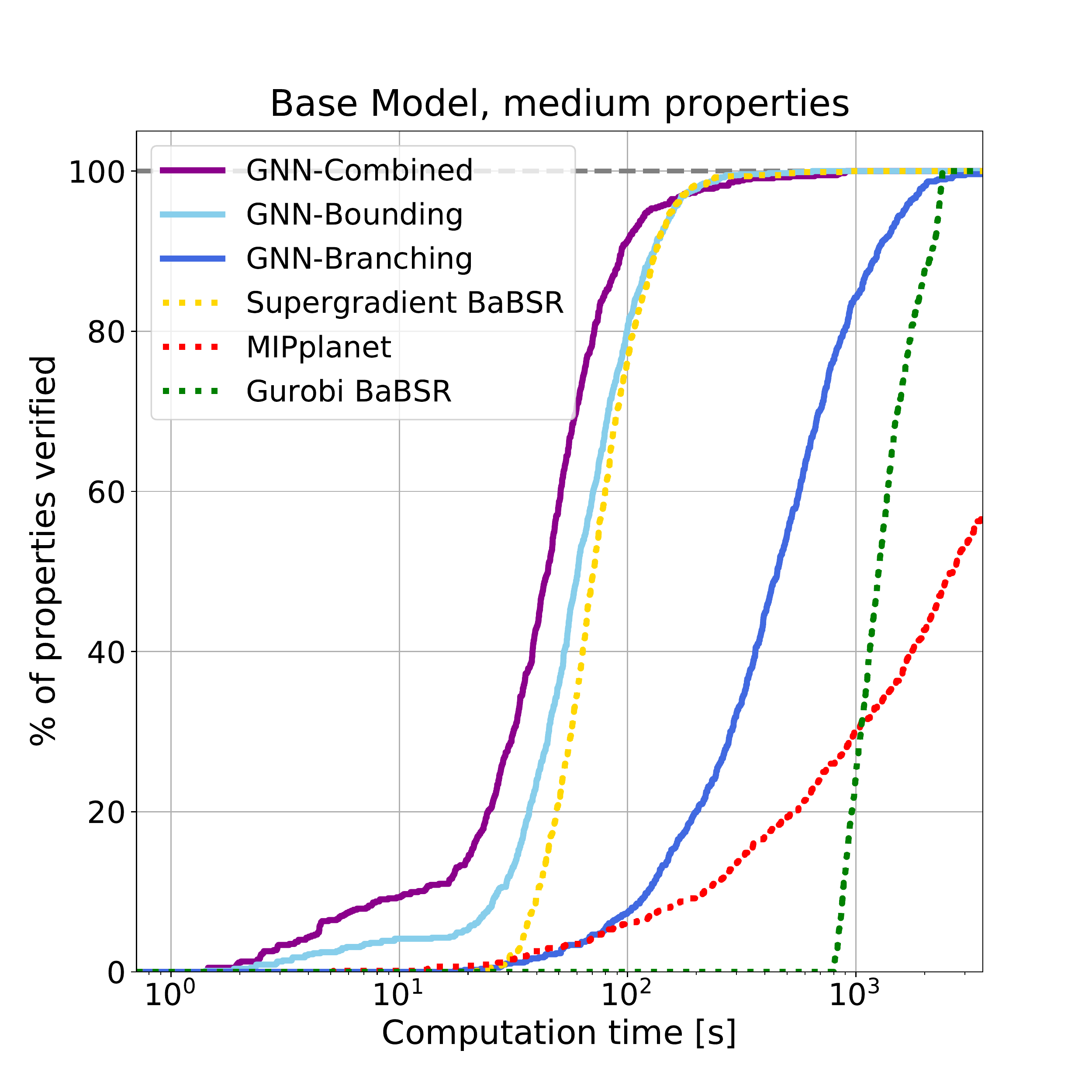}
	\end{subfigure}
	\begin{subfigure}{.33\textwidth}
		\centering
		\includegraphics[width=\textwidth]{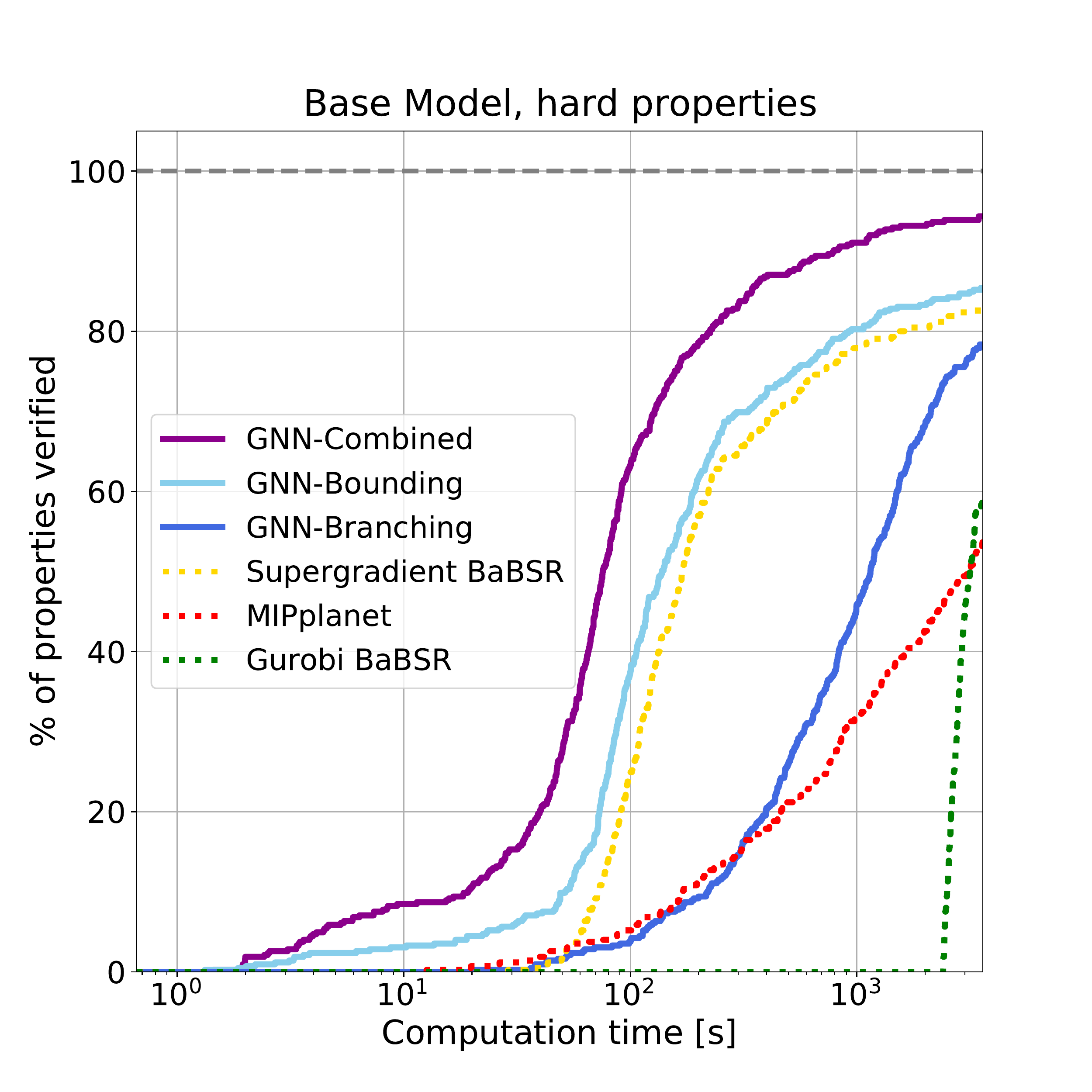}
	\end{subfigure}
	\caption{Cactus plots for the base model, separated into three different graphs based on the difficulty of the properties. We compare the different bounding methods by plotting the percentage of properties that have been solved for any given time.}
	\label{fig:verification-appendix}
\end{figure*}

We now provide a more in-depth analysis of the results on the base model in Figure \ref{fig:verification-appendix}. 
The properties are separated into three sets based on the time $t_i$ it takes GUROBI BaBSR to solve them: ``easy" ($t_i < 800$), ``medium" and a ``hard " ($t_i > 2400$).
Both GNNs are trained on easy properties only, but the combined GNN method beats the baselines on all three types of properties thus showing good generalization performance.

\import{Tables/}{table_all_base_h.tex}

\subsection{Constant Epsilon Experiments}
\label{sec:app:costant_epsilons}
We will now compare our method against the supergradient ascent method, the strongest baselines used in this paper on a new dataset with constant perturbation norms.
We run both methods on 100 properties on the
``\textit{Base}" model. 
We set the epsilon value for all 100 properties to 0.1, 0.15, 0.2, and 0.25 and plot the percentage of properties successfully verified for each experiment. The smaller the perturbation value the more properties are verified by both methods. Our method outperforms the baseline for all four epsilon values demonstrating that the improved performance seen in the experiments above does not depend on whether we use constant or unique $\epsilon$ values. We further note that our method generalizes well to perturbation norms that differ from the ones seen at training time. 
\begin{figure*}[h!]
	\centering
	\begin{subfigure}{.8\textwidth}
		\centering
		\includegraphics[width=\textwidth]{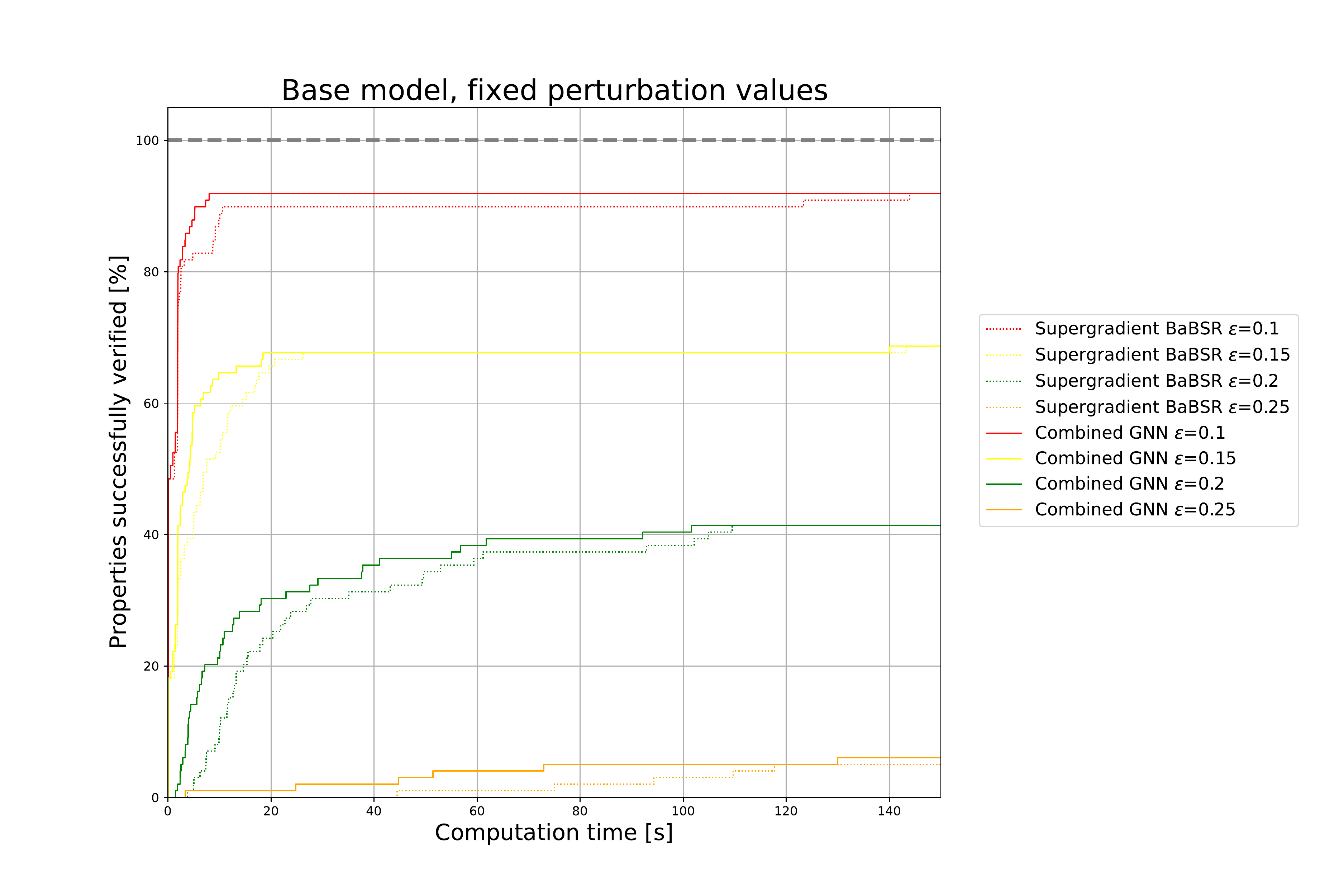}
	\end{subfigure}
	\caption{Cactus plots for the ``\textit{Base}" model and for different constant $\epsilon$ values. We compare our method against Supergradient BaBSR, the strongest baseline, by plotting the percentage of properties that have been solved for any given time.}
	\label{fig:constant_eps}
\end{figure*}

\subsection{Comparison against ERAN}
\label{sec:app:further_experiments_combined}

We will now compare our results against ERAN \citep{singh2020eran}, a state-of-the-art complete verification method. We run ERAN, the baselines described above, and our methods on a subset of the OVAL dataset used in the VNN-COMP competition \citep{VNNComp}.
Our combined GNN method leads to a 50\% reduction in verification time compared to ERAN. While ERAN timesout on fewer properties on the ``\textit{Base}" model, the combined GNN method verifies more properties on the ``\textit{Wide}" model.

\begin{figure*}[h!]
	\centering
	\begin{subfigure}{.33\textwidth}
		\centering
		\includegraphics[width=\textwidth]{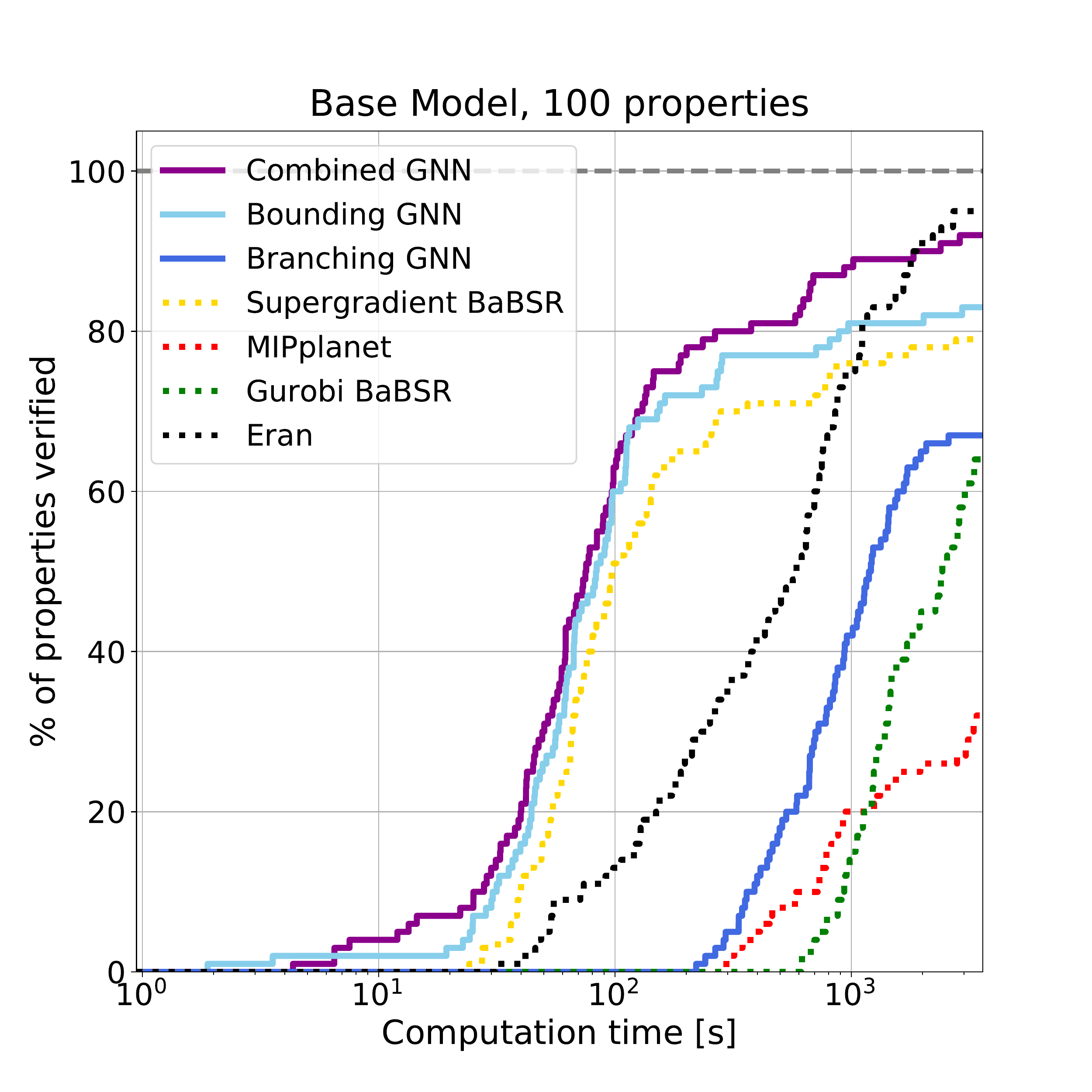}
	\end{subfigure}
	\begin{subfigure}{.33\textwidth}
		\centering
		\includegraphics[width=\textwidth]{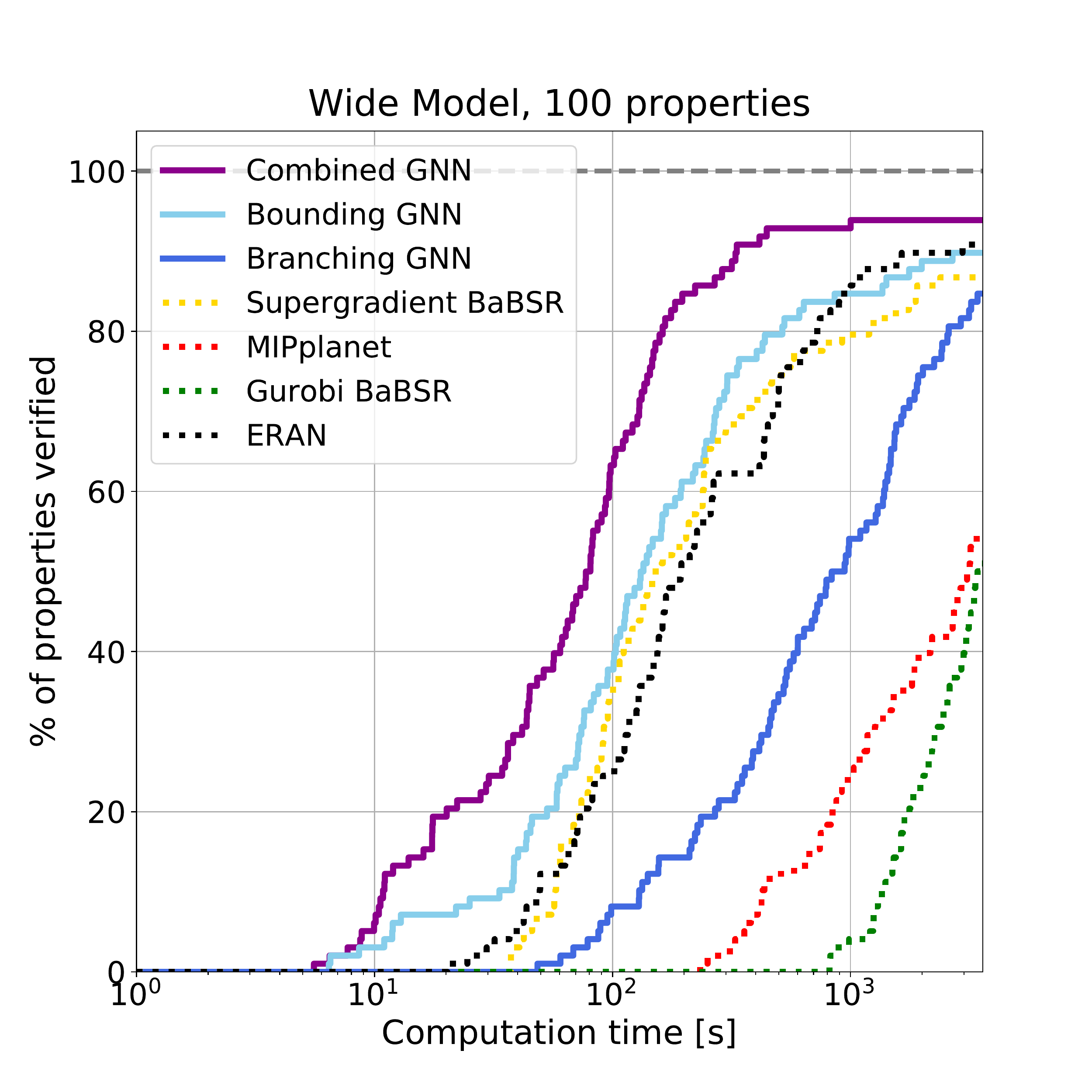}
	\end{subfigure}
	\begin{subfigure}{.33\textwidth}
		\centering
		\includegraphics[width=\textwidth]{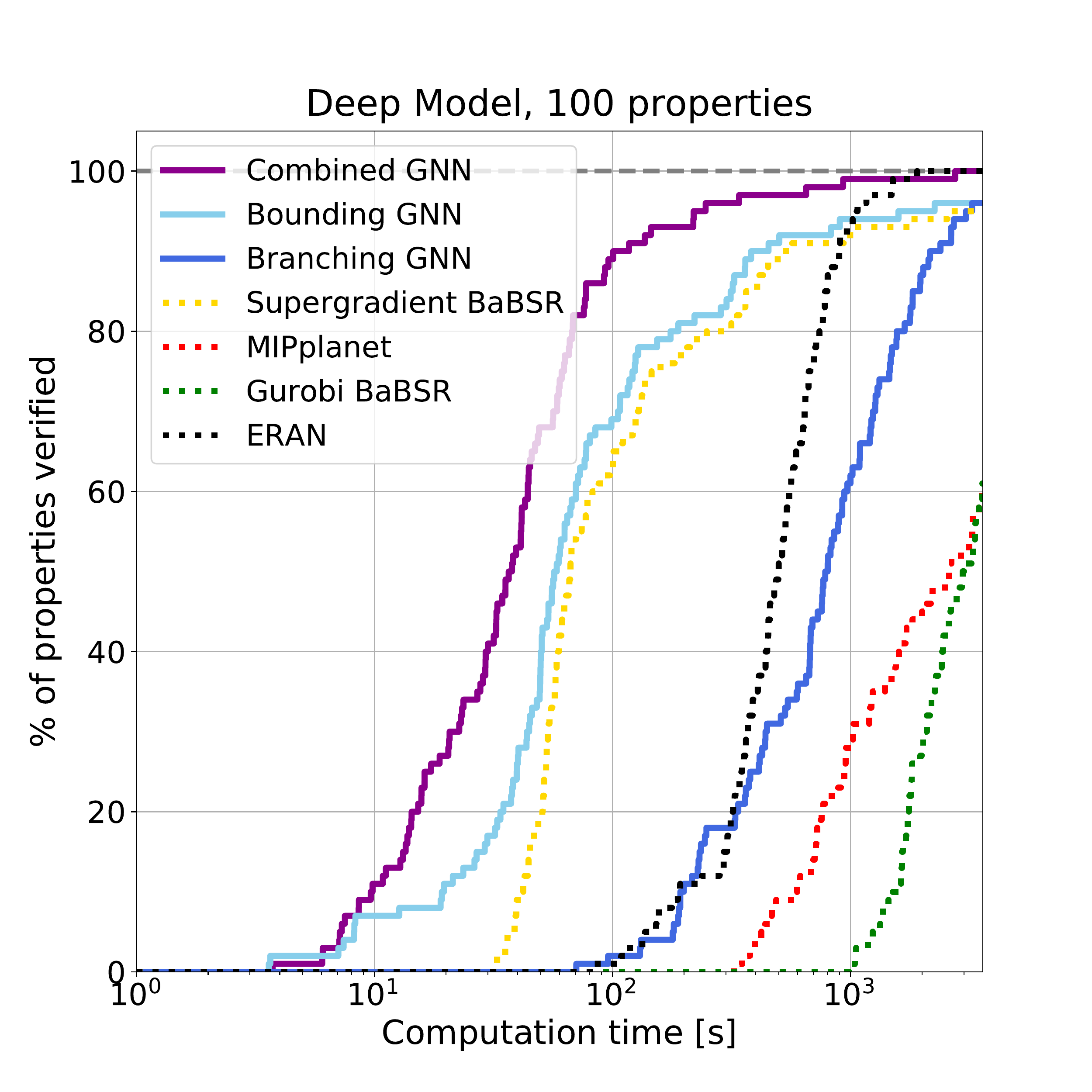}
	\end{subfigure}
	\caption{Cactus plots for the base model, separated into three different graphs based on the difficulty of the properties. We compare the different bounding methods by plotting the percentage of properties that have been solved for any given time.}
	\label{fig:eran}
\end{figure*}
\import{Tables/}{table_eran.tex}

\section{Extending our Method beyond Piece-wise Linearities}\label{sec:app:non-linearities}
The Branch-and-Bound verification method with ReLU splitting is complete as the leaf nodes in the BaB tree form a convex problem. We note that in practice we run our method with a given timeout which makes our method incomplete. As complete neural verification is NP-hard it is unlikely that there exists an efficient algorithm that is complete for all cases when using a short timeout.
Other versions of BaB including BaB with input domain splitting is also complete, as in most cases all the subdomains are small enough to make all ReLU nodes non-ambiguous thus making the problem convex and easy to solve. In the worst case we end up evaluating the network at every single input point which is possible as there is a finite number of input points due to floating-point arithmetic.

Our bounding method can be extended to other non-linearities such as the sigmoid activation or the hyperbolic tangent. \cite{de2021improved} and \cite{zhang2018efficient} provide suitable relaxations for both functions given a pair of lower and upper bounds which allows us to form a dual formulation. A GNN can then estimate better dual directions as it has done for the ReLU case.
Similar to our bounding method, our branching approach can also be extended to work on all piece-wise convex/concave functions. Rather than splitting the ReLU into 2 linear pieces when branching on a particular node we split it into $k$ convex or concave pieces. 
We note that from a theoretical point of view the method is not complete anymore as the composition of a convex and a concave function is not guaranteed to be either convex or concave. However, if we allow repeatedly splitting on the same non-linear activations then our method stays complete.

In practice we don't focus on whether verification methods are complete as all methods become incomplete when using a short timeout. Instead we care about the efficiency of different methods shown empirically.

%% file: Algorithms/BAB.tex
\begin{algorithm}[H]
  \caption{Branch and Bound}\label{alg:bab}
  \small 
  \begin{algorithmic}[1]
        \algrenewcommand\algorithmicindent{1.0em}%
        \Function{BaB}{$\mtt{net}, \mtt{problem}$}
        \State $\mtt{global\_lb} \gets {\mtt{compute\_LB}}(\mtt{net}, \mtt{problem})$ \Comment{global lower bound}
        \State $\mtt{global\_ub} \gets {\mtt{compute\_UB}}(\mtt{net}, \mtt{problem})$ \Comment{global upper bound}
        \State $\mtt{probs}\gets \left[ (\mtt{global\_lb}, \mtt{problem}) \right]$ \Comment{set of all current domains}
        \While{$\mtt{probs}$ is not empty}
        \State $(\_\ , \mtt{prob}) \gets \mtt{pick\_out}(\mtt{probs})$
        \Comment{the pick_out function picks an ambiguous ReLU to split on}
        \State $\left[ \mtt{subprob\_1}, \mtt{subprob\_2} \right] \gets \mtt{split}(\mtt{prob})$
        \For{$i = 1, 2 $}
        \State $\mtt{sub\_lb} \gets {\mtt{compute\_LB}}(\mtt{net}, \mtt{subprob\_i})$
        \State $\mtt{sub\_ub} \gets \mtt{compute\_UB}(\mtt{net}, \mtt{subprob\_i})$
        \If{$\mtt{sub\_ub} < 0$}
        \State \Return SAT
        \Comment{we've found an adversarial example}
        \EndIf
        \If{$\mtt{sub\_lb} < 0$}
        \State $\mtt{probs}.\mtt{append}((\mtt{sub\_lb}, \mtt{subprob\_i}))$
        \EndIf
        \Comment if {$\mtt{sub\_lb} > 0$} then the subdomain gets pruned away
        \EndFor
        \State $\mtt{global\_lb} \gets \min\{ \mtt{lb}\ |\ (\mtt{lb}, \mtt{prob}) \in \mtt{probs}\}$ 
        \Comment{If $\mtt{probs}$ is non-empty then $\mtt{global\_lb}$ is negative}
        \EndWhile
        \State\Return UNSAT
        \Comment{all subproblems have a positive lower bound, therefore $\mtt{global\_lb}$ is positive}
        \EndFunction
      \end{algorithmic}
\end{algorithm}

We will now describe the parallelized version of the BaB algorithm for 
when we use supergradient ascent or the bounding GNN to compute final bounds. Compared to the standard BaB algorithm the number of times the \textit{compute$\_$UB} and \textit{compute$\_$LB} functions are called can be reduced by a factor of $batch\_size$, which can lead to a significant speed up as these two functions tend to be the bottleneck of the algorithm.
\begin{algorithm}[H]
  \caption{Branch and Bound --- parallelized version}\label{alg:bab_parallel}
  \small 
  \begin{algorithmic}[1]
        \algrenewcommand\algorithmicindent{1.0em}%
        \Function{BaB}{$\mtt{net}, \mtt{problem}$}
        \State $\mtt{global\_lb} \gets {\mtt{compute\_LB}}(\mtt{net}, \mtt{problem})$ \Comment{global lower bound}
        \State $\mtt{global\_ub} \gets {\mtt{compute\_UB}}(\mtt{net}, \mtt{problem})$ \Comment{global upper bound}
        \State $\mtt{probs}\gets \left[ (\mtt{global\_lb}, \mtt{problem}) \right]$ \Comment{set of all current domains}
        \While{$\mtt{probs}$ is not empty}
        \State $s = \mtt{min\{batch\_size / 2, len({probs})\}}$
        \State $\mtt{subproblems} = []$
        \For{$i = 1 \dots s $}
        \State $(\_\ , \mtt{prob}) \gets \mtt{pick\_out}(\mtt{probs})$
        \Comment{the pick_out function picks an ambiguous ReLU to split on}
        \State $\left[ \mtt{subprob\_{i_1}}, \mtt{subprob\_{i_2}} \right] \gets
        \mtt{split}(\mtt{prob})$
        \State $\mtt{subproblems} \gets \mtt{subproblems} + \left[ \mtt{subprob\_{i_1}}, \mtt{subprob\_{i_2}} \right]$
        \EndFor
        \State $\left[\mtt{sub\_lb_{1_1}}, \, \mtt{sub\_lb_{1_2}}, \dots, \mtt{sub\_lb_{s_1}}, \, \mtt{sub\_lb_{s_2}} \right] \gets {\mtt{compute\_LBs}}(\mtt{net}, \mtt{subproblems})$
        \State $\left[\mtt{sub\_ub_{1_1}}, \, \mtt{sub\_ub_{1_2}}, \dots, \mtt{sub\_ub_{s_1}}, \, \mtt{sub\_ub_{s_2}} \right] \gets {\mtt{compute\_UBs}}(\mtt{net}, \mtt{subproblems})$
        \For{$i = 1 \dots s $}
        \For{$j = 1, \, 2 $}
        \If{$\mtt{sub\_ub_{i_j}} < 0$}
        \State \Return SAT
        \Comment{we've found an adversarial example}
        \EndIf
        \If{$\mtt{sub\_lb_{i_j}} < 0$}
        \State $\mtt{probs}.\mtt{append}((\mtt{sub\_lb_{i_j}}, \mtt{subprob\_{i_j}}))$
        \EndIf
        \EndFor
        \EndFor
        \State $\mtt{global\_lb} \gets \min\{ \mtt{lb}\ |\ (\mtt{lb}, \mtt{prob}) \in \mtt{probs}\}$ 
        \Comment{If $\mtt{probs}$ is non-empty then $\mtt{global\_lb}$ is negative}
        \EndWhile
        \State\Return UNSAT
        \Comment{all subproblems have a positive lower bound, therefore $\mtt{global\_lb}$ is positive}
        \EndFunction
      \end{algorithmic}
\end{algorithm}

%% file: Figures/relu/interceptKW.tex
\begin{tikzpicture}
  \tikzset{dummy/.style= {inner sep=0, outer sep=0}}
  \tikzset{cross/.style={cross out, draw,
      minimum size=3*(#1-\pgflinewidth),
      inner sep=0pt, outer sep=0pt,
      thick}}

  \draw[-, ultra thick](0, 0) to (1, 0) to (2, 1);
  \draw[dashed, ultra thick](0, 0) to (2, 1);

  \draw[dashed](0, -0.5) to (0, 1.5);
  \draw[dashed](2, -0.5) to (2, 1.5);

  \draw[fill=white!80!green](0, 0) -- (1,0) -- (2, 1);
  \draw[-, ultra thick, red] (1,0) to (1, 0.5);

  \node[cross=2pt] at (0, 0) {};
  \node[dummy](lb-lab) at (-0.3, -0.3) {$l_{i[j]}$};
  \node[cross=2pt] at (2, 0) {};
  \node[dummy](ub-lab) at (2.35, -0.3) {$u_{i[j]}$};

  \draw[-latex](-0.5,0) to (3, 0);
  \node[dummy](x-label) at (3.3, 0) {$\hat{x}_{i[j]}$};
  \draw[-latex](1,-0.5) to (1, 1.5);
  \node[dummy](x-label) at (1, 1.8) {$x_{i[j]}$};
      
\end{tikzpicture}

%% file: Figures/relu/relusplitN.tex
\begin{tikzpicture}
  \tikzset{dummy/.style= {inner sep=0, outer sep=0}}
  \tikzset{cross/.style={cross out, draw,
      minimum size=3*(#1-\pgflinewidth),
      inner sep=0pt, outer sep=0pt,
      thick}}
  \tikzset{arrow/.style = {thick, double,  ->, >=stealth}}  

  \draw[-, ultra thick](0, 0) to (1, 0) to (2, 1);
  \draw[-, ultra thick, red](0, 0) to (2, 1);    

  \draw[dashed](0, -0.5) to (0, 1.5);
  \draw[dashed](2, -0.5) to (2, 1.5);

  \draw[fill=white!80!green](0, 0) -- (1,0) -- (2, 1);

  \node[cross=2pt] at (0, 0) {};
  \node[dummy](lb-lab) at (-0.3, -0.3) {$l_{i[j]}$};
  \node[cross=2pt] at (2, 0) {};
  \node[dummy](ub-lab) at (2.35, -0.3) {$u_{i[j]}$};

  \draw[-latex](-0.5,0) to (3, 0);
  \node[dummy](x-label) at (3.3, 0) {$\hat{x}_{i[j]}$};
  \draw[-latex](1,-0.5) to (1, 1.5);
  \node[dummy](x-label) at (1, 1.8) {$x_{i[j]}$};
 \node[dummy](ambi) at (1, -1) {(a) Ambiguous Node};
      
      
  \draw[-, ultra thick, red](5, 0) to (5.8, 0);

  \draw[dashed](5, -0.5) to (5, 1.5);
  \draw[dashed](5.8, -0.5) to (5.8, 1.5);

  \node[cross=2pt] at (5, 0) {};
  \node[dummy](lb-lab) at (4.7, -0.3) {$l_{i[j]}$};
  \node[cross=2pt] at (5.8, 0) {};
  \node[dummy](ub-lab) at (6.15, -0.3) {$u_{i[j]}$};

  \draw[-latex](4.5,0) to (7, 0);
  \node[dummy](x-label) at (7.3, 0) {$\hat{x}_{i[j]}$};
  \draw[-latex](6,-0.5) to (6, 1.5);
  \node[dummy](x-label) at (6, 1.8) {$x_{i[j]}$};   
  \node[dummy](blocking) at (6, -1) {(b) Complete Blocking};
  

  \draw[-, ultra thick, red] (9.7, 0.2) to (10.5, 1);

  \draw[dashed](9.7, -0.5) to (9.7, 1.5);
  \draw[dashed](10.5, -0.5) to (10.5, 1.5);

  \node[cross=2pt] at (9.7, 0) {};
  \node[dummy](lb-lab) at (9.4, -0.3) {$l_{i[j]}$};
  \node[cross=2pt] at (10.5, 0) {};
  \node[dummy](ub-lab) at (10.85, -0.3) {$u_{i[j]}$};

  \draw[-latex](8.5,0) to (11, 0);
  \node[dummy](x-label) at (11.3, 0) {$\hat{x}_{i[j]}$};
  \draw[-latex](9.5,-0.5) to (9.5, 1.5);
  \node[dummy](x-label) at (9.5, 1.8) {$x_{i[j]}$};
  \node[dummy](passing) at (10, -1) {(c) Complete Passing}; 

\end{tikzpicture}

%% file: Algorithms/Supergradient.tex
\begin{minipage}{.90\textwidth}
	\begin{algorithm}[H]
		\caption{Supergradient method}\label{alg:subg}
		\begin{algorithmic}[1]
			\algrenewcommand\algorithmicindent{1.0em}%
			\Function{superg\_compute\_bounds}{$\{W_k, \mathbf{b}_k, \mathbf{l}_k, \mathbf{u}_k\}_{k=1..n}$}
			\State Initialise dual variables $\rrho^0$ using the duals of the parent domain or the algo of \citet{Wong2018}
			\For{$\mtt{nb\_iterations}$}\label{alg:inner_loopstart}
			\State $\zhat^*, \, \zhat^*_A \, \zhat^*_B \leftarrow$ inner minimization as proposed by \citet{bunel2020lagrangian} \label{alg:inner_min}
			\State Compute supergradient using $\nabla_{\rrho} q(\rrho^t) = \zhat_B^* - \zhat_A^* $ 
			\State $\rrho^{t+1} \leftarrow$ Adam's update rule \cite{kingma2014adam} \label{alg:adam-update}
			\EndFor\label{alg:inner_loopend}
			\State\Return $q(\rrho)$
			\EndFunction
		\end{algorithmic}
	\end{algorithm}
\end{minipage}

%% file: Tables/table_all_median.tex
 \sisetup{detect-weight=true,detect-inline-weight=math,detect-mode=true}
\begin{table*}[h]
	\centering
	\scriptsize
	\setlength{\tabcolsep}{4pt}
	\aboverulesep = 0.1mm  
	\belowrulesep = 0.2mm  
	
    
	\begin{adjustbox}{center}
		\begin{tabular}{
				l
				S[table-format=4.3]
				S[table-format=4.3]
				S[table-format=4.3]
				S[table-format=4.3]
				S[table-format=4.3]
				S[table-format=4.3]
				S[table-format=4.3]
				S[table-format=4.3]
				S[table-format=4.3]
			}
			& \multicolumn{3}{ c }{Base} & \multicolumn{3}{ c }{Wide} & \multicolumn{3}{ c }{Deep} \\
			\toprule
			
			\multicolumn{1}{ c }{Method} &
			\multicolumn{1}{ c }{time(s)} &
			\multicolumn{1}{ c }{subdomains} &
			\multicolumn{1}{ c }{$\%$Timeout} &
			\multicolumn{1}{ c }{time(s)} &
			\multicolumn{1}{ c }{subdomains} &
			\multicolumn{1}{ c }{$\%$Timeout} &
			\multicolumn{1}{ c }{time(s)} &
			\multicolumn{1}{ c }{subdomains} &
			\multicolumn{1}{ c }{$\%$Timeout} \\
			
			\cmidrule(lr){1-1} \cmidrule(lr){2-4} \cmidrule(lr){5-7} \cmidrule(lr){8-10}
			
			\multicolumn{1}{ c }{\textsc{Gurobi BaBSR}}
            &1231.39
            &1034.00
            &10.57
			
            &3600.00
            &704.00
            &50.17
			
            &3600.00
            &364.00
            &54.00 \\
			
			\multicolumn{1}{ c }{\textsc{MIPplanet}}
            &2045.19
            &
            &36.40
			
            &3600.00
            &
            &79.54
			
            &3600.00
            &
            &73.60 \\
			
			\multicolumn{1}{ c }{\textsc{Adam}}
            &71.04
            &3962.00
            &4.68
			
            &95.14
            &3452.00
            &13.20
			
            &62.72
            &1484.00
            &4.00 \\
			
			\multicolumn{1}{ c }{\textsc{Branching-GNN}}
            &419.73
            &396.00
            &5.77
			
            &1480.08
            &330.00
            &20.79
			
            &1822.88
            &175.00
            &20.80 \\

			\multicolumn{1}{ c }{\textsc{Bounding-GNN}}
            &64.42
            &2835.00
            &2.94
			
            &92.42
            &1910.00
            &10.23
			
            &55.45
            &1162.00
            &2.80 \\
			
			\multicolumn{1}{ c }{\textsc{Combined-GNN}}
            &\B53.63
            &1506.00
            &\B1.44
			
            &\B72.50
            &1442.00
            &\B5.61
			
            &\B45.65
            &784.00
            &\B0.80 \\
			\bottomrule
			
		\end{tabular}
	\end{adjustbox}
	\caption{\small We compare average (median) solving time, average (median) number of subdomains solved, and the percentage of properties that the methods time out on when using a cut-off time of 3600s. The best performing method for each subcategory is highlighted in bold.}
	\label{table:all_median}
\end{table*}

%% file: Tables/table_geometric_mean.tex
 \sisetup{detect-weight=true,detect-inline-weight=math,detect-mode=true}
\begin{table*}[h]
	\centering
	\scriptsize
	\setlength{\tabcolsep}{4pt}
	\aboverulesep = 0.1mm  
	\belowrulesep = 0.2mm  
	
    
	\begin{adjustbox}{center}
		\begin{tabular}{
				l
				S[table-format=4.3]
				S[table-format=4.3]
				S[table-format=4.3]
				S[table-format=4.3]
				S[table-format=4.3]
				S[table-format=4.3]
				S[table-format=4.3]
				S[table-format=4.3]
				S[table-format=4.3]
			}
			& \multicolumn{3}{ c }{Base} & \multicolumn{3}{ c }{Wide} & \multicolumn{3}{ c }{Deep} \\
			\toprule
			
			\multicolumn{1}{ c }{Method} &
			\multicolumn{1}{ c }{time(s)} &
			\multicolumn{1}{ c }{subdomains} &
			\multicolumn{1}{ c }{$\%$Timeout} &
			\multicolumn{1}{ c }{time(s)} &
			\multicolumn{1}{ c }{subdomains} &
			\multicolumn{1}{ c }{$\%$Timeout} &
			\multicolumn{1}{ c }{time(s)} &
			\multicolumn{1}{ c }{subdomains} &
			\multicolumn{1}{ c }{$\%$Timeout} \\
			
			\cmidrule(lr){1-1} \cmidrule(lr){2-4} \cmidrule(lr){5-7} \cmidrule(lr){8-10}
			
			\multicolumn{1}{ c }{\textsc{Gurobi BaBSR}}
            &1222.41
            &1013.63
            &10.57

            &2724.47
            &698.59
            &50.17

            &2869.54
            &366.85
            &54.00 \\
			
			\multicolumn{1}{ c }{\textsc{MIPplanet}}
            &1208.08
            &
            &36.40

            &2611.10
            &
            &79.54

            &2502.48
            &
            &73.60 \\
			
			\multicolumn{1}{ c }{\textsc{Adam}}
            &92.56
            &4142.24
            &4.68

            &163.99
            &3954.07
            &13.20

            &90.35
            &1690.18
            &4.00 \\
			
			\multicolumn{1}{ c }{\textsc{Branching-GNN}}
            &408.52
            &373.36
            &5.77

            &1258.83
            &301.92
            &20.79

            &1423.57
            &152.80
            &20.80 \\

			\multicolumn{1}{ c }{\textsc{Bounding-GNN}}
            &70.28
            &2734.70
            &2.94

            &130.13
            &2183.76
            &10.23

            &66.36
            &1133.28
            &2.80 \\

            \multicolumn{1}{ c }{\textsc{Combined-GNN}}
            &\B54.91
            &1296.13
            &\B1.44

            &\B80.61
            &1176.62
            &\B5.61

            &\B47.39
            &641.62
            &\B0.80 \\
			\bottomrule
			
		\end{tabular}
	\end{adjustbox}
	\caption{\small We use the geometric mean to compare solving time, number of subdomains solved, and the percentage of properties that the methods time out on when using a cut-off time of 3600s. The best performing method for each subcategory is highlighted in bold.}
	\label{table:geometric_mean}
\end{table*}

%% file: Tables/table_all_base_h.tex
 \sisetup{detect-weight=true,detect-inline-weight=math,detect-mode=true}
\begin{table*}[h]
	\centering
	\scriptsize
	\setlength{\tabcolsep}{4pt}
	\aboverulesep = 0.1mm  
	\belowrulesep = 0.2mm  
	

	\begin{adjustbox}{center}
		\begin{tabular}{
				l
				S[table-format=4.3]
				S[table-format=4.3]
				S[table-format=4.3]
				S[table-format=4.3]
				S[table-format=4.3]
				S[table-format=4.3]
				S[table-format=4.3]
				S[table-format=4.3]
				S[table-format=4.3]
			}
			& \multicolumn{3}{ c }{Easy} & \multicolumn{3}{ c }{Med} & \multicolumn{3}{ c }{Hard} \\
			\toprule
			
			\multicolumn{1}{ c }{Method} &
			\multicolumn{1}{ c }{time(s)} &
			\multicolumn{1}{ c }{subdomains} &
			\multicolumn{1}{ c }{$\%$Timeout} &
			\multicolumn{1}{ c }{time(s)} &
			\multicolumn{1}{ c }{subdomains} &
			\multicolumn{1}{ c }{$\%$Timeout} &
			\multicolumn{1}{ c }{time(s)} &
			\multicolumn{1}{ c }{subdomains} &
			\multicolumn{1}{ c }{$\%$Timeout} \\
			
			\cmidrule(lr){1-1} \cmidrule(lr){2-4} \cmidrule(lr){5-7} \cmidrule(lr){8-10}
			
			\multicolumn{1}{ c }{\textsc{Gurobi BaBSR}}
			&550.48
			&580.43
			&0.00
			
			&1374.32
			&1408.75
			&0.00
			
			&3129.08
			&2551.63
			&42.41 \\
			
			\multicolumn{1}{ c }{\textsc{MIPplanet}}
			&1499.35
			& 
			&16.49
			
			&2240.92
			& 
			&42.95

			&2255.53
			& 
			&46.35 \\
			
			\multicolumn{1}{ c }{\textsc{Supergradient}}
			&59.40
			&2528.48
			&0.21
			
			&87.71
			&5289.22
			&\B0.00
			
			&923.45
			&22103.66
			&18.12 \\
			
			\multicolumn{1}{ c }{\textsc{GNN_Branching}}
			&272.69
			&285.68
			&0.21
			
			&592.12
			&583.21
			& 0.39
			
			&1573.16
			&1098.28
			&21.65 \\

			\multicolumn{1}{ c }{\textsc{GNN_Bounding}}
			&\B46.70
			&2201.70
			&\B0.00
			
			&78.85
			&4108.27
			&\B 0.00
			
			&665.64
			&16442.30
			&11.53 \\

			\multicolumn{1}{ c }{\textsc{GNN_Combined}}
			&50.61
			&1188.36
			&0.21
			
			&\B68.52
			&2419.44
			&\B 0.00
			
			&\B380.15
			&8488.50
			&\B5.41 \\

			\bottomrule
			
		\end{tabular}
	\end{adjustbox}
	\caption{\small We compare average (mean) solving time, average number of subdomains solved, and the percentage of properties solved for easy, medium, and hard properties on the base model. The best performing method for each subcategory is highlighted in bold (note that by definition Gurobi BaBSR doesn't time out on easy and med experiments).}.
	\label{table:combined_base_exp[h]}
\end{table*}

%% file: Tables/table_eran.tex
 \sisetup{detect-weight=true,detect-inline-weight=math,detect-mode=true}
\begin{table*}[h]
	\centering
	\scriptsize
	\setlength{\tabcolsep}{4pt}
	\aboverulesep = 0.1mm  
	\belowrulesep = 0.2mm  
	
    
	\begin{adjustbox}{center}
		\begin{tabular}{
				l
				S[table-format=4.3]
				S[table-format=4.3]
				S[table-format=4.3]
				S[table-format=4.3]
				S[table-format=4.3]
				S[table-format=4.3]
				S[table-format=4.3]
				S[table-format=4.3]
				S[table-format=4.3]
			}
			& \multicolumn{3}{ c }{Base} & \multicolumn{3}{ c }{Wide} & \multicolumn{3}{ c }{Deep} \\
			\toprule
			
			\multicolumn{1}{ c }{Method} &
			\multicolumn{1}{ c }{time(s)} &
			\multicolumn{1}{ c }{subdomains} &
			\multicolumn{1}{ c }{$\%$Timeout} &
			\multicolumn{1}{ c }{time(s)} &
			\multicolumn{1}{ c }{subdomains} &
			\multicolumn{1}{ c }{$\%$Timeout} &
			\multicolumn{1}{ c }{time(s)} &
			\multicolumn{1}{ c }{subdomains} &
			\multicolumn{1}{ c }{$\%$Timeout} \\
			
			\cmidrule(lr){1-1} \cmidrule(lr){2-4} \cmidrule(lr){5-7} \cmidrule(lr){8-10}
			
			\multicolumn{1}{ c }{\textsc{Gurobi BaBSR}}
            &2367.15
            &1313.41
            &36.00
			
            &2860.04
            &1072.58
            &48.98
			
            &2750.12
            &441.26
            &39.00 \\
			
			\multicolumn{1}{ c }{\textsc{MIPplanet}}
            &2849.10
            &
            &68.00
			
            &2423.04
            &
            &45.92
			
            &2302.79
            &
            &40.00 \\
			
			\multicolumn{1}{ c }{\textsc{Supergradient}}
            &922.35
            &13738.41
            &21.00
			
            &743.50
            &13184.12
            &13.27
			
            &348.59
            &3672.74
            &5.00 \\
			
			\multicolumn{1}{ c }{\textsc{ERAN}}
            &805.89
            &
            &\B5.00
			
            &635.48
            &
            &9.18
			
            &545.72
            &
            &\B0.00 \\
			
			\multicolumn{1}{ c }{\textsc{GNN-branching}}
            &1794.88
            &734.93
            &33.00
			
            &1360.97
            &397.77
            &15.31
			
            &1055.06
            &128.14
            &4.00\\

			\multicolumn{1}{ c }{\textsc{GNN-bounding}}
            &758.28
            &9875.30
            &17.00
			
            &599.50
            &9371.14
            &10.20
			
            &285.25
            &3392.71
            &4.00\\

			\multicolumn{1}{ c }{\textsc{GNN-combined}}
            &\B479.61
            &10072.93
            &8.00
			
            &\B317.16
            &2515.74
            &\B6.12
			
            &\B90.78
            &1471.12
            &\B0.00\\

			\bottomrule
			
		\end{tabular}
	\end{adjustbox}
	\caption{\small We compare average (mean) solving time, average number of subdomains solved, and the percentage of properties that the methods time out on when using a cut-off time of 3600s. The best performing method for each subcategory is highlighted in bold.}
	\label{table:eran}
\end{table*}

%% file: main.bbl
\begin{thebibliography}{43}
\providecommand{\natexlab}[1]{#1}
\providecommand{\url}[1]{\texttt{#1}}
\expandafter\ifx\csname urlstyle\endcsname\relax
  \providecommand{\doi}[1]{doi: #1}\else
  \providecommand{\doi}{doi: \begingroup \urlstyle{rm}\Url}\fi

\bibitem[Alvarez et~al.(2017)Alvarez, Louveaux, and
  Wehenkel]{alvarez2017machine}
Alejandro~Marcos Alvarez, Quentin Louveaux, and Louis Wehenkel.
\newblock A machine learning-based approximation of strong branching.
\newblock \emph{INFORMS Journal on Computing}, 29\penalty0 (1):\penalty0
  185--195, 2017.

\bibitem[Anderson et~al.(2019{\natexlab{a}})Anderson, Pailoor, Dillig, and
  Chaudhuri]{Anderson2019}
Greg Anderson, Shankara Pailoor, Isil Dillig, and Swarat. Chaudhuri.
\newblock Optimization and abstraction: a synergistic approach for analyzing
  neural network robustness.
\newblock \emph{ACM SIGPLAN Conference on Programming Language Design and
  Implementation}, 2019{\natexlab{a}}.

\bibitem[Anderson et~al.(2019{\natexlab{b}})Anderson, Huchette, Tjandraatmadja,
  and Vielma]{anderson2019strong}
Ross Anderson, Joey Huchette, Christian Tjandraatmadja, and Juan~Pablo Vielma.
\newblock Strong mixed-integer programming formulations for trained neural
  networks.
\newblock In \emph{International Conference on Integer Programming and
  Combinatorial Optimization}, pages 27--42. Springer, 2019{\natexlab{b}}.

\bibitem[Balunovic and Vechev(2020)]{Balunovic2020Adversarial}
Mislav Balunovic and Martin Vechev.
\newblock Adversarial training and provable defenses: Bridging the gap.
\newblock In \emph{International Conference on Learning Representations}, 2020.

\bibitem[Bello et~al.(2017a)Bello, Pham, Le, Norouzi, and
  Bengio]{bello2016neural}
Irwan Bello, Hieu Pham, Quoc~V Le, Mohammad Norouzi, and Samy Bengio.
\newblock Neural combinatorial optimization with reinforcement learning.
\newblock \emph{The International Conference on Learning Representations
  Workshop}, 2017a.

\bibitem[Bojarski et~al.(2016)Bojarski, Del~Testa, Dworakowski, Firner, Flepp,
  Goyal, Jackel, Monfort, Muller, Zhang, et~al.]{bojarski2016end}
Mariusz Bojarski, Davide Del~Testa, Daniel Dworakowski, Bernhard Firner, Beat
  Flepp, Prasoon Goyal, Lawrence~D Jackel, Mathew Monfort, Urs Muller, Jiakai
  Zhang, et~al.
\newblock End to end learning for self-driving cars.
\newblock \emph{arXiv preprint arXiv:1604.07316}, 2016.

\bibitem[Bunel et~al.(2018{\natexlab{a}})Bunel, Turkaslan, Torr, Kohli, and
  Kumar]{rudy2018}
Rudy Bunel, Ilker Turkaslan, Philip~H.S Torr, Pushmeet Kohli, and M.~Pawan
  Kumar.
\newblock A unified view of piecewise linear neural network verification.
\newblock \emph{Advances in Neural Information Processing Systems}, pages
  4790--4799, 2018{\natexlab{a}}.

\bibitem[Bunel et~al.(2020{\natexlab{a}})Bunel, De~Palma, Desmaison, Dvijotham,
  Kohli, Torr, and Kumar]{bunel2020lagrangian}
Rudy Bunel, Alessandro De~Palma, Alban Desmaison, Krishnamurthy Dvijotham,
  Pushmeet Kohli, Philip~HS Torr, and M~Pawan Kumar.
\newblock Lagrangian decomposition for neural network verification.
\newblock \emph{arXiv preprint arXiv:2002.10410}, 2020{\natexlab{a}}.

\bibitem[Bunel et~al.(2020{\natexlab{b}})Bunel, Lu, Turkaslan, Kohli, Torr, and
  Kumar]{journal2019}
Rudy Bunel, Jingyue Lu, Ilker Turkaslan, P~Kohli, P~Torr, and M.~Pawan Kumar.
\newblock Branch and bound for piecewise linear neural network verification.
\newblock \emph{Journal of Machine Learning Research}, 21\penalty0 (2020),
  2020{\natexlab{b}}.

\bibitem[Bunel et~al.(2018{\natexlab{b}})Bunel, Turkaslan, Torr, Kohli, and
  Kumar]{bunel2018unified}
Rudy~R Bunel, Ilker Turkaslan, Philip Torr, Pushmeet Kohli, and M~Pawan Kumar.
\newblock A unified view of piecewise linear neural network verification.
\newblock In \emph{Advances in Neural Information Processing Systems}, pages
  4790--4799, 2018{\natexlab{b}}.

\bibitem[Dai et~al.(2017)Dai, Khalil, Zhang, Dilkina, and Song]{Dai2017}
Hanjun Dai, Elias~B. Khalil, Yuyu Zhang, Bistra Dilkina, and Le~Song.
\newblock Learning combinatorial optimization algorithms over graphs.
\newblock \emph{Conference on Neural Information Processing Systems}, 2017.

\bibitem[De~Palma et~al.(2021)De~Palma, Bunel, Desmaison, Dvijotham, Kohli,
  Torr, and Kumar]{de2021improved}
Alessandro De~Palma, Rudy Bunel, Alban Desmaison, Krishnamurthy Dvijotham,
  Pushmeet Kohli, Philip~HS Torr, and M~Pawan Kumar.
\newblock Improved branch and bound for neural network verification via
  lagrangian decomposition.
\newblock \emph{arXiv preprint arXiv:2104.06718}, 2021.

\bibitem[Dvijotham et~al.(2018{\natexlab{a}})Dvijotham, Gowal, Stanforth,
  Arandjelovic, O'Donoghue, Uesato, and Kohli]{dvijotham2018training}
Krishnamurthy Dvijotham, Sven Gowal, Robert Stanforth, Relja Arandjelovic,
  Brendan O'Donoghue, Jonathan Uesato, and Pushmeet Kohli.
\newblock Training verified learners with learned verifiers.
\newblock \emph{arXiv preprint arXiv:1805.10265}, 2018{\natexlab{a}}.

\bibitem[Dvijotham et~al.(2018{\natexlab{b}})Dvijotham, Stanforth, Gowal, Mann,
  and Kohli]{dvijotham2018dual}
Krishnamurthy Dvijotham, Robert Stanforth, Sven Gowal, Timothy~A Mann, and
  Pushmeet Kohli.
\newblock A dual approach to scalable verification of deep networks.
\newblock In \emph{Conference on Uncertainty in Artificial Intelligence}, pages
  550--559, 2018{\natexlab{b}}.

\bibitem[Ehlers(2017{\natexlab{a}})]{Ehlers2017}
Ruediger Ehlers.
\newblock Formal verification of piece-wise linear feed-forward neural
  networks.
\newblock \emph{Automated Technology for Verification and Analysis},
  2017{\natexlab{a}}.

\bibitem[Ehlers(2017{\natexlab{b}})]{ehlers2017formal}
Ruediger Ehlers.
\newblock Formal verification of piece-wise linear feed-forward neural
  networks.
\newblock In \emph{International Symposium on Automated Technology for
  Verification and Analysis}, pages 269--286. Springer, 2017{\natexlab{b}}.

\bibitem[Gasse et~al.(2019)Gasse, Ch{\'e}telat, Ferroni, Charlin, and
  Lodi]{gasse2019exact}
Maxime Gasse, Didier Ch{\'e}telat, Nicola Ferroni, Laurent Charlin, and Andrea
  Lodi.
\newblock Exact combinatorial optimization with graph convolutional neural
  networks.
\newblock In \emph{Advances in Neural Information Processing Systems}, pages
  15554--15566, 2019.

\bibitem[Goodfellow et~al.(2015)Goodfellow, Shlens, and
  Szegedy]{goodfellow2014explaining}
Ian~J Goodfellow, Jonathon Shlens, and Christian Szegedy.
\newblock Explaining and harnessing adversarial examples.
\newblock \emph{The International Conference on Learning Representations},
  2015.

\bibitem[Gowal et~al.(2018)Gowal, Dvijotham, Stanforth, Bunel, Qin, Uesato,
  Mann, and Kohli]{gowal2018effectiveness}
Sven Gowal, Krishnamurthy Dvijotham, Robert Stanforth, Rudy Bunel, Chongli Qin,
  Jonathan Uesato, Timothy Mann, and Pushmeet Kohli.
\newblock On the effectiveness of interval bound propagation for training
  verifiably robust models.
\newblock \emph{arXiv preprint arXiv:1810.12715}, 2018.

\bibitem[Gowal et~al.(2019)Gowal, Dvijotham, Stanforth, Mann, and
  Kohli]{gowal2019dual}
Sven Gowal, Krishnamurthy Dvijotham, Robert Stanforth, Timothy Mann, and
  Pushmeet Kohli.
\newblock A dual approach to verify and train deep networks.
\newblock In \emph{Proceedings of the 28th International Joint Conference on
  Artificial Intelligence}, pages 6156--6160. AAAI Press, 2019.

\bibitem[Guignard and Kim(1987)]{guignard1987lagrangean}
Monique Guignard and Siwhan Kim.
\newblock Lagrangean decomposition: A model yielding stronger lagrangean
  bounds.
\newblock \emph{Mathematical programming}, 39\penalty0 (2):\penalty0 215--228,
  1987.

\bibitem[Gurobi~Optimization(2020)]{gurobi}
LLC Gurobi~Optimization.
\newblock Gurobi optimizer reference manual, 2020.
\newblock URL \url{http://www.gurobi.com}.

\bibitem[Hansknecht et~al.(2018)Hansknecht, Joormann, and
  Stiller]{hansknecht2018cuts}
Christoph Hansknecht, Imke Joormann, and Sebastian Stiller.
\newblock Cuts, primal heuristics, and learning to branch for the
  time-dependent traveling salesman problem.
\newblock \emph{arXiv preprint arXiv:1805.01415}, 2018.

\bibitem[Katz et~al.(2017{\natexlab{a}})Katz, Barrett, Dill, Julian, and
  Kochenderfer]{Katz2017}
Guy Katz, Clark Barrett, David Dill, Kyle Julian, and Mykel Kochenderfer.
\newblock Reluplex: An efficient smt solver for verifying deep neural networks.
\newblock \emph{International Conference on Computer Aided Verification},
  2017{\natexlab{a}}.

\bibitem[Katz et~al.(2017{\natexlab{b}})Katz, Barrett, Dill, Julian, and
  Kochenderfer]{katz2017reluplex}
Guy Katz, Clark Barrett, David~L Dill, Kyle Julian, and Mykel~J Kochenderfer.
\newblock Reluplex: An efficient smt solver for verifying deep neural networks.
\newblock In \emph{International Conference on Computer Aided Verification},
  pages 97--117. Springer, 2017{\natexlab{b}}.

\bibitem[Katz et~al.(2019)Katz, Huang, Ibeling, Julian, Lazarus, Lim, Shah,
  Thakoor, Wu, Zelji{\'c}, et~al.]{katz2019marabou}
Guy Katz, Derek~A Huang, Duligur Ibeling, Kyle Julian, Christopher Lazarus,
  Rachel Lim, Parth Shah, Shantanu Thakoor, Haoze Wu, Aleksandar Zelji{\'c},
  et~al.
\newblock The marabou framework for verification and analysis of deep neural
  networks.
\newblock In \emph{International Conference on Computer Aided Verification},
  pages 443--452. Springer, 2019.

\bibitem[Khalil et~al.(2016)Khalil, Le~Bodic, Song, Nemhauser, and
  Dilkina]{khalil2016learning}
Elias~Boutros Khalil, Pierre Le~Bodic, Le~Song, George Nemhauser, and Bistra
  Dilkina.
\newblock Learning to branch in mixed integer programming.
\newblock \emph{Thirtieth AAAI Conference on Artificial Intelligence}, 2016.

\bibitem[Kingma and Ba(2015)]{kingma2014adam}
Diederik~P Kingma and Jimmy Ba.
\newblock Adam: A method for stochastic optimization.
\newblock In \emph{International Conference on Learning Representations}, 2015.

\bibitem[Krizhevsky et~al.(2012)Krizhevsky, Sutskever, and
  Hinton]{krizhevsky2012imagenet}
Alex Krizhevsky, Ilya Sutskever, and Geoffrey~E Hinton.
\newblock Imagenet classification with deep convolutional neural networks.
\newblock In \emph{Advances in neural information processing systems}, pages
  1097--1105, 2012.

\bibitem[Maas et~al.(2013)Maas, Hannun, and Ng]{maas2013rectifier}
Andrew~L Maas, Awni~Y Hannun, and Andrew~Y Ng.
\newblock Rectifier nonlinearities improve neural network acoustic models.
\newblock In \emph{in ICML Workshop on Deep Learning for Audio, Speech and
  Language Processing}. Citeseer, 2013.

\bibitem[Paszke et~al.(2017)Paszke, Gross, Chintala, Chanan, Yang, DeVito, Lin,
  Desmaison, Antiga, and Lerer]{paszke2017automatic}
Adam Paszke, Sam Gross, Soumith Chintala, Gregory Chanan, Edward Yang, Zachary
  DeVito, Zeming Lin, Alban Desmaison, Luca Antiga, and Adam Lerer.
\newblock \emph{Automatic differentiation in pytorch}, 2017.

\bibitem[Raghunathan et~al.(2018)Raghunathan, Steinhardt, and
  Liang]{raghunathan2018certified}
Aditi Raghunathan, Jacob Steinhardt, and Percy Liang.
\newblock Certified defenses against adversarial examples.
\newblock In \emph{International Conference on Learning Representations}, 2018.

\bibitem[Royo et~al.(2019)Royo, Calandra, Stipanovic, and Tomlin]{royo2019fast}
Vicenc~Rubies Royo, Roberto Calandra, Dusan~M Stipanovic, and Claire Tomlin.
\newblock Fast neural network verification via shadow prices.
\newblock \emph{arXiv preprint arXiv:1902.07247}, 2019.

\bibitem[Singh et~al.(2020)Singh, Maurer, Mller, Mirman, Gehr, Hoffmann,
  Tsankov, Drachsler~Cohen, Pschel, and Vechev]{singh2020eran}
Gagandeep Singh, Jonathan Maurer, Christoph Mller, Matthew Mirman, Timon Gehr,
  Adrian Hoffmann, Petar Tsankov, Dana Drachsler~Cohen, Markus Pschel, and
  Martin Vechev.
\newblock Eth robustness analyzer for neural networks (eran), 2020.
\newblock URL \url{https://github.com/eth-sri/eran}.

\bibitem[Szegedy et~al.(2013)Szegedy, Zaremba, Sutskever, Bruna, Erhan,
  Goodfellow, and Fergus]{szegedy2013intriguing}
Christian Szegedy, Wojciech Zaremba, Ilya Sutskever, Joan Bruna, Dumitru Erhan,
  Ian Goodfellow, and Rob Fergus.
\newblock Intriguing properties of neural networks.
\newblock \emph{arXiv preprint arXiv:1312.6199}, 2013.

\bibitem[Tjeng et~al.(2019)Tjeng, Xiao, and Tedrake]{newMIP}
Vincent Tjeng, Kai Xiao, and Russ Tedrake.
\newblock Evaluating robustness of neural networks with mixed integer
  programming.
\newblock \emph{International Conference on Learning Representations}, 2019.

\bibitem[VNN-COMP(2020)]{VNNComp}
VNN-COMP.
\newblock International verification of neural networks competition (vnn-comp),
  2020.
\newblock URL \url{https://sites.google.com/view/vnn20/vnncomp}.

\bibitem[Wang et~al.(2018{\natexlab{a}})Wang, Pei, Whitehouse, Yang, and
  Jana]{Wang2018}
Shiqi Wang, Kexin Pei, Justin Whitehouse, Junfeng Yang, and Suman Jana.
\newblock Efficient formal safety analysis of neural networks.
\newblock \emph{Conference on Neural Information Processing Systems},
  2018{\natexlab{a}}.

\bibitem[Wang et~al.(2018{\natexlab{b}})Wang, Pei, Whitehouse, Yang, and
  Jana]{wang2018formal}
Shiqi Wang, Kexin Pei, Justin Whitehouse, Junfeng Yang, and Suman Jana.
\newblock Formal security analysis of neural networks using symbolic intervals.
\newblock \emph{27th $\{$USENIX$\}$ Security Symposium ($\{$USENIX$\}$ Security
  18)}, pages 1599--1614, 2018{\natexlab{b}}.

\bibitem[Weiss et~al.(2012)Weiss, Natarajan, Peissig, McCarty, and
  Page]{weiss2012machine}
Jeremy~C Weiss, Sriraam Natarajan, Peggy~L Peissig, Catherine~A McCarty, and
  David Page.
\newblock Machine learning for personalized medicine: Predicting primary
  myocardial infarction from electronic health records.
\newblock \emph{AI Magazine}, 33\penalty0 (4):\penalty0 33--33, 2012.

\bibitem[Weng et~al.(2018)Weng, Zhang, Chen, Song, Hsieh, Boning, Dhillon, and
  Daniel]{Weng2018}
Tsui-Wei Weng, Huan Zhang, Hongge Chen, Zhao Song, Cho-Jui Hsieh, Duane Boning,
  Inderjit~S Dhillon, and Luca Daniel.
\newblock Towards fast computation of certified robustness for relu networks.
\newblock \emph{International Conference on Machine Learning}, 2018.

\bibitem[Wong and Kolter(2018)]{Wong2018}
Eric Wong and Zico Kolter.
\newblock Provable defenses against adversarial examples via the convex outer
  adversarial polytope.
\newblock \emph{International Conference on Machine Learning}, 2018.

\bibitem[Zhang et~al.(2018)Zhang, Weng, Chen, Hsieh, and
  Daniel]{zhang2018efficient}
Huan Zhang, Tsui-Wei Weng, Pin-Yu Chen, Cho-Jui Hsieh, and Luca Daniel.
\newblock Efficient neural network robustness certification with general
  activation functions.
\newblock In \emph{Advances in neural information processing systems}, pages
  4939--4948, 2018.

\end{thebibliography}
